%% file: main.tex
\documentclass{article}
\usepackage[final]{neurips_2024}

\usepackage[utf8]{inputenc}
\usepackage[T1]{fontenc}
\usepackage[colorlinks=true,linkcolor=Blue,urlcolor=Blue, citecolor=Blue,anchorcolor=Blue]{hyperref}
\usepackage{url}
\usepackage{booktabs}
\usepackage{amsfonts}
\usepackage{nicefrac}
\usepackage{microtype}

\usepackage{amsmath}
\usepackage[flushleft]{threeparttable}
\usepackage{algorithm}
\usepackage{algorithmic}
\usepackage{thmtools}
\usepackage{multirow}
\usepackage{adjustbox}
\usepackage{caption}
\usepackage{subcaption}
\usepackage[flushleft]{threeparttable}
\usepackage{enumitem}
\usepackage{array}
\usepackage{lipsum}
\usepackage[dvipsnames]{xcolor}

\title{NeuralFuse: Learning to Recover the Accuracy of Access-Limited Neural Network Inference in Low-Voltage Regimes}

\author{%
  Hao-Lun Sun$^{1}$,  Lei Hsiung$^{2}$,  Nandhini Chandramoorthy$^{3}$,  Pin-Yu Chen$^{3}$, Tsung-Yi Ho$^{4}$\\\\
  $^1$ National Tsing Hua University ~ $^2$ Dartmouth College ~ $^3$ IBM Research \\
  $^4$ The Chinese University of Hong Kong\\
  \texttt{s109062594@m109.nthu.edu.tw} \\
  \texttt{lei.hsiung.gr@dartmouth.edu} \\
  \texttt{\{pin-yu.chen, nandhini.chandramoorthy\}@ibm.com} \\
  \texttt{tyho@cse.cuhk.edu.hk}
}

\begin{document}
\maketitle

\input{sections/0_abstract}
\input{sections/1_introduction}
\input{sections/2_related_work}
\input{sections/3_method}

\input{sections/4_experiments}
\input{sections/5_conclusions}

\begin{ack}
We thank the anonymous reviewers for their insightful comments and valuable suggestions. The research described in this paper was conducted in the JC STEM Lab of Intelligent Design Automation, which is funded by The Hong Kong Jockey Club Charities Trust in support of Tsung-Yi Ho.
\end{ack}

{
\small
\bibliographystyle{plainnat}
\bibliography{reference}
}

\newpage
\appendix
\onecolumn
\input{sections/6_supplementary}
\end{document}

%% file: sections/0_abstract.tex
\begin{abstract}
Deep neural networks (DNNs) have become ubiquitous in machine learning, but their energy consumption remains problematically high. An effective strategy for reducing such consumption is supply-voltage reduction, but if done too aggressively, it can lead to accuracy degradation. This is due to random bit-flips in static random access memory (SRAM), where model parameters are stored. To address this challenge, we have developed NeuralFuse, a novel add-on module that handles the energy-accuracy tradeoff in low-voltage regimes by learning input transformations and using them to generate error-resistant data representations, thereby protecting DNN accuracy in both nominal and low-voltage scenarios. As well as being easy to implement, NeuralFuse can be readily applied to DNNs with limited access, such cloud-based APIs that are accessed remotely or non-configurable hardware. Our experimental results demonstrate that, at a 1\% bit-error rate, NeuralFuse can reduce SRAM access energy by up to 24\% while recovering accuracy by up to 57\%. To the best of our knowledge, this is the first approach to addressing low-voltage-induced bit errors that requires no model retraining.\let\thefootnote\relax\footnotetext{Project Page: \url{https://trustsafeai-neuralfuse.static.hf.space}}\let\thefootnote\relax\footnotetext{Code: \url{https://github.com/IBM/NeuralFuse}}
\end{abstract}

%% file: sections/1_introduction.tex
\vspace{-3mm}\section{Introduction}\label{sec:introduction}\vspace{-1mm}
Energy-efficient computing is of primary importance to the effective deployment of deep neural networks (DNNs), particularly in edge devices and in on-chip AI systems. Increasing DNN computation’s energy efficiency and lowering its carbon footprint require iterative efforts from both chip designers and algorithm developers. Processors with specialized hardware accelerators for AI computing, capable of providing orders of magnitude better performance and energy efficiency for AI computation, are now ubiquitous. However, alongside reduced precision/quantization and architectural optimizations, endowing such systems with the capacity for low-voltage operation is a powerful lever for reducing their power consumption.

The computer engineering literature contains ample evidence of the effects of undervolting and low-voltage operation on accelerator memories that store weights and activations during computation. Aggressive scaling-down of static random access memory’s (SRAM’s) supply voltage to below the rated value saves power, thanks to the quadratic dependence of dynamic power on voltage.  Crucially, however, it also leads to an exponential increase in bit failures. Memory bit flips cause errors in the stored weight and activation values \citep{ResilientLowVolNandhini, BitErrorDAC}, leading to catastrophic accuracy loss.

A recent wave of research has proposed numerous techniques for allowing low-voltage operation of DNN accelerators while preserving their accuracy. Most of these have been either hardware-based error-mitigation techniques or error-aware robust training of DNN models. On-chip error mitigation methods have significant performance and power overheads \citep{ResilientLowVolNandhini, Minerva}. On the other hand, some have proposed to generate models that are robust to bit errors via a specific learning algorithm \citep{matic, eden, ErrorAwareTraining}, thereby eliminating the need for on-chip error mitigation. However, error-aware robust training to find the optimal set of robust parameters for each model is time- and energy-intensive and may not be possible in all access-limited settings.

\input{assets/_figures/introduction.tex}

In this paper, therefore, we propose a novel model-agnostic approach: \textit{NeuralFuse}. This proof-of-concept machine-learning module offers trainable input transformation parameterized by a relatively small DNN; and, by enhancing input’s robustness, it mitigates bit errors caused by very low-voltage operation, thus serving the wider goal of more accurate inferencing. The pipeline of NeuralFuse is illustrated in Figure \ref{fig:Introduction}. To protect the deployed models from making wrong predictions under low-power conditions, NeuralFuse accepts scenarios under access-limited neural networks (e.g., non-configurable hardware or cloud-based APIs). Specifically, we consider two access-limited scenarios that are common in the real world: 1) \textit{relaxed access}, in which `black box' model details are unknown, but backpropagation through those models is possible; and 2) \textit{restricted access}, in which the model details are unknown and backpropagation is disallowed. To enable it to deal with relaxed access, we trained NeuralFuse via backpropagation, and for restricted-access cases, we trained it on a white-box surrogate model. To the best of our knowledge, this is the first study that leverages a learning-based method to address random bit errors as a means of recovering accuracy in low-voltage and access-limited settings.

We summarize our \textbf{main contributions} as follows:
\begin{itemize}[leftmargin=*]
  \setlength{\itemsep}{2pt}
  \item We propose \textit{NeuralFuse}, a novel learning-based input-transformation module aimed at enhancing the accuracy of DNNs that are subject to random bit errors caused by very low voltage operation. NeuralFuse is model-agnostic, i.e., operates on a plug-and-play basis at the data-input stage and does not require any re-training of deployed DNN models.
  
  \item We explore two practical limited-access scenarios for neural-network inference: relaxed access and restricted access. In the former setting, we use gradient-based methods for module training. In the latter one, we use a white-box surrogate model for training, which is highly transferable to other types of DNN architecture.
  
  \item We report the results of an extensive program of experiments with various combinations of DNN models (ResNet18, ResNet50, VGG11, VGG16, and VGG19), datasets (CIFAR-10, CIFAR-100, GTSRB, and ImageNet-10), and NeuralFuse implementations of different architectures and sizes.
  These show that NeuralFuse can consistently increase the perturbed accuracy (accuracy evaluated under random bit errors in weights) by up to $57\%$, while simultaneously saving up to $24\%$ of the energy normally required for SRAM access, based on our realistic characterization of bit-cell failures for a given memory array in a low-voltage regime inducing a $0.5\%/1\%$ bit-error rate.

  \item 
  We demonstrate NeuralFuse's transferability (i.e., adaptability to unseen base models), versatility (i.e., ability to recover low-precision quantization loss), and competitiveness (i.e., state-of-the-art performance) in various scenarios, establishing it as a promising proof-of-concept for energy-efficient, resilient DNN inference.
\end{itemize}

%% file: assets/_figures/introduction.tex
\begin{figure*}
     \begin{subfigure}[b]{.648\linewidth}
         \centering
         \includegraphics[width=\textwidth]{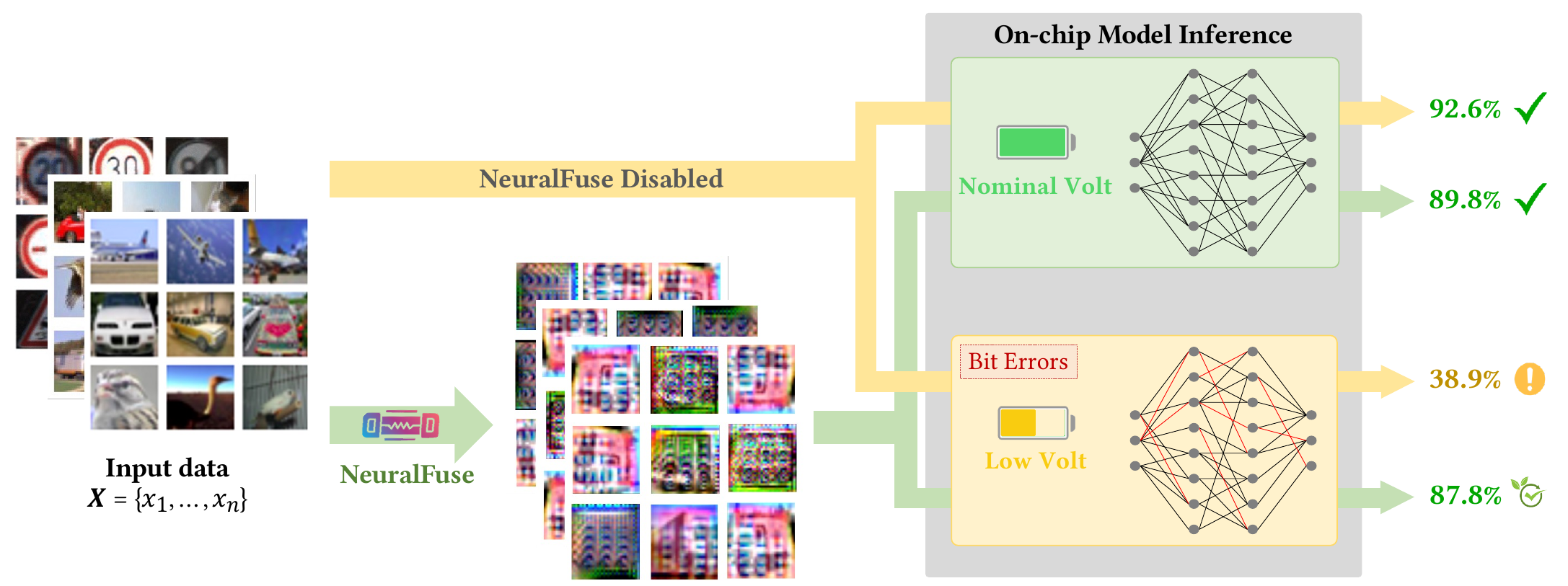}
         \caption{The pipeline of the NeuralFuse framework at inference.}
         \label{fig:system_plot}
     \end{subfigure}
     \hfill
     \begin{subfigure}[b]{0.348\linewidth}
         \centering
         \includegraphics[width=\textwidth]{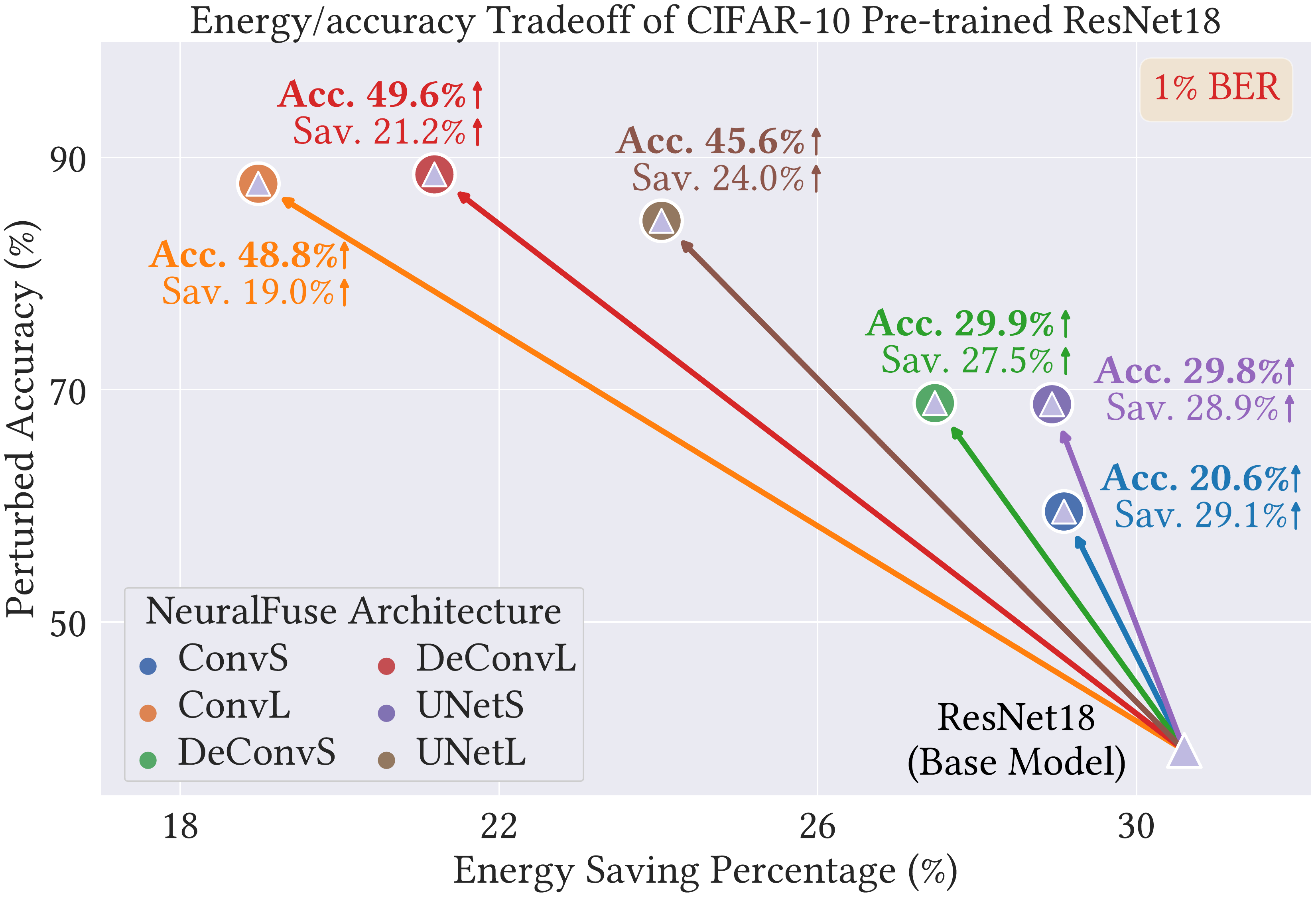}
         \caption{Energy/accuracy tradeoff example.}
         \label{fig:result}
     \end{subfigure}
     \vspace{-5mm}
\caption{(a) At inference, NeuralFuse transforms input samples $\mathbf{x}$ into robust data representations. The \textit{nominal} voltage allows models to work as expected, whereas at \textit{low voltage}, one would encounter bit errors (e.g., $1\%$) that cause incorrect inferences. The percentages reflect the accuracy of a CIFAR-10 pre-trained ResNet18 with and without NeuralFuse in both those voltage cases. (b) On the same base model (ResNet18), we illustrate the energy/accuracy tradeoff of six NeuralFuse implementations.
The x-axis represents the percentage reduction in dynamic-memory access energy at low-voltage settings (base model protected by NeuralFuse), as compared to the bit-error-free (nominal) voltage. The y-axis represents the perturbed accuracy (evaluated at low voltage) with a $1\%$ bit-error rate.}\vspace{-3mm}
\label{fig:Introduction}
\end{figure*}

%% file: sections/2_related_work.tex
\section{Related Work and Background}\label{sec:related_work_and_background}
\paragraph{Software-based Energy-saving Strategies.} Various recent studies have proposed software-based methods of reducing computing’s energy consumption. For instance, quantization techniques have been reported to reduce the precision required for storing model weights, and thus to decrease total memory storage \citep{CompressNNQuantization, XNORNETBinaryQuantization, QuantizationCNNMobile}. On the other hand, \citet{EnergyAwarePuring} - who proposed energy-aware pruning on each layer and fine-tuning of weights to maximize final accuracy - suggested several ways to reduce DNNs’ energy consumption. For example, they devised the ECC framework, which compresses DNN models to meet a given energy constraint \citep{ECCModelCompression}, and a method of compressing such models via joint pruning and quantization \citep{PruneAndQuantize}. It is also feasible, during DNN training, to treat energy constraints as an optimization problem and thereby reduce energy consumption while maximizing training accuracy \citep{EnergyConstrainedCompression}. However, unlike ours, all these methods imply changing either model architectures or model weights.

\paragraph{Hardware-based Energy-saving Strategies.} Prior studies have also explored ways of improving energy efficiency via specially designed hardware. Several of them have focused on the undervolting of DNN accelerators and proposed methods to maintain accuracy in the presence of bit errors. For instance, \citet{Minerva} proposed an SRAM fault-mitigation technique that rounds faulty weights to zero to avoid degradation of prediction accuracy. \citet{SRAMMSBPRESERVE} recommended storing sensitive MSBs (most significant bits) in robust SRAM cells to preserve accuracy.  \citet{ResilientLowVolNandhini} proposed dynamic supply-voltage boosting to improve the resilience of memory-access operations; and the learning-based approach proposed by \citet{ErrorAwareTraining} aims to find models that are robust to bit errors. The latter paper discusses several techniques for improving such robustness, notably quantization, weight-clipping, random bit-error training, and adversarial bit-error training. Its authors concluded from their experiments that a combination of quantization, weight-clipping, and adversarial bit-error training will yield excellent performance. However, they also admitted that the relevant training process was sensitive to hyperparameter settings, and hence, it might come with a challenging training procedure. 

However, we suggest that all the methods mentioned above are difficult to implement and/or unsuitable for use in real-world access-limited settings. For example, the weights of DNN models packed on embedded systems may not be configurable or updatable, making model retraining (e.g., \citet{ErrorAwareTraining}) non-viable in that scenario. Moreover, DNN training is already a tedious and time-consuming task, so adding error-aware training to it may further increase its complexity and, in particular, make hyperparameter searches more challenging. 
\citet{robustnessWBF} also reported that error-aware training was ineffective for large DNNs with millions of bits. NeuralFuse obviates the need for model retraining via an add-on trainable input-transformation function parameterized by a relatively small secondary DNN.

\input{assets/_figures/random_bit_error.tex}

\paragraph{SRAM Bit Errors in DNNs.} Low voltage-induced memory bit-cell failures can cause bit-flips from 0 to 1 and vice versa. In practice, SRAM bit errors increase exponentially when the supply voltage is scaled below $V_{min}$, i.e., the minimum voltage required to avoid them. This phenomenon has been studied extensively in the prior literature, including work by \citet{ResilientLowVolNandhini} and \citet{BitErrorDAC}. The specific increases in bit errors as voltage scales down, in the case of an SRAM array of $512\times64$ bits with a 14nm technology node, is illustrated in Figure \ref{fig:ber_energy_voltage}. The corresponding dynamic energy per SRAM read access, measured at each voltage at a constant frequency, is shown on the right-hand side of the figure. In this example, accessing the SRAM at 0.83$V_{min}$ leads to a $1\%$ bit-error rate, and at the same time, dynamic energy per access is reduced by approximately $30\%$. This can lead to DNNs making inaccurate inferences, particularly when bit-flips occur at the MSBs. However, improving robustness to bit errors can allow us to lower $V_{min}$ and exploit the resulting energy savings.

It has been observed that bit-cell failures for a given memory array are randomly distributed and independent of each other. That is, the spatial distribution of bit-flips can be assumed to be random, as it generally differs from one array to another, within as well as between chips. Below, following \citet{ResilientLowVolNandhini}, we model bit errors in a memory array of a given size by generating a random distribution of such errors with equal likelihood of 0-to-1 and 1-to-0 bit-flipping. More specifically, we assume that the model weights are quantized to 8-bit precision (i.e., from 32-bit floats to 8-bit integers), and generate perturbed models by injecting our randomly distributed bit errors into the two's complement representation of weights. For more implementation details, please refer to Section \ref{subsec:experiment_setup}.\looseness-1

%% file: assets/_figures/random_bit_error.tex
\begin{figure}[t]
\centering
\includegraphics[width=.7\linewidth]{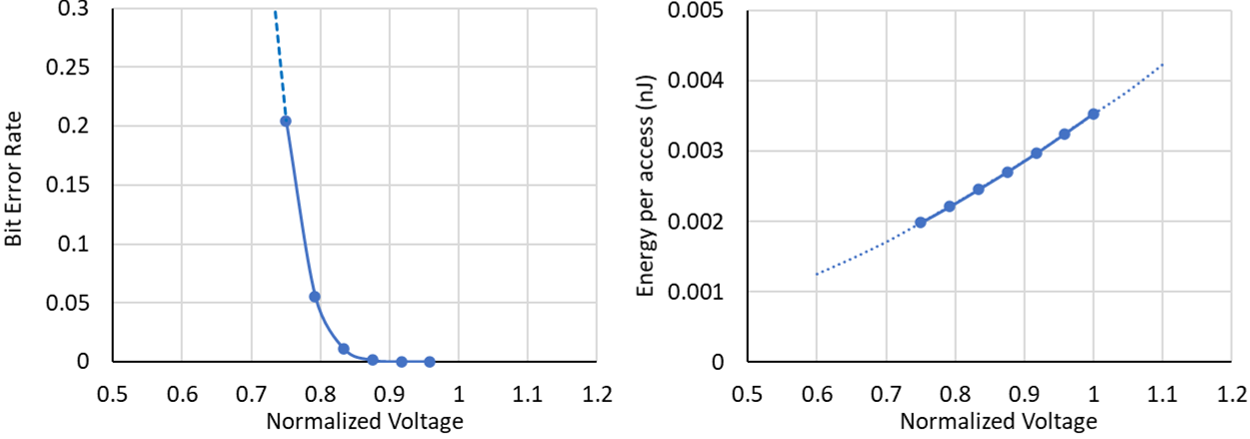}\vspace{-2mm}
\caption{The bit-error rates (left) and dynamic energy per memory access versus voltage for static random access memory arrays (right) as reported by \citet{ResilientLowVolNandhini}. The x-axis shows voltages normalized with respect to the minimum bit error-free voltage ($V_{min}$).}\label{fig:ber_energy_voltage}\vspace{-3mm}
\end{figure}

%% file: sections/3_method.tex
\section{NeuralFuse: Framework and Algorithms}\label{sec:NeuralFuse}
\subsection{Error-Resistant Input Transformation}\label{subsec:error_resistant_input_transformation}
As illustrated in Figure \ref{fig:Introduction}, we propose a novel trainable input-transformation module, NeuralFuse, parametrized by a relatively small DNN, to mitigate the accuracy-energy tradeoff for model inference and thus overcome the drawback of performance degradation in low-voltage regimes. A specially designed loss function and training scheme are used to derive NeuralFuse and apply it to the input data such that the transformed inputs will become robust to low voltage-induced bit errors.

Consider the input $\mathbf{x}$ sampled from the data distribution $\mathcal{X}$ and a set of models $\mathcal{M}_{p}$ with $p\%$ random bit errors on weights (i.e., \textit{perturbed} models). When it is not manifesting any bit errors (i.e., at normal-voltage settings), the perturbed model operates as a nominal deterministic one, denoted by $M_{0}$. NeuralFuse aims to ensure that a model $M_p\in\mathcal{M}_p$ can make correct inferences on the transformed inputs while also delivering consistent results in its $M_{0}$ state. To adapt to various data characteristics, NeuralFuse -- designated as $\mathcal{F}$ in Eq. (\ref{eqn:neuralfuse_formulation}), below -- is designed to be input-aware. This characteristic can be formally defined as
\begin{equation}
\mathcal{F}(\mathbf{x})=\text{clip}_{[-1,1]}\big(\mathbf{x}+\mathcal{G}(\mathbf{x})\big)\text{,}
\label{eqn:neuralfuse_formulation}
\end{equation}
where $\mathcal{G}(\mathbf{x})$ is a ``generator'' (i.e., an input-transformation function) that can generate a perturbation based on input $\mathbf{x}$. As transformed by NeuralFuse, i.e., as $\mathcal{F}(\mathbf{x})$, that input is passed to the deployed model ($M_0$ or $M_p$) for final inference. Without loss of generality, we assume the transformed input lies within a scaled input range $\mathcal{F}(\cdot) \in [-1,1]^d$, where $d$ is the (flattened) dimension of $\mathbf{x}$.

\subsection{Training Objective and Optimizer}\label{subsec:training_objectives}
To train our generator $\mathcal{G}(\cdot)$, which ought to be able to ensure the correctness of both the perturbed model $M_p$ and the clean model $M_0$, we parameterized it with a neural network and apply our training objective function
\begin{align}
 \arg\max_{\mathcal{W}_{\mathcal{G}}}{\log\mathcal{P}_{M_0}(y|{\mathcal{F}(\mathbf{x};\mathcal{W}_{\mathcal{G}})})} + \lambda \cdot{\mathbf{E}_{M_p \sim \mathcal{M}_p }}[ {\log\mathcal{P}_{M_p}(y|{\mathcal{F}(\mathbf{x};\mathcal{W}_{\mathcal{G}})})}]\text{,}
\end{align}
where $\mathcal{W}_{\mathcal{G}}$ is the set of trainable parameters for $\mathcal{G}$; $y$ is the ground-truth label of $\mathbf{x}$; $\mathcal{P}_M$ denotes the likelihood of $y$ as computed by a model $M$ being given a transformed input $\mathcal{F}(\mathbf{x};\mathcal{W}_{\mathcal{G}})$;
$\mathcal{M}_p$ is the distribution of the perturbed models inherited from the clean model $M_0$, under a $p\%$ random bit-error rate; and $\lambda$ is a hyperparameter that balances the importance of the nominal and perturbed models.

The training objective function can be readily converted to a loss function ($\mathcal{L}$) that evaluates cross-entropy between the ground-truth label $y$ and the prediction $\mathcal{P}_{M}(y|{\mathcal{F}(\mathbf{x};\mathcal{W}_{\mathcal{G}})}$. That is, the total loss function can be calculated as
\begin{equation}
  \mathcal{L}_\text{Total} = \mathcal{L}_{M_0} + \lambda \cdot{\mathcal{L}_{\mathcal{M}_p}}\text{.}
\label{eqn:loss_function}
\end{equation}
In particular, optimizing the loss function requires evaluation of the impact of the loss term $\mathcal{L}_{\mathcal{M}_p}$ on randomly perturbed models. Our training process is inspired by expectation over transformation (EOT) attacks \citep{EOTAttack}, which aim to produce robust adversarial examples that are simultaneously adversarial over the entire transformation distribution. Based on that idea, we propose a new optimizer for solving Eq.~\eqref{eqn:loss_function}, which we call expectation over perturbed models (EOPM). EOPM-trained generators can generate error-resistant input transformations and mitigate inherent bit errors. However, it would be computationally impossible to enumerate all possible perturbed models with random bit errors, and the number of realizations for perturbed models is constrained by the memory size of the GPUs used for training. In practice, therefore, we only use $N$ perturbed models per iteration to calculate empirical average loss, i.e.,
\begin{equation}
\mathcal{L}_{\mathcal{M}_p} \approx \frac{ {\mathcal{L}_{M_{p_{1}}}+\cdots+{\mathcal{L}_{M_{p_{N}} }}}}{N}\text{,}
\label{eqn:empirical_loss_function}
\end{equation}
where $N$ is the number of perturbed models $\{M_{p_{1}},\cdots,M_{p_{N}}\}$ that are simulated to calculate the loss caused by random bit errors. Therefore, the gradient used to update the generator can be calculated as follows:
\begin{equation}
\scalebox{1.1}{$
\frac{\partial{\mathcal{L}_\text{Total}}}{\partial{\mathcal{W}_{\mathcal{G}}}}= \frac{\partial{\mathcal{L}_{M_0}}}{\partial{\mathcal{W}_{\mathcal{G}}}} + \frac{\lambda}{N} \left({\frac{\partial{\mathcal{L}_{M_{p_{1}}}}}{\partial{\mathcal{W}_{\mathcal{G}}}}+\cdots+\frac{\partial{\mathcal{L}_{M_{p_{N}}}}}{\partial{\mathcal{W}_{\mathcal{G}}}}} \right)\text{.}$
}
\label{eqn:gradient_calculation}
\end{equation}
Through our implementation, we found that stable performance could be delivered when $N=10$, and that there was little to be gained by using a larger value. The results of our ablation study for different values of $N$ can be found in Appendix \ref{sec:appendix_ablation}.

\subsection{Training Algorithm}\label{subsec:training_algorithm}
Algorithm \ref{alg:EOPM_Algo} in Appendix \ref{sec:appendix_training_algorithm} summarizes NeuralFuse’s training steps. Briefly, this involves splitting the training data $\mathcal{X}$ into $B$ mini-batches for training the generator in each epoch. For each mini-batch, we first feed these data into $\mathcal{F}(\mathbf{\cdot})$ to obtain the transformed inputs. Also, we simulate $N$ perturbed models using a $p\%$ random bit-error rate, denoted by $M_{p_{1}}, \cdots, M_{p_{N}}$, from $\mathcal{M}_p$. Then, the transformed inputs are fed into those $N$ perturbed models as well as into the clean model $\mathcal{M}_0$, and their respective losses and gradients are calculated. Finally, NeuralFuse parameters $\mathcal{W}_{\mathcal{G}}$ are updated based on the gradient obtained by EOPM.

%% file: sections/4_experiments.tex
\section{Experiments}\label{sec:experiments}
\subsection{Experiment Setups}\label{subsec:experiment_setup}
\textbf{Datasets.}~ We evaluate NeuralFuse on four different datasets: CIFAR-10 \citep{CIFAR10}, CIFAR-100 \citep{CIFAR10}, the German Traffic Sign Recognition Benchmark (GTSRB) \citep{GTSRB}, and ImageNet-10 \citep{imagenet}. CIFAR-10 consists of 10 classes, with 50,000 training images and 10,000 testing images in total. Similarly, CIFAR-100 consists of 100 classes, with 500 training images and 100 testing images in each. The GTSRB contains 43 classes with a total of 39,209 training images and 12,630 testing images. Similar to CIFAR-10 and CIFAR-100, we resize GTSRB into 32$\times$32$\times$3 in our experiment. For ImageNet-10, we chose the same ten categories as \citet{ImageNet-10}, in which there are 13,000 training images and 500 test images cropped into 224$\times$224$\times$3. Due to space limitations, our CIFAR-100 results are presented in Appendices \ref{sec:appendix_relaxed_access} and \ref{sec:appendix_transferability}.

\textbf{Base Models.}~ We selected several common architectures for our base models: ResNet18, ResNet50 \citep{ResNet}, VGG11, VGG16, and VGG19 \citep{VGG}. To replicate the deployment of models on chips, all our based models were given quantization-aware training that followed \citet{ErrorAwareTraining}.

\textbf{NeuralFuse Generators.}~ The architecture of the NeuralFuse generator ($\mathcal{G}$) ) is based on an encoder-decoder structure. We designed and compared three types of generators, namely convolution-based, deconvolution-based, and UNet-based. We also considered large(L)/small(S) network sizes for each type. Further details can be found below and in Appendix \ref{sec:appendix_NeuralFuse_implementation}.
\begin{itemize}[leftmargin=*]

\item {\bf Convolution-based (Conv).}~ Conv uses convolution with MaxPool layers for its encoder and \textit{convolution with UpSample layers} for its decoder. This architecture has previously been shown to be efficient and effective at generating \textit{input-aware} backdoor triggers \citep{InputAware}.

\item {\bf Deconvolution-based (DeConv).}~ DeConv uses convolution with MaxPool layers for its encoder and \textit{deconvolution layers} for its decoder. We expected this modification both to enhance its performance and to reduce its energy consumption.

\item {\bf UNet-based (UNet).}~ UNet uses convolution with MaxPool layers for its encoder, and deconvolution layers for its decoder. UNet is known for its robust performance in image segmentation \citep{UNet}.
\end{itemize}
\textbf{Energy-consumption Calculation.}~ The energy consumption reported in Figure \ref{fig:Introduction} is based on the product of the total number of SRAM memory accesses in a systolic array-based convolution neural network (CNN) accelerator and the dynamic energy per read access at a given voltage. Research by \citet{eyeriss} previously showed that energy consumption by SRAM buffers and arrays accounts for a high proportion of total system energy consumption. 
We assume that there are no bit errors on NeuralFuse, given its status as an add-on data preprocessing module whose functions could also be performed by a general-purpose core. In this work we assume it is implemented on the accelerator equipped with dynamic voltage scaling and therefore NeuralFuse computation is performed at nominal error-free voltage. We report a reduction in overall weight-memory energy consumption (i.e., NeuralFuse + Base Model under a $p\%$ bit-error rate) with respect to the unprotected base model in the regular-voltage mode (i.e., $0\%$ bit-error rate and without NeuralFuse).

To quantify memory accesses, we used the SCALE-SIM simulator \citep{ScaleSim}, and our chosen configuration simulated an output-stationary dataflow and a 32$\times$32 systolic array with 256KB of weight memory. We collected data on the dynamic energy per read access of the SRAM both at $V_{min}$ and at the voltage corresponding to a $1\%$ bit-error rate ($V_{ber}\approx0.83V_{min}$) from Cadence ADE Spectre simulations, both at the same clock frequency.

\textbf{Relaxed and Restricted Access Settings.}~ In the first of our experiments’ two scenarios, relaxed access, the base-model information was not entirely transparent, but allowed us to obtain gradients from the black-box model through backpropagation. Therefore, this scenario allowed direct training of NeuralFuse with the base model using EOPM. In the restricted-access scenario, on the other hand, only the inference function was allowed for the base model, and we therefore trained NeuralFuse using a white-box surrogate base model and then transfering the generator to the access-restricted model.\looseness-1

\input{assets/_figures/figure_3}
\textbf{Computing Resources.}~ Our experiments were performed using eight Nvidia Tesla V100 GPUs and implemented with PyTorch. NeuralFuse was found to generally take 150 epochs to converge, and its training time was similar to that of the base model it incorporated. On both the CIFAR-10 and CIFAR-100 datasets, average training times were 17 hours (ResNet18), 50 hours (ResNet50), 9 hours (VGG11), 13 hours (VGG16), and 15 hours (VGG19). For GTSRB, the average training times were 9 hours (ResNet18), 27 hours (ResNet50), 5 hours (VGG11), 7 hours (VGG16), and 8 hours (VGG19); and for ImageNet-10, the average training times were 32 hours (ResNet18), 54 hours (ResNet50), 50 hours (VGG11), 90 hours (VGG16), and 102 hours (VGG19).

\subsection{Performance Evaluation, Relaxed-access Scenario}\label{subsec:relaxed_access_performance}
Our experimental results pertaining to the relaxed-access scenario are shown in Figure \ref{fig:experimental_results_main1}. The bit-error rate (BER) due to low voltage was $1\%$ in the cases of CIFAR-10 and GTSRB, and $0.5\%$ for ImageNet-10. The BER of ImageNet-10 was lower than that of the other two because, being pre-trained, it has more parameters than either of them. For each experiment, we sampled and evaluated $N = 10$ perturbed models (independent from training), and below, we report the means and standard deviations of their respective accuracies. Below, clean accuracy (CA) refers to a model’s accuracy measured at nominal voltage, and perturbed accuracy (PA) to its accuracy measured at low voltage.\looseness-1

In the cases of CIFAR-10 and the GTSRB, we observed that large generators like ConvL and UNetL recovered PA considerably, i.e., in the range of $41\%$ to $63\%$ on ResNet18, VGG11, VGG16, and VGG19. ResNet50’s recovery percentage was slightly worse than those of the other base models, but it nevertheless attained up to $51\%$ recovery on the GTSRB. On the other hand, the recovery percentages achieved when we used small generators like DeConvS were worse than those of their larger counterparts. This could be explained by larger-sized networks’ better ability to learn error-resistant generators (though perhaps at the cost of higher energy consumption). In the case of ImageNet-10, using larger generators also yielded better PA performance recovery, further demonstrating NeuralFuse’s ability to work well with large input sizes and varied datasets.

\subsection{Performance Evaluation, Restricted-access Scenario (Transferability)}\label{subsec:restricted_access_transferability}
The experimental results of our restricted-access scenario are shown in Table \ref{table:Maintransferabilityp1}. We adopted ResNet18 and VGG19 as our white-box surrogate source models for training the generators under a $1.5\%$ bit-error rate. We chose ConvL and UNetL as our generators because they performed best out of the six we tested (see Figure \ref{fig:experimental_results_main1}).

\input{assets/_tables/_tables_experiments_maintransferabilityp1}

From Table \ref{table:Maintransferabilityp1}, we can see that transferring from a larger BER such as $1.5\%$ can endow a smaller one (e.g., $1\%$) with strong resilience. The table also makes it clear that using VGG19 as a surrogate model with UNet-based generators like UNetL can yield better recovery performance than other model/generator combinations. On the other hand, we observed in some cases that if we transferred between source and target models of the same type (but with different BERs for training and testing), performance results could exceed those we had obtained during the original relaxed-access scenario. For instance, when we transferred VGG19 with UNetL under a $1.5\%$ BER to VGG19 or VGG11 under a $0.5\%$ BER, the resulting accuracies were $85.86\%$ (as against $84.99\%$ for original VGG19) and $84.81\%$ (as against $82.42\%$ for original VGG11). We conjecture that generators trained on relatively large BERs can cover the error patterns of smaller BERs, and even help improve the latter's generalization. These findings indicate the considerable promise of access-limited base models in low-voltage settings to recover accuracy.

\input{assets/_tables/_tables_Appendix_EnergySaving}

\subsection{Energy/Accuracy Tradeoff}\label{subsec:energy_acc_tradeoff}
We report total dynamic energy consumption as the total number of SRAM-access events multiplied by the dynamic energy of a single such event. Specifically, we used SCALE-SIM to calculate total weight-memory access (TWMA), specifics of which can be found in Appendix \ref{sec:appendix_energy_accuracy_tradeoff}'s Table \ref{table:Total_Mem_E}. 
In Table \ref{table:EnergySaving}, below, we report the percentages of energy saved (ES) at voltages that yield a $1\%$ bit-error rate for various base-model and generator combinations. The formula for computing ES is
\begin{equation}
\scalebox{1.05}{
$\text{ES} = \frac{\text{Energy}_{\textit{NV}} - \big(\text{Energy}_{\textit{LV}}+\text{Energy}_{\text{NeuralFuse at }\textit{NV}}\big)}{\text{Energy}_{\textit{NV}}}\times{100\%}$,
}
\end{equation}
where \textit{NV} denotes nominal voltage regime, and \textit{LV}, a low-voltage one.

Our results indicate that when ResNet18 is utilized as a base model, NeuralFuse can recover model accuracy by $20-49\%$ while reducing energy use by $19-29\%$. In Appendix \ref{sec:appendix_energy_accuracy_tradeoff}, we provide more results on the tradeoff between energy and accuracy of different NeuralFuse and base-model combinations. Overall, it would appear that using NeuralFuse can effectively restore model accuracy when SRAM encounters low-voltage-induced random bit errors.

\textbf{Runtime and Latency.}~ On the other hand, runtime and its corresponding energy consumption may also affect overall energy savings. For instance, previous research has shown that multiply-and-accumulate (MAC) operations account for more than 99\% of all operations in state-of-the-art CNNs, dominating processing runtime and energy consumption alike \cite{EnergyAwarePuring}. Therefore, we also report the results of our MAC-based energy-consumption estimation in Appendix \ref{sec:appendix_energy_accuracy_tradeoff}, and of our latency analysis in Appendix \ref{sec:appendix_latency}. Here, it should also be noted that an additional latency overhead is an inevitable tradeoff for reducing energy consumption in our scenarios. Although neither runtime nor latency is a major focus of this paper, future researchers could usefully design a lighter-weight version of the NeuralFuse module, or apply model-compression techniques to it, to reduce these two factors.\looseness-1

\input{assets/_figures_and_tables}

\subsection{Model Size and NeuralFuse Efficiency}\label{subsec:additional_analysis}
To arrive at a full performance characterization of NeuralFuse, we analyzed the relationship between the final recovery achieved by each base model in combination with generators of varying parameter counts. For this purpose, we defined \textit{efficiency ratio} as the recovery percentage in PA divided by NeuralFuse’s parameter count. Table \ref{table:EffRat} compares the efficiency ratios of all NeuralFuse generators trained on CIFAR-10. Those results show that UNet-based generators had better efficiency per million parameters than either convolution-based or deconvolution-based ones.

\subsection{NeuralFuse’s Robustness to Reduced-precision Quantization}\label{sec:appendix_quant_precision_loss_neuralfuse}
Lastly, we also explored NeuralFuse’s robustness to low-precision quantization on model weights. Uniform quantization is the usual method for quantizing model weights \citep{Quantizationsurvey}. However, it is possible for it to cause an accuracy drop due to lack of precision. Given our aim of offering protection from bit errors, we hoped to understand whether NeuralFuse could also recover model-accuracy drops caused by this phenomenon. We therefore uniformly quantized the model weights to a lower bit precision and measured the resulting accuracy. Specifically, we applied symmetric uniform quantization to our base models with various numbers of bits to induce precision loss, and defined the quantized weight $\mathbf{W_{q}}$ (integer) as $\mathbf{W_{q}=\lfloor\frac{\mathbf{W}}{s}\rceil}$, where $\mathbf{W}$ denotes the original model weight (full precision), $s=\frac{\max{\left|\mathbf{W}\right|}}{2^{b-1}-1}$ is the quantization scale parameter, and $b$ is the precision (number of bits) used to quantize the models. Bit errors induced by low voltage operation as previously described, are also applied to low precision weights.

We used the GTSRB pre-trained ResNet18 as our example in an evaluation of two NeuralFuse generators, i.e., ConvL and UNetL trained with $0.5\%$ BER, and varied precision $b$ from 8 bits to 2 bits (integer). The results, shown in Figure \ref{fig:low_bit_precision}, indicated that when $b>3$ bits, NeuralFuse could effectively recover accuracy in both the low-voltage and low-precision scenarios. When $b=3$, while NeuralFuse could still handle the bit-error-free model (Fig. \ref{fig:low_bit_precision} top), it exhibited a limited ability to recover the random bit-error case (Fig. \ref{fig:low_bit_precision} bottom). We find these results encouraging, insofar as NeuralFuse was only trained on random bit errors, yet demonstrated high accuracy in dealing with unseen bit-quantization errors. Further experimental results derived from other base models and datasets can be found in Appendix \ref{sec:appendix_low_precision_quantization}.

\subsection{Extended Analysis}
Here, we would like to highlight some key findings from the additional results in the Appendices. In Appendix \ref{sec:appendix_ablation}, we compare NeuralFuse against a simple baseline of learning a universal input perturbation. We found that such baseline performed much worse than NeuralFuse at that task, validating the necessity of adopting input-aware transformation if the goal is to learn error-resistant data representations in low-voltage scenarios. In Appendix \ref{sec:appendix_transferability}, we report that ensemble training of white-box surrogate base models could further improve the transferability of NeuralFuse in restricted-access scenarios. Appendices \ref{sec:visualization} and \ref{sec:visual_transformed} present visualization results of NeuralFuse’s data embeddings and transformed inputs. In Appendix \ref{sec:appendix_robust_model_neuralfuse}, we show that NeuralFuse can further recover the accuracy of a base model trained with adversarial weight perturbation in a low-voltage setting.

%% file: assets/_figures/figure_3.tex
\begin{figure*}[t]
     \centering
     \hfill
     \begin{subfigure}[b]{\textwidth}
         \centering
         \includegraphics[width=\textwidth, trim=-0.2cm -0.1cm 0.2cm 0.1cm, clip]{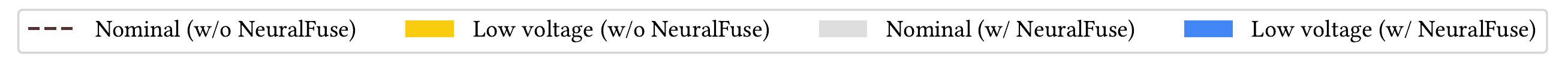}
     \end{subfigure}
     \hfill
     \begin{subfigure}[b]{0.195\textwidth}
         \centering
         \includegraphics[width=\textwidth]{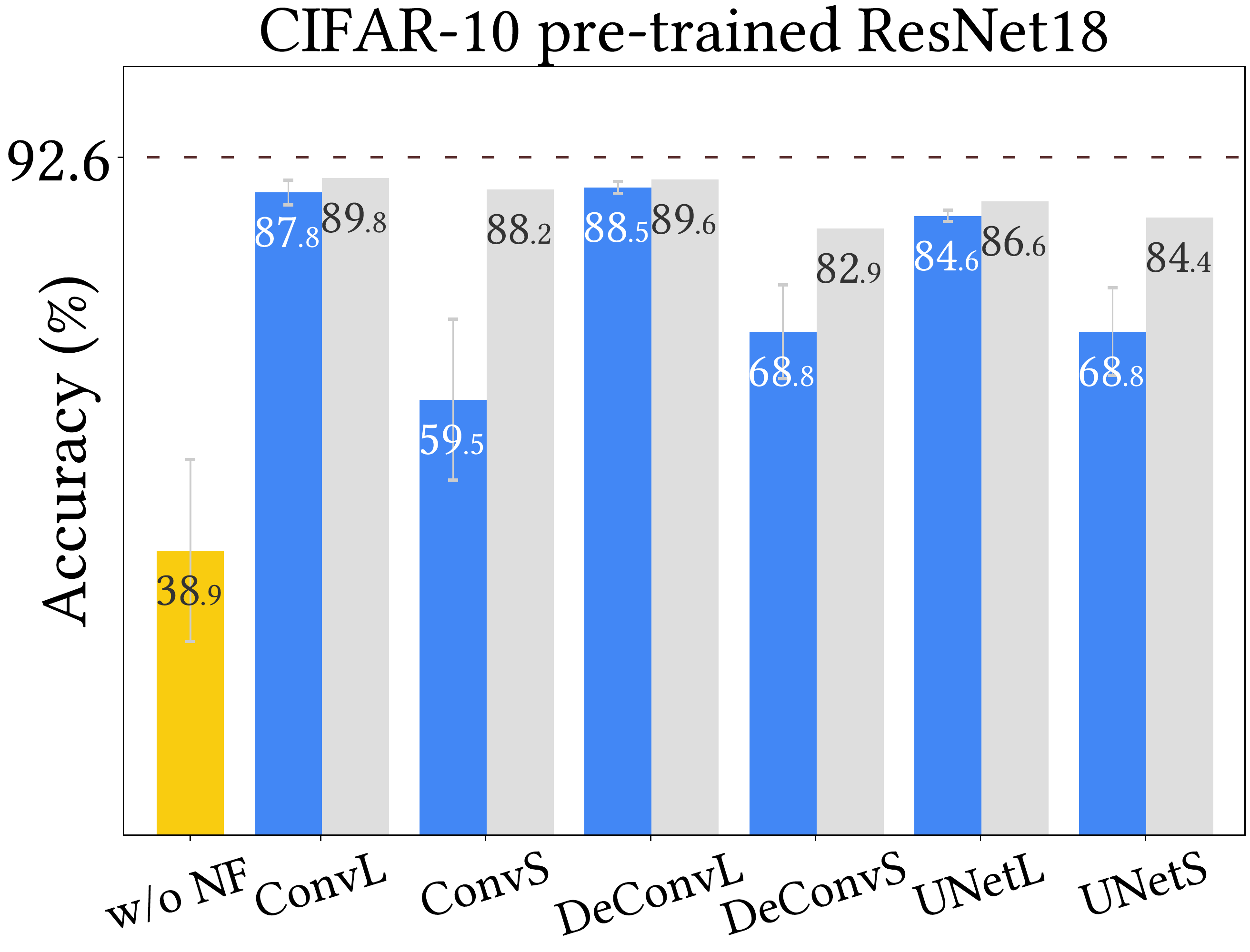}
     \end{subfigure}
     \hfill
     \begin{subfigure}[b]{0.195\textwidth}
         \centering
         \includegraphics[width=\textwidth]{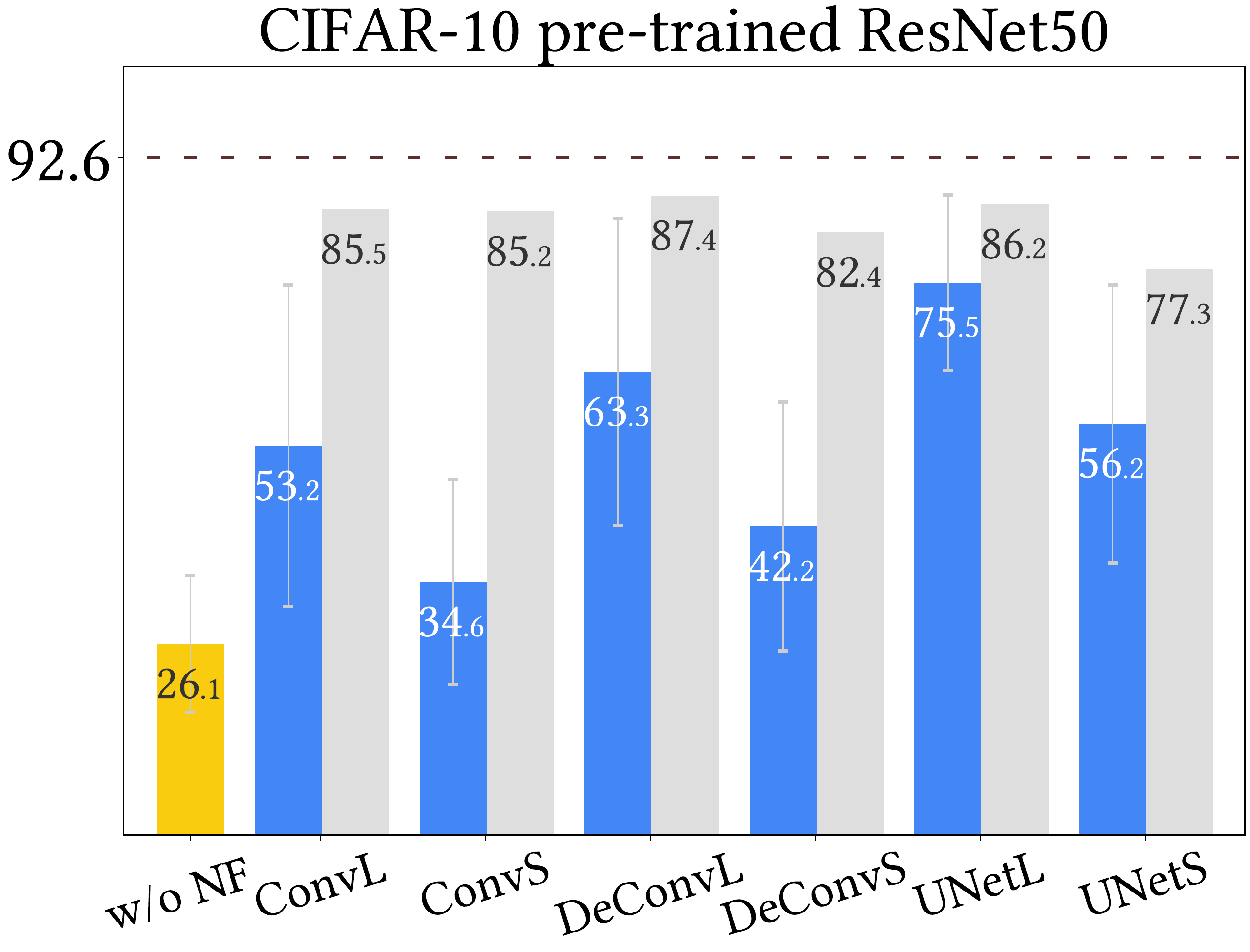}
     \end{subfigure}
     \hfill
     \begin{subfigure}[b]{0.195\textwidth}
         \centering
         \includegraphics[width=\textwidth]{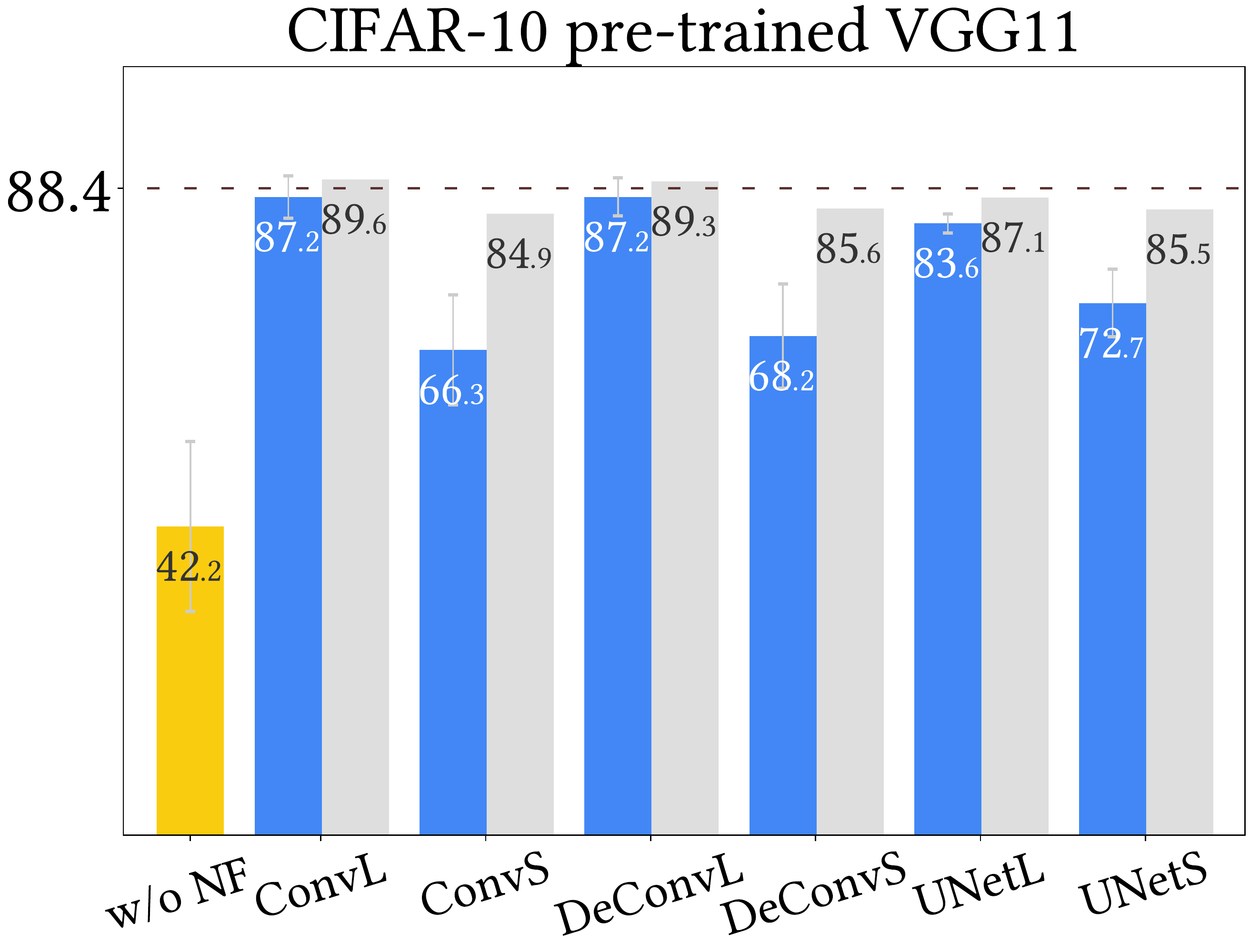}
     \end{subfigure}
     \hfill
     \begin{subfigure}[b]{0.195\textwidth}
         \centering
         \includegraphics[width=\textwidth]{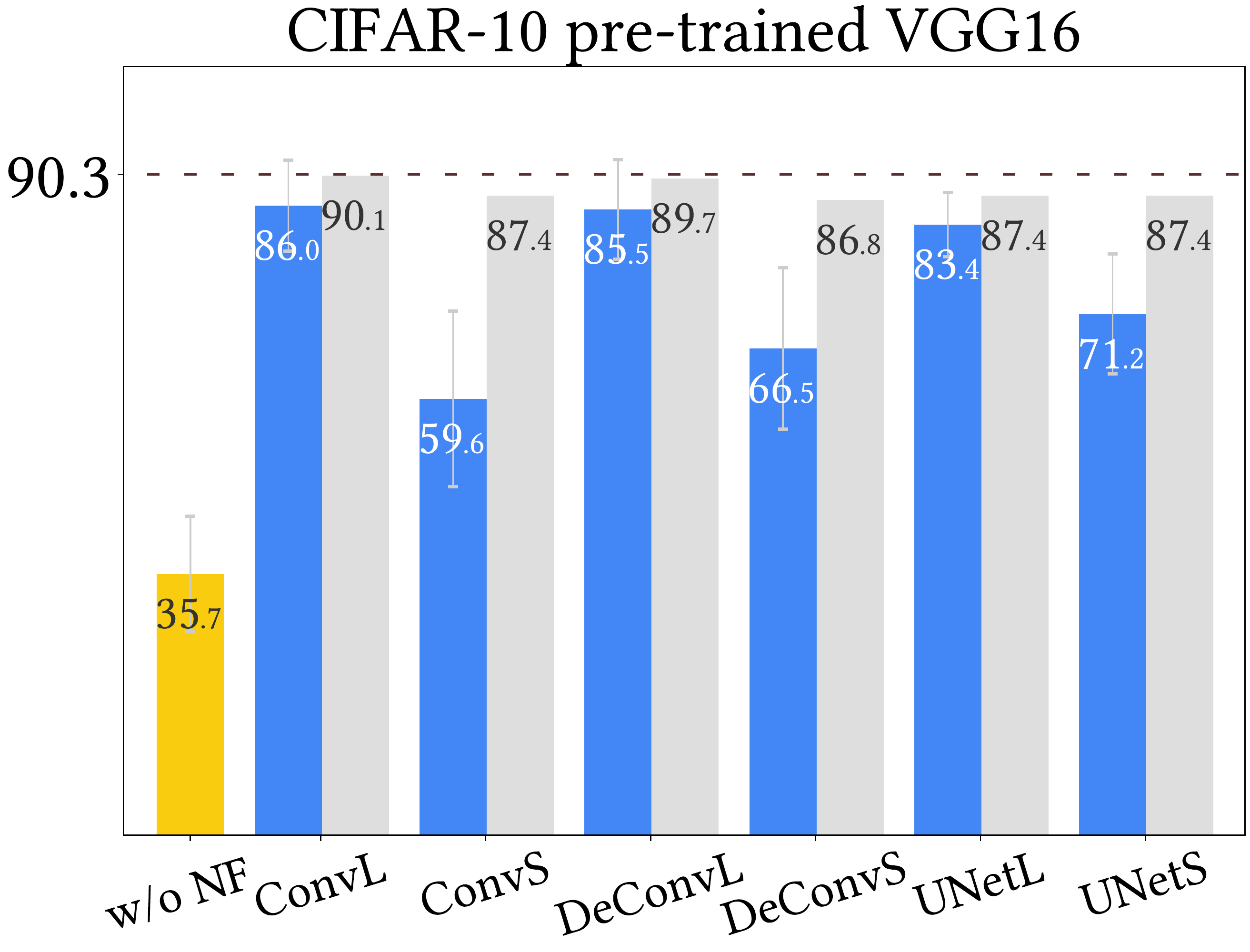}
     \end{subfigure}
     \hfill
     \begin{subfigure}[b]{0.195\textwidth}
         \centering
         \includegraphics[width=\textwidth]{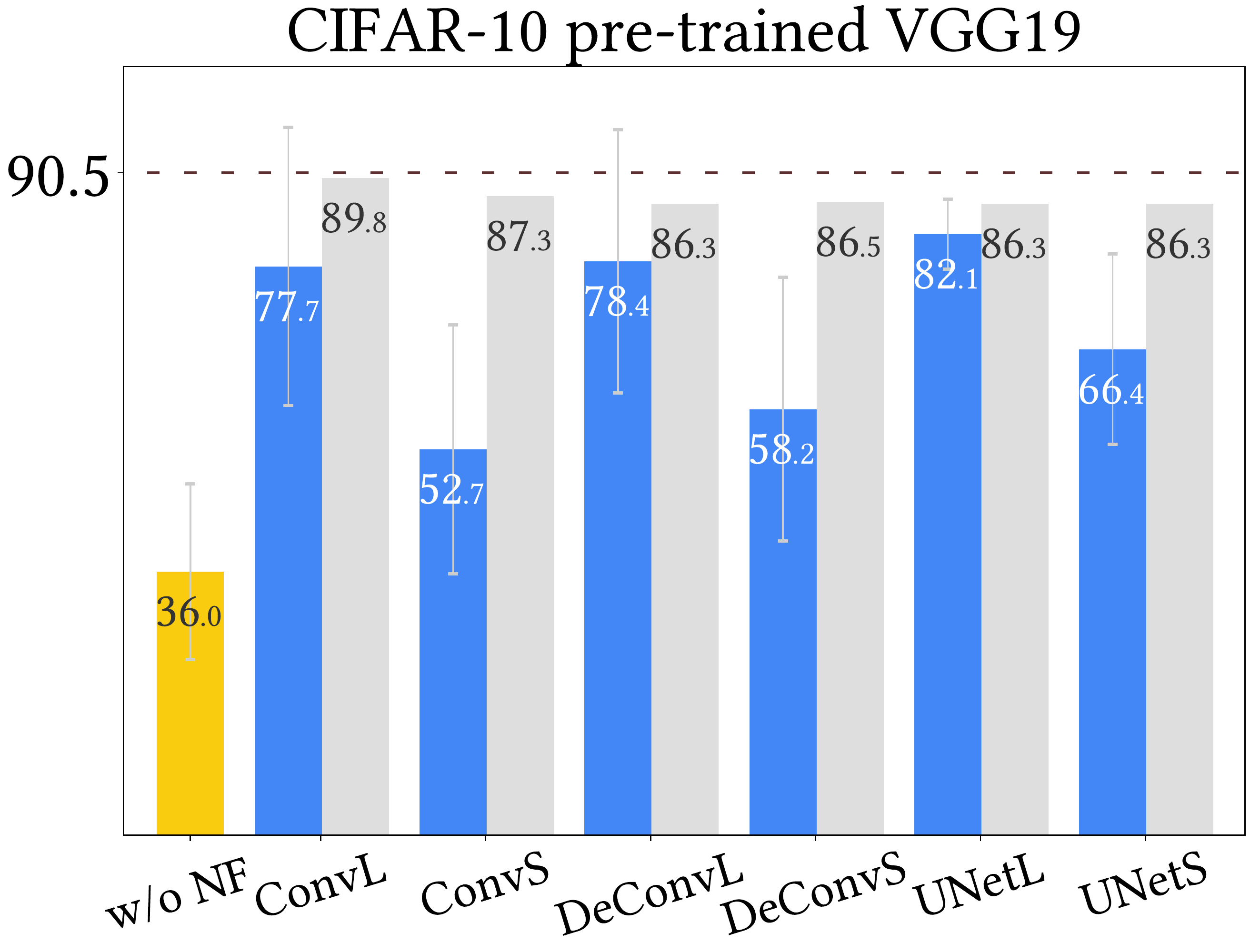}
     \end{subfigure}
     {\small (a) CIFAR-10, $1\%$ Bit-error Rate\\}\vspace{2mm}
     \begin{subfigure}[b]{0.195\textwidth}
         \centering
         \includegraphics[width=\textwidth]{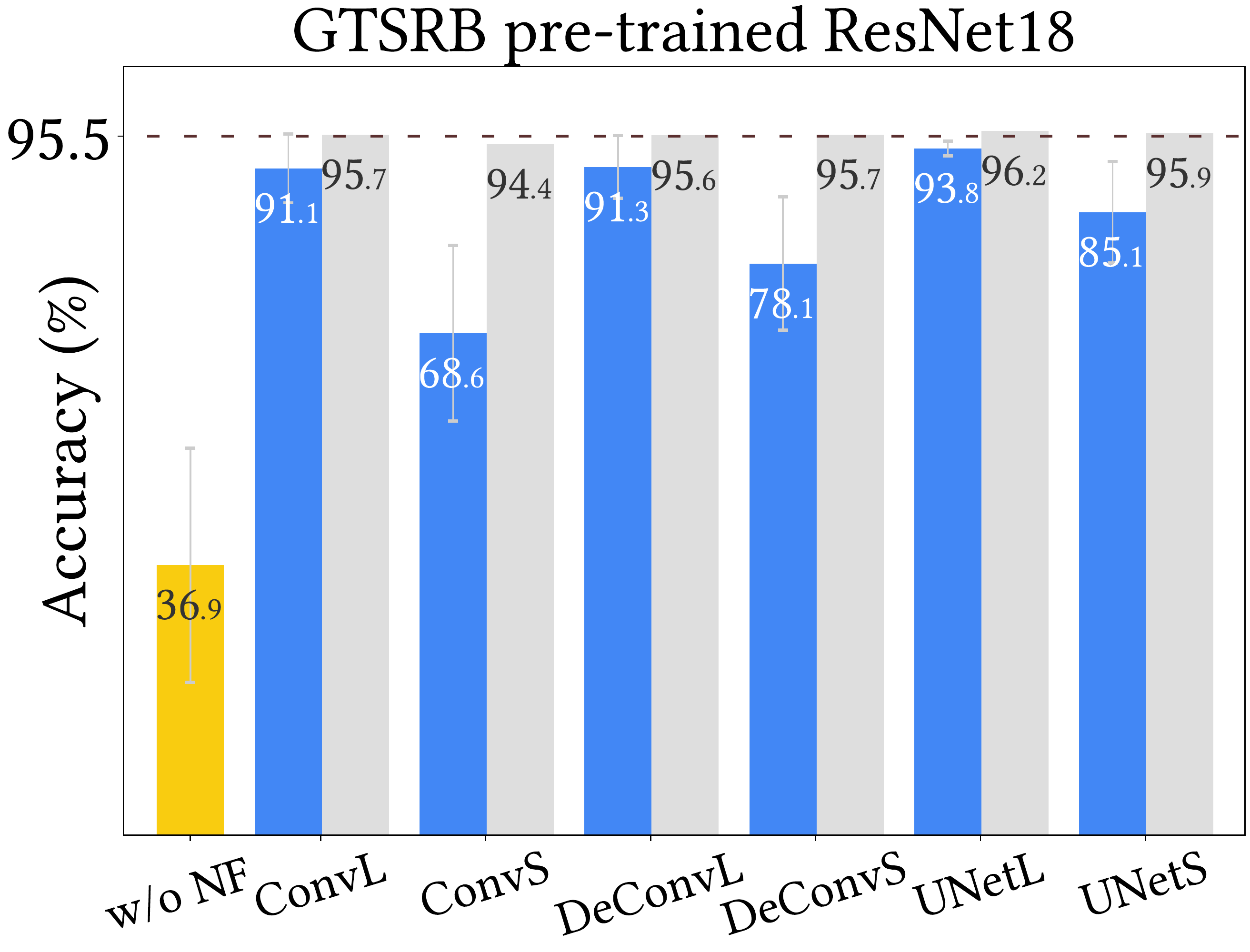}
     \end{subfigure}
     \hfill
     \begin{subfigure}[b]{0.195\textwidth}
         \centering
         \includegraphics[width=\textwidth]{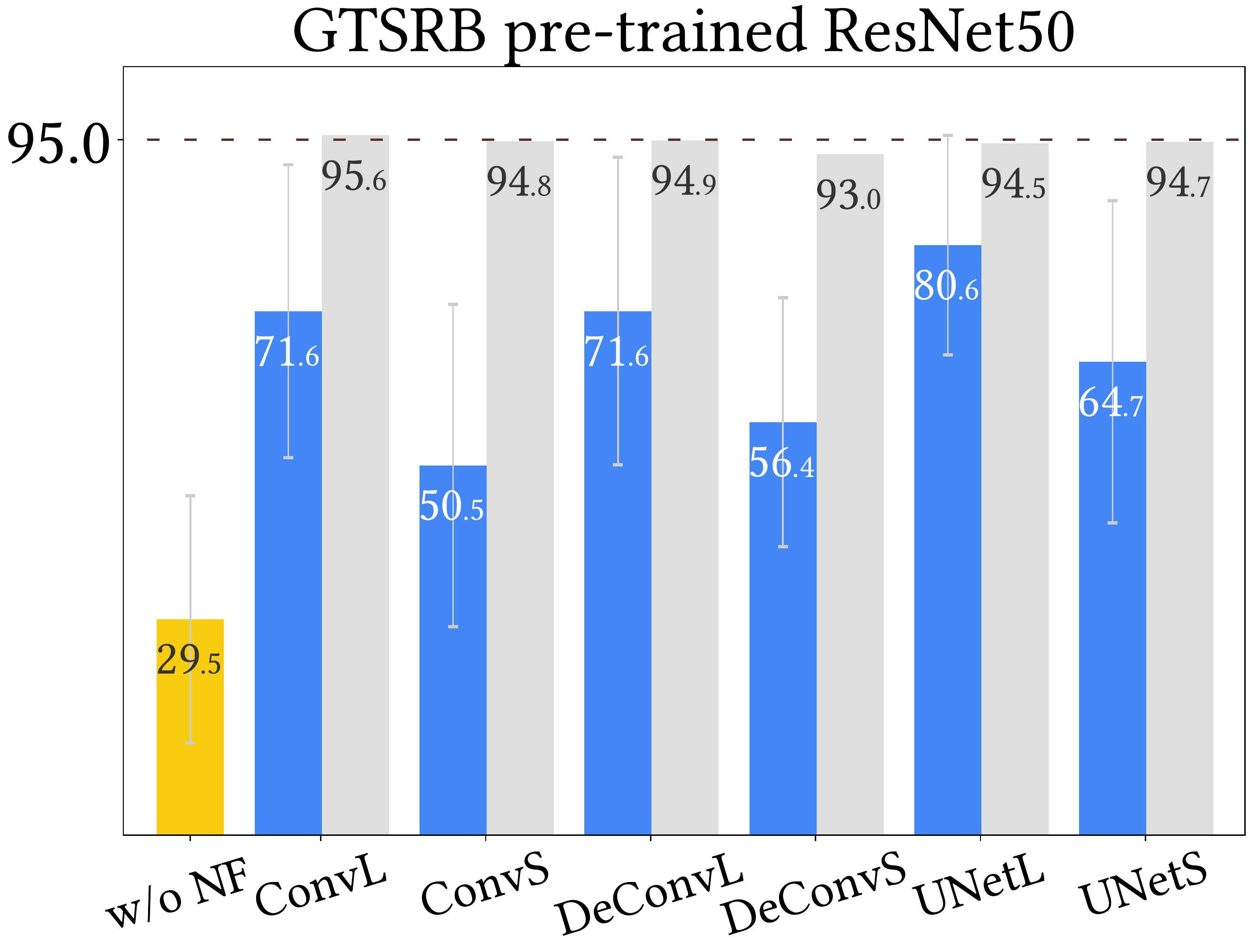}
     \end{subfigure}
     \hfill
     \begin{subfigure}[b]{0.195\textwidth}
         \centering
         \includegraphics[width=\textwidth]{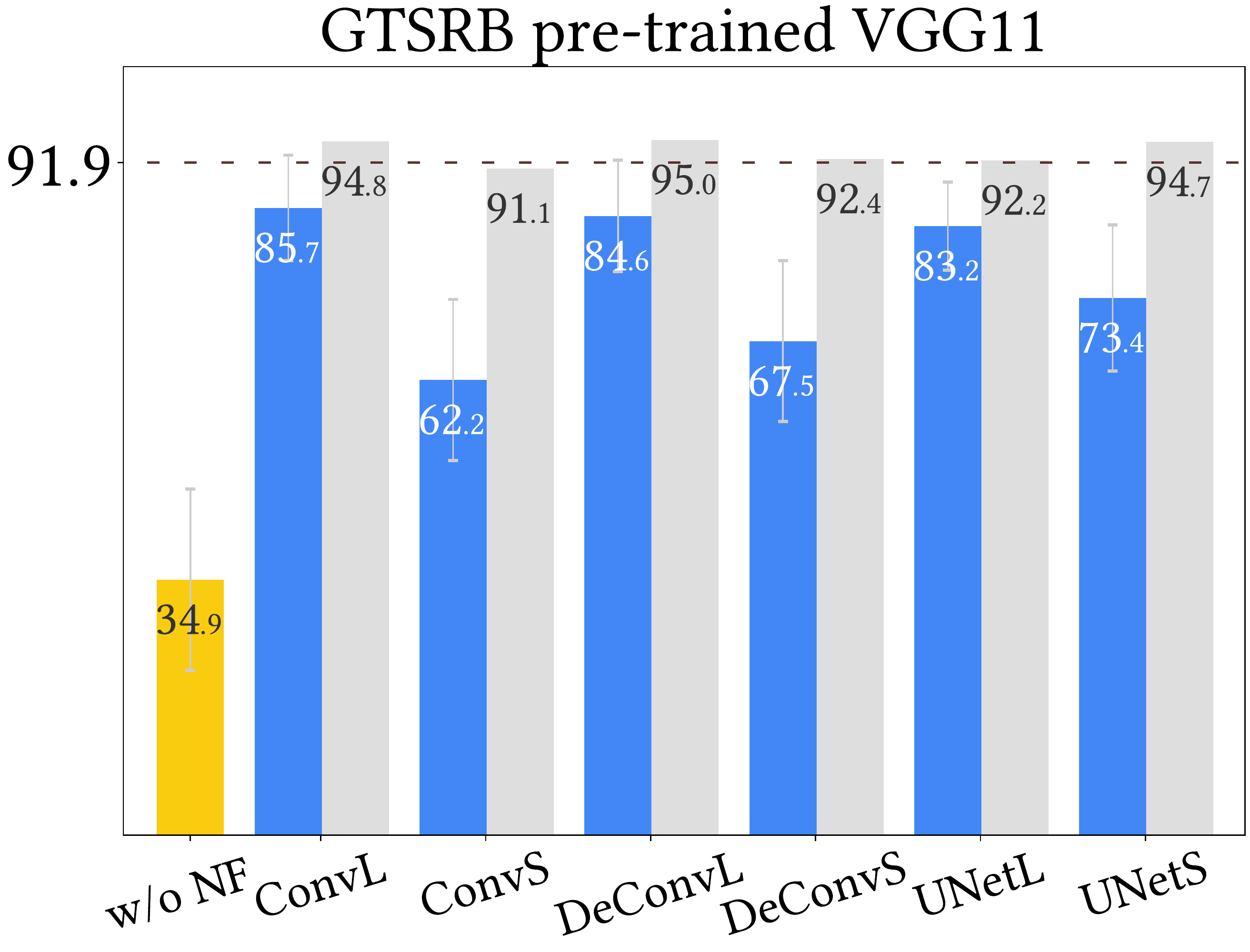}
     \end{subfigure}
     \hfill
     \begin{subfigure}[b]{0.195\textwidth}
         \centering
         \includegraphics[width=\textwidth]{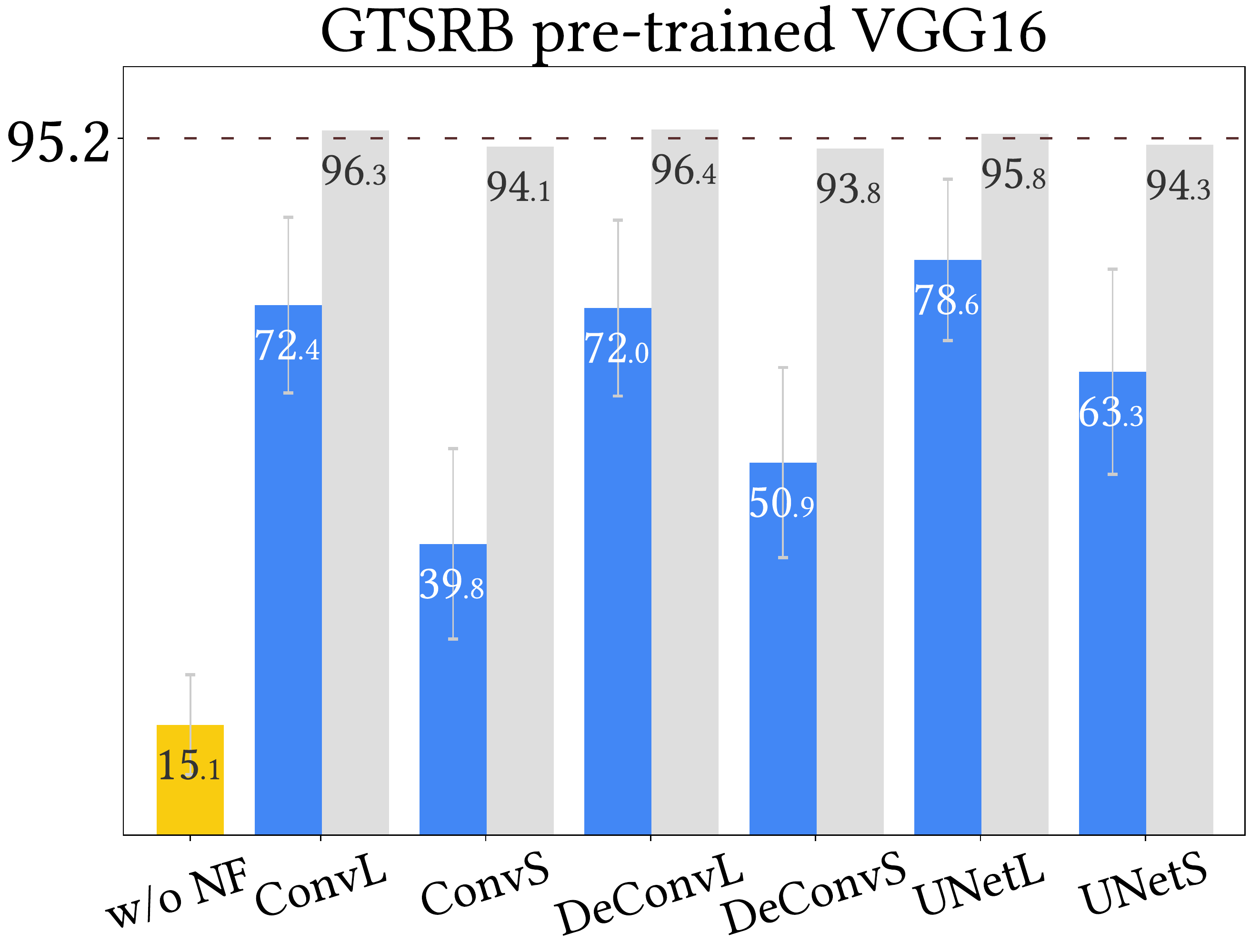}
     \end{subfigure}
     \hfill
     \begin{subfigure}[b]{0.195\textwidth}
         \centering
         \includegraphics[width=\textwidth]{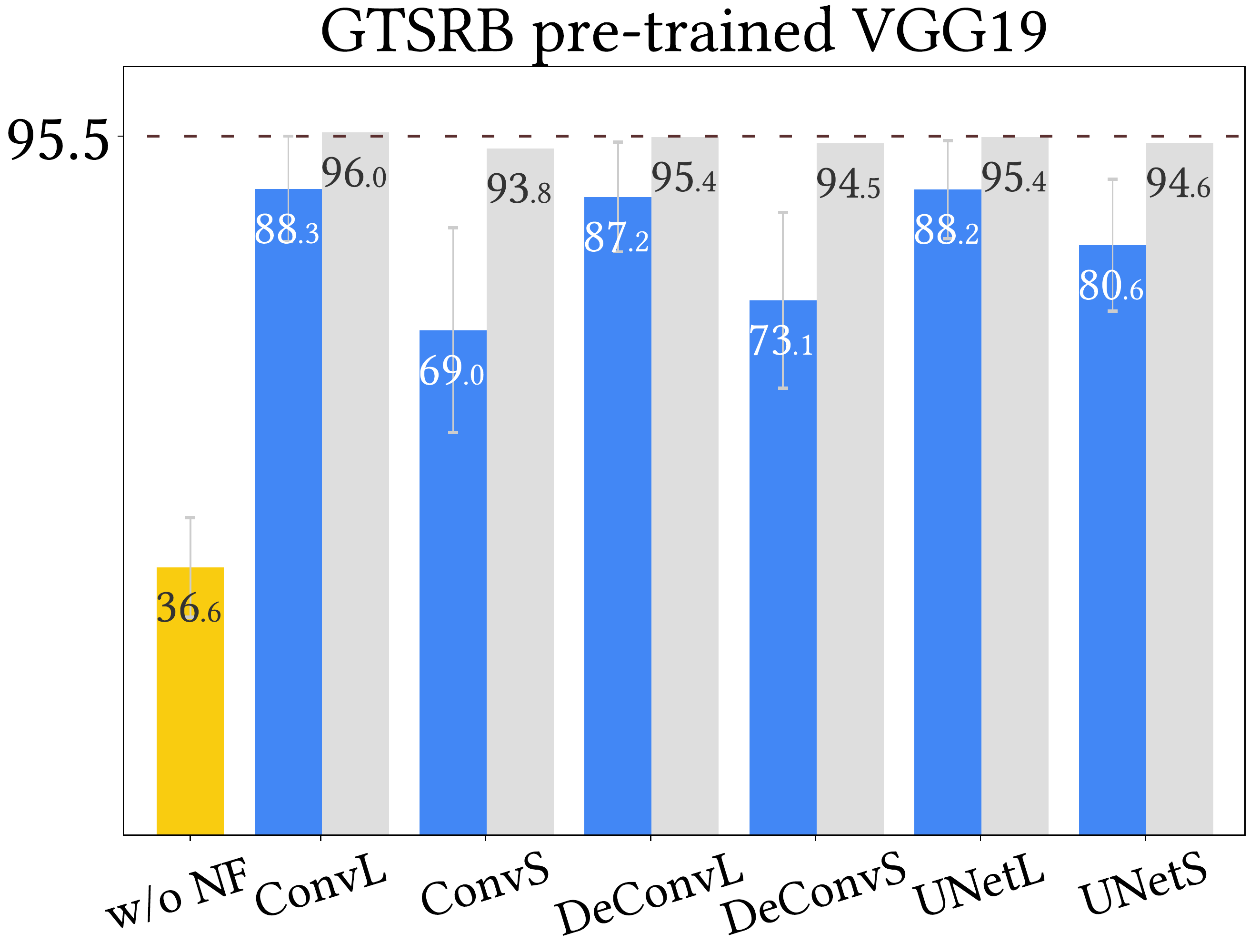}
     \end{subfigure}
     {\small (b) GTSRB, $1\%$ Bit-error Rate\\}\vspace{2mm}
     \begin{subfigure}[b]{0.195\textwidth}
         \centering
         \includegraphics[width=\textwidth]{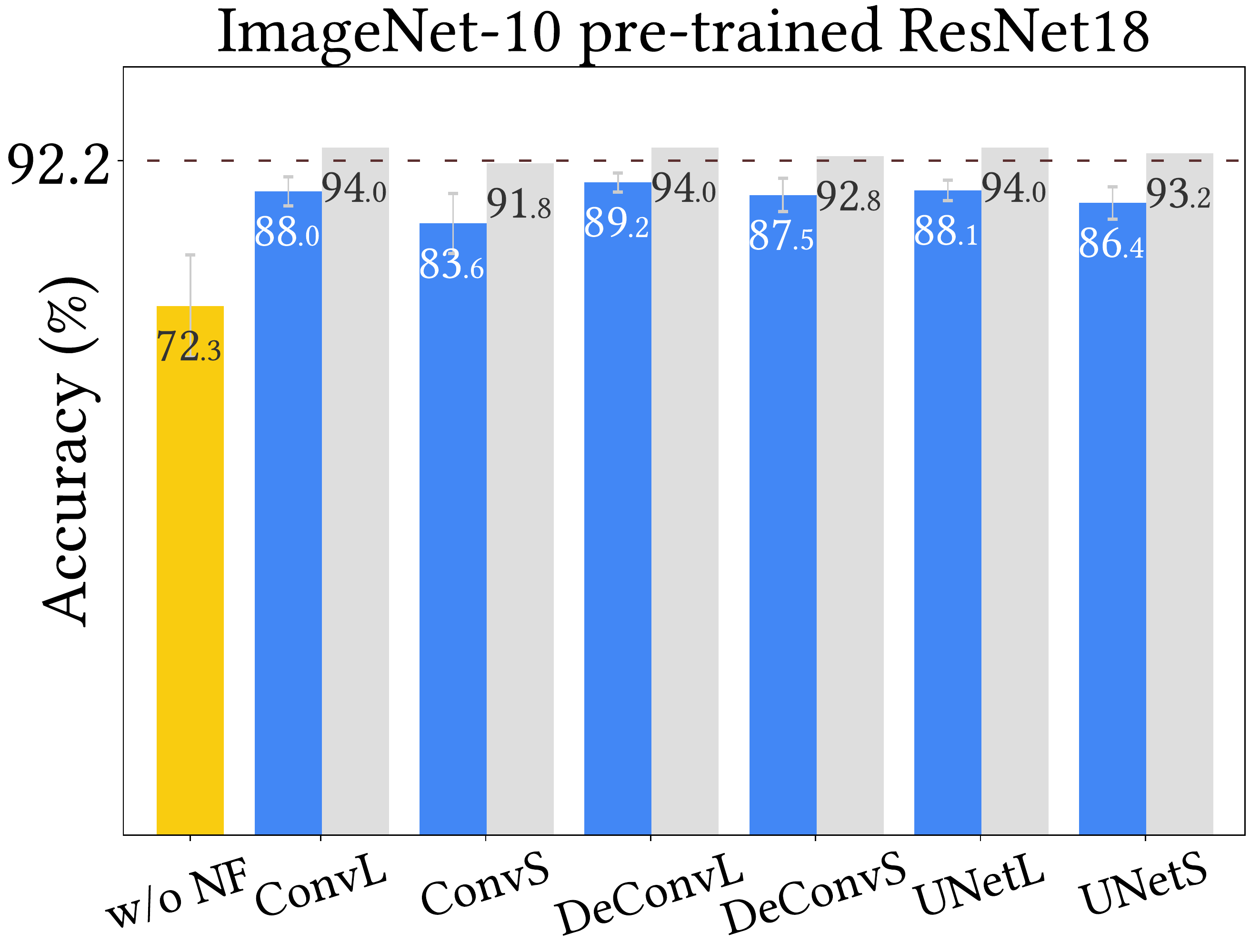}
     \end{subfigure}
     \hfill
     \begin{subfigure}[b]{0.195\textwidth}
         \centering
         \includegraphics[width=\textwidth]{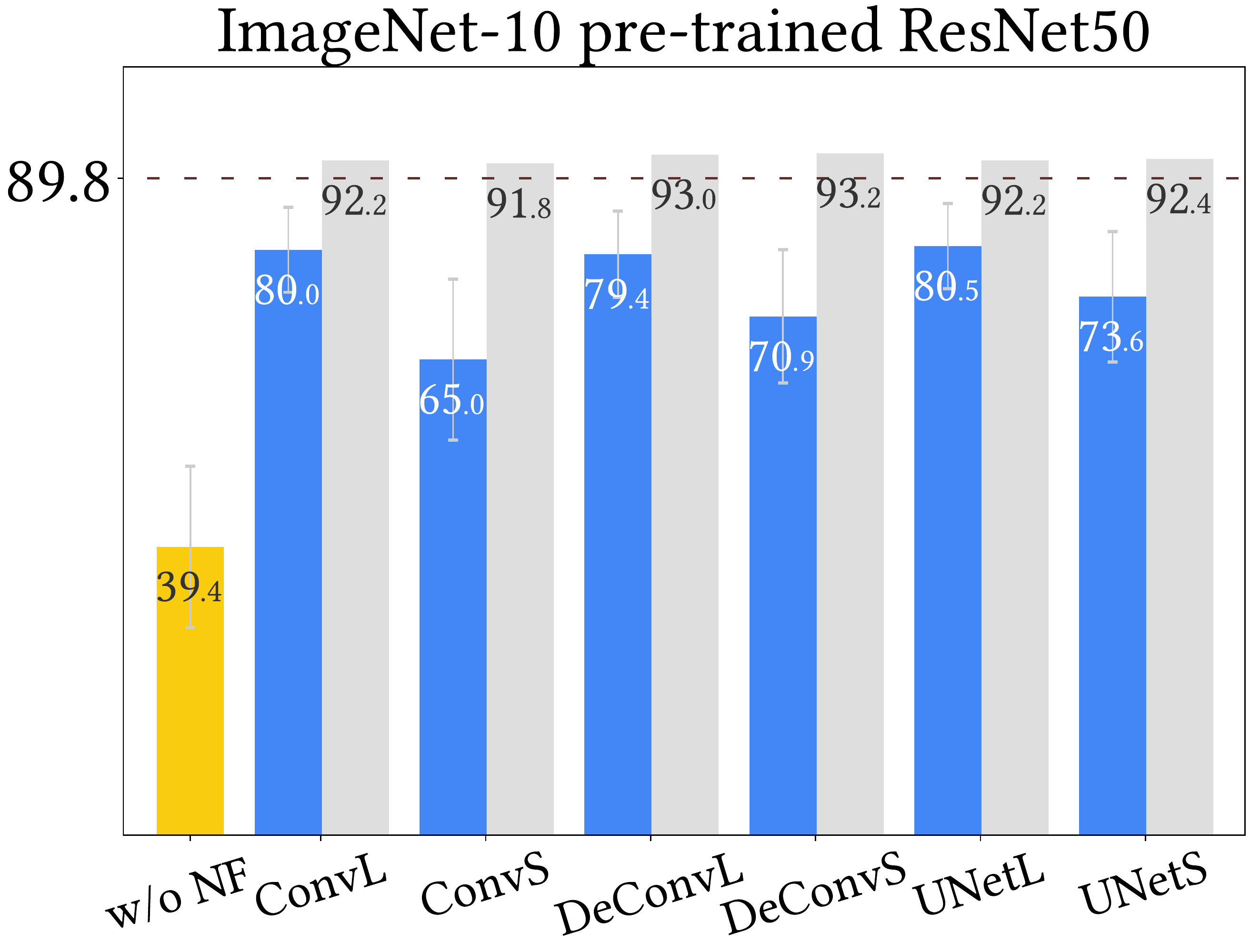}
     \end{subfigure}
     \hfill
     \begin{subfigure}[b]{0.195\textwidth}
         \centering
         \includegraphics[width=\textwidth]{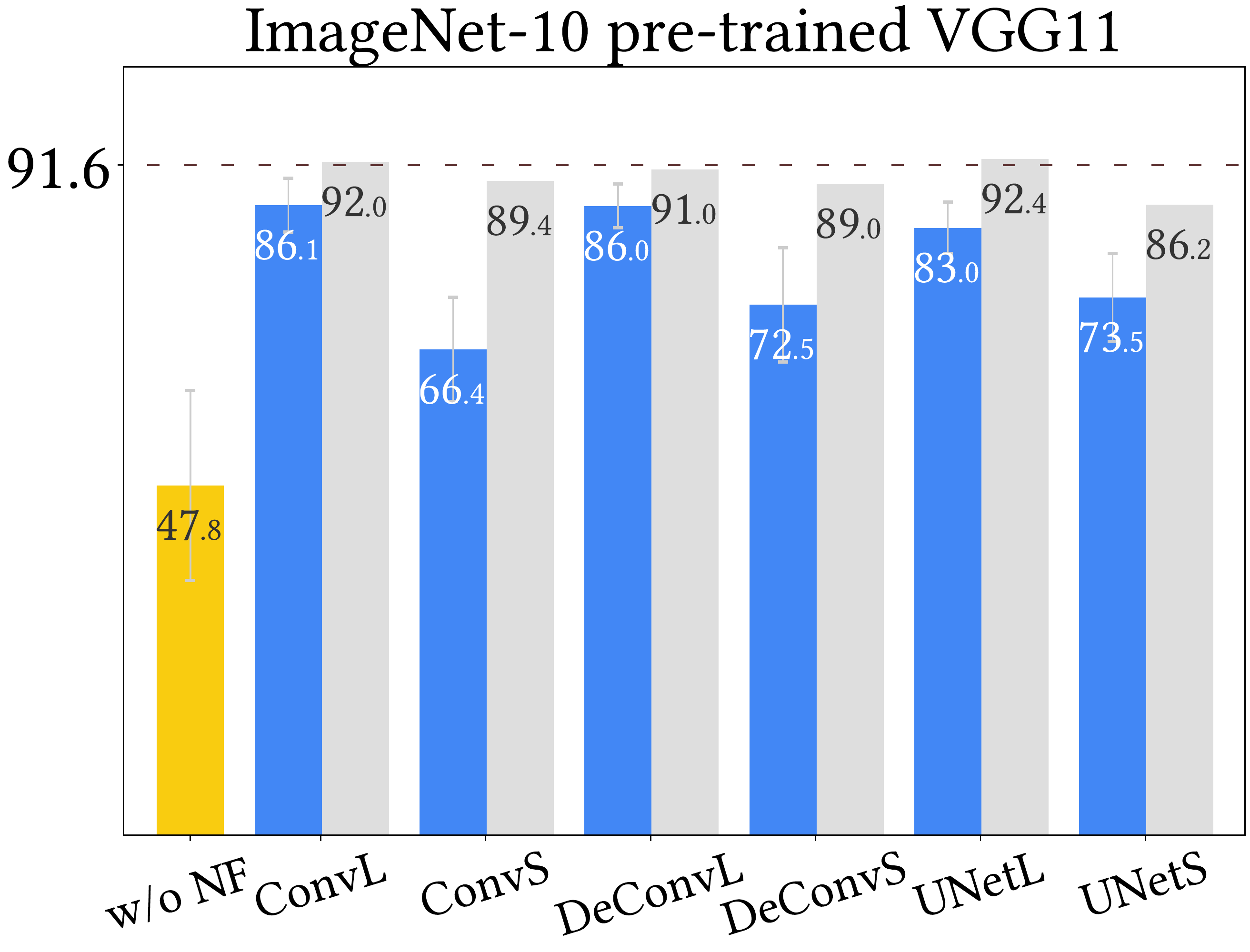}
     \end{subfigure}
     \hfill
     \begin{subfigure}[b]{0.195\textwidth}
         \centering
         \includegraphics[width=\textwidth]{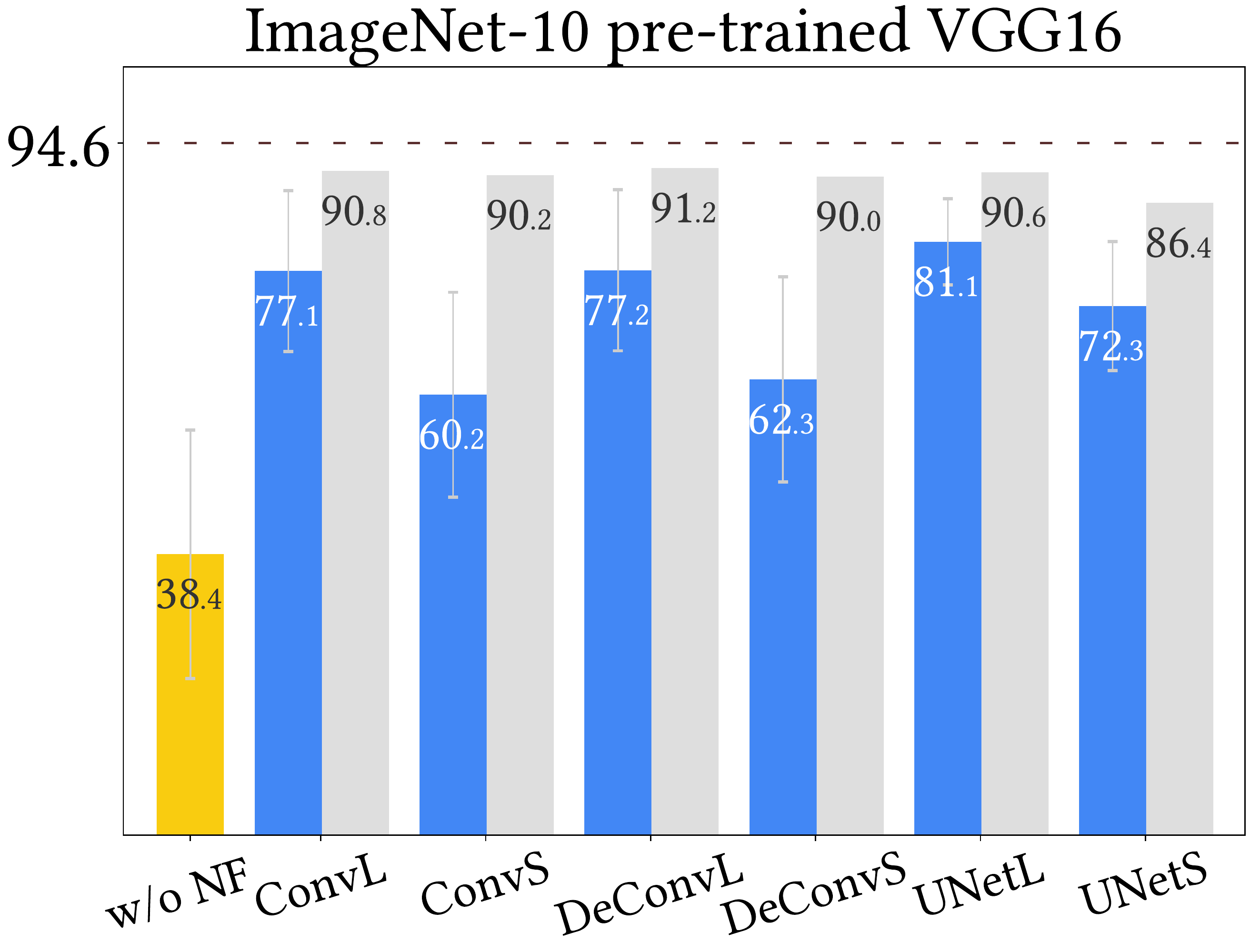}
     \end{subfigure}
     \hfill
     \begin{subfigure}[b]{0.195\textwidth}
         \centering
         \includegraphics[width=\textwidth]{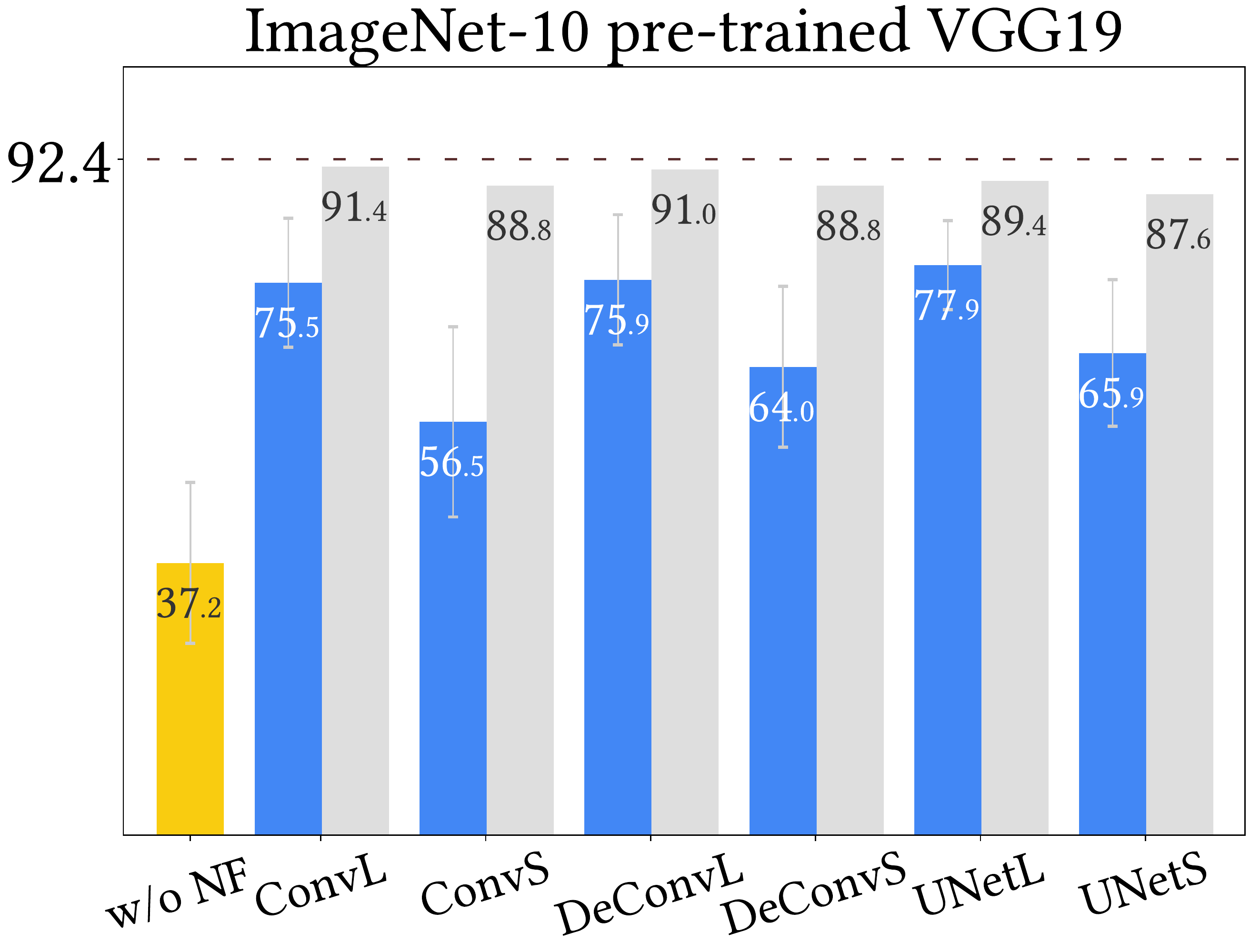}
     \end{subfigure}
     {\small (c) ImageNet-10, $0.5\%$ Bit-error Rate}
     \vspace{-1mm}
        \caption{Relaxed-access scenario test accuracies ($\%$) of various pre-trained models with and without NeuralFuse, compared at nominal voltage ($0\%$ bit-error rate) or low voltage (with specified bit-error rates). The results demonstrate that NeuralFuse consistently recovered perturbation accuracy.}
        \label{fig:experimental_results_main1}
\end{figure*}

%% file: assets/_tables/_tables_experiments_maintransferabilityp1.tex
\begin{table*}[t]
\centering
\caption{Restricted-access scenario: Transfer results on CIFAR-10 with $1.5\%$ bit-error rate}
\label{table:Maintransferabilityp1}
\begin{adjustbox}{max width=\linewidth}
\begin{threeparttable}
\begin{tabular}{c|c|c|cc|crr|crr }
\hline
\multirow{2}{*}{SM} & \multirow{2}{*}{TM} & \multirow{2}{*}{BER} & \multirow{2}{*}{CA} & \multirow{2}{*}{PA} & \multicolumn{3}{c|}{ConvL (1.5\%)} & \multicolumn{3}{c}{UNetL (1.5\%)} \\ 
& & & & & CA (NF) & PA (NF) & RP & CA (NF) & PA (NF) & RP \\
\hline

\multirow{10}{*}{ResNet18} & \multirow{2}{*}{ResNet18} & 1\% & \multirow{2}{*}{92.6} & 38.9 $\pm$ 12.4 & \multirow{2}{*}{89.8} & 89.0 $\pm$ 0.5 & 50.1 & \multirow{2}{*}{85.8} & 85.2 $\pm$ 0.5 & 46.3 \\
&  & 0.5\% & & 70.1 $\pm$ 11.6 &  & 89.6 $\pm$ 0.2 & 19.5 &  & 85.7 $\pm$ 0.2 & 15.6 \\
\cline{2-11}

& \multirow{2}{*}{ResNet50} & 1\% & \multirow{2}{*}{92.6} & 26.1 $\pm$ ~~9.4 & \multirow{2}{*}{89.2} & 36.1 $\pm$ ~18 & 10.0 & \multirow{2}{*}{84.4} & 38.9 $\pm$ ~16 & 12.8 \\
&  & 0.5\% & & 61.0 $\pm$ 10.3 & & 74.1 $\pm$ ~10 &  13.1 & & 72.7 $\pm$ 4.6 & 11.7\\ 
\cline{2-11}

& \multirow{2}{*}{VGG11} & 1\% & \multirow{2}{*}{88.4} & 42.2 $\pm$ 11.6 & \multirow{2}{*}{86.3} & 59.2 $\pm$ ~10 & 17.0 & \multirow{2}{*}{82.3} & 69.8 $\pm$ 7.5 & 27.6 \\
&  & 0.5\% & & 63.6 $\pm$ ~~9.3 & & 78.9 $\pm$ 4.9 &  15.3 & & 77.0 $\pm$ 4.0 & 13.4\\ 
\cline{2-11}

& \multirow{2}{*}{VGG16} & 1\% & \multirow{2}{*}{90.3} & 35.7 $\pm$ ~~7.9 & \multirow{2}{*}{89.4} & 62.2 $\pm$ ~18 & 26.5 & \multirow{2}{*}{84.7} & 68.9 $\pm$ ~14 & 33.2 \\
&  & 0.5\% & & 66.6 $\pm$ ~~8.1 & & 83.4 $\pm$ 5.5 &  16.8 &  & 80.5 $\pm$ 5.9 & 13.9\\ 
\cline{2-11}

& \multirow{2}{*}{VGG19} & 1\% & \multirow{2}{*}{90.5} & 36.0 $\pm$ 12.0 & \multirow{2}{*}{89.8} & 49.9 $\pm$ ~23 & 13.9 &  \multirow{2}{*}{85.0} & 55.1 $\pm$ ~17 & 19.1 \\
&  & 0.5\% & & 64.2 $\pm$ 12.4 & & 81.8 $\pm$ 8.5 & 17.6 & & 78.5 $\pm$ 6.8 & 14.3\\ 
\hline

\multirow{10}{*}{VGG19} & \multirow{2}{*}{ResNet18} & 1\% & \multirow{2}{*}{92.6} & 38.9 $\pm$ 12.4 & \multirow{2}{*}{88.9} & 62.6 $\pm$ ~13 & 23.7 & \multirow{2}{*}{85.0} & 72.3 $\pm$ ~11 & 33.4 \\
&  & 0.5\% & & 70.1 $\pm$ 11.6 & & 84.2 $\pm$ 7.2 & 14.1 & & 82.1 $\pm$ 2.2 & 12.0 \\
\cline{2-11}

& \multirow{2}{*}{ResNet50} & 1\% & \multirow{2}{*}{92.6} & 26.1 $\pm$ ~~9.4 & \multirow{2}{*}{88.8} & 37.9 $\pm$ ~18 & 11.8 & \multirow{2}{*}{85.2} & 46.7 $\pm$ ~17 & 20.6 \\
&  & 0.5\% & & 61.0 $\pm$ 10.3 & & 76.6 $\pm$ 7.8 & 15.6 & & 78.3 $\pm$ 3.7 & 17.3 \\
\cline{2-11}

& \multirow{2}{*}{VGG11} & 1\% & \multirow{2}{*}{88.4} & 42.2 $\pm$ 11.6 & \multirow{2}{*}{88.9} & 76.0 $\pm$ 6.1 & 33.8 & \multirow{2}{*}{85.5} & 81.9 $\pm$ 3.9 & 39.7 \\
&  & 0.5\% & & 63.6 $\pm$ ~~9.3 & & 85.9 $\pm$ 2.6 & 22.3 & & 84.8 $\pm$ 0.5 & 21.2 \\
\cline{2-11}

& \multirow{2}{*}{VGG16} & 1\% & \multirow{2}{*}{90.3} & 35.7 $\pm$ ~~7.9 & \multirow{2}{*}{89.0} & 76.5 $\pm$ 9.0 & 40.8 &  \multirow{2}{*}{85.9} & 79.2 $\pm$ 7.5 & 43.5 \\
&  & 0.5\% & & 66.6 $\pm$ ~~8.1 & & 87.7 $\pm$ 0.7 & 21.1 & & 84.7 $\pm$ 0.9 & 18.1 \\
\cline{2-11}

& \multirow{2}{*}{VGG19} & 1\% & \multirow{2}{*}{90.5} & 36.0 $\pm$ 12.0 & \multirow{2}{*}{89.1} & 80.2 $\pm$ ~12 & 44.2 & \multirow{2}{*}{86.3} & 84.3 $\pm$ 1.2 & 48.3 \\
&  & 0.5\% & & 64.2 $\pm$ 12.4 & & 88.8 $\pm$ 0.4 & 24.6 & & 85.9 $\pm$ 0.3 & 21.7 \\
\hline

\end{tabular}
    \begin{tablenotes}
      \small
      \item {\textit{Note}. SM: source model, used for training generators; TM: target model, used for testing generators; BER: the bit-error rate of the target model; CA ($\%$): clean accuracy; PA ($\%$): perturbed accuracy; NF: NeuralFuse; and RP: total recovery percentage of PA (NF) vs. PA}
    \end{tablenotes}
\end{threeparttable}
\end{adjustbox}
\vspace{-4mm}
\end{table*}

%% file: assets/_tables/_tables_Appendix_EnergySaving.tex
\begin{table}[t]
\setlength{\tabcolsep}{1mm}
\centering
\caption{Energy saving ($\%$) by NeuralFuse for 30 combinations of base models and generators.}\label{table:EnergySaving}
\vspace{1mm}
\begin{adjustbox}{max width=\linewidth}
\begin{tabular}{c|rrrrrr}
\hline
& ConvL & ConvS & DeConvL & DeConvS & UNetL & UNetS \\
\hline
ResNet18 & 19.0 & 29.1 & 21.2 & 27.5 & 24.0 & 28.9 \\
ResNet50 & 25.4 & 29.9 & 26.4 & 29.2 & 27.7 & 29.9 \\
VGG11 & 6.6 & 27.5 & 11.2 & 24.1 & 17.1 & 27.2 \\
VGG16 & 17.1 & 28.9 & 19.7 & 27.0 & 23.0 & 28.7 \\
VGG19 & 20.3 & 29.7 & 22.3 & 27.8 & 24.8 & 29.1 \\
\hline
\end{tabular}
\end{adjustbox}\vspace{-6mm}
\end{table}

%% file: assets/_figures_and_tables.tex
\begin{figure*}
  \begin{minipage}[c]{0.60\textwidth}
    \centering
      \captionof{table}{The efficiency ratio for all NeuralFuse generators.\label{table:EffRat}}
\setlength{\tabcolsep}{2pt}
\begin{adjustbox}{max width=\linewidth}
\begin{tabular}{ c|c|rrrrrr }
\hline

\multirow{2}{*}{\begin{tabular}{@{}c@{}}Base \\ Model\end{tabular}} & \multirow{2}{*}{BER} & \multicolumn{6}{c}{NeuralFuse} \\ 
& & ConvL & ConvS & DeConvL & DeConvS & UNetL & UNetS \\
\hline 

\multirow{2}{*}{ResNet18} & 1\% & 67.5 & 182 & 76.6 & 190.7 & 94.5 & 245.9 \\
 & 0.5\% & 24.7 & 73.3 & 30.7 & 62.5 & 33.6 & 88.3 \\
\hline

\multirow{2}{*}{ResNet50} & 1\% & 37.4 & 75.1 & 57.4 & 102.7 & 102.3 & 248.4 \\
 & 0.5\% & 35.2 & 108.7 & 40.4 & 92.5 & 47.4 & 124.6 \\
\hline

\multirow{2}{*}{VGG11} & 1\% & 62.3 & 212.9 & 69.5 & 165.8 & 92.0 & 251.7 \\
 & 0.5\% & 32.3 & 96.3 & 35.8 & 77.2 & 38.9 & 100.7 \\
\hline

\multirow{2}{*}{VGG16} & 1\% & 69.6 & 211.2 & 76.9 & 196.5 & 98.8 & 292.9 \\
 & 0.5\% & 30.3 & 98.1 & 33.0 & 75.3 & 40.6 & 113 \\
\hline

\multirow{2}{*}{VGG19} & 1\% & 57.6 & 147.5 & 65.5 & 141.6 & 95.4 & 250.8 \\
 & 0.5\% & 33.0 & 91.0 & 37.5 & 70.2 & 43.1 & 106.4 \\
\hline

\end{tabular}
\vspace{-2mm}
\end{adjustbox}
    \end{minipage}
  \hfill
  \begin{minipage}[c]{0.38\textwidth}
    \vspace{5mm}
    \centering
         \includegraphics[width=\textwidth, trim=-0.2cm -0.1cm 0.2cm 0.1cm, clip]{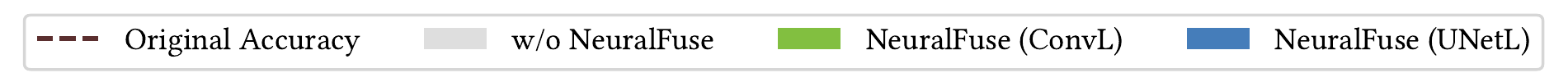}
         \includegraphics[width=\textwidth, trim=-0.2cm -0.1cm 0cm 0.1cm, clip]{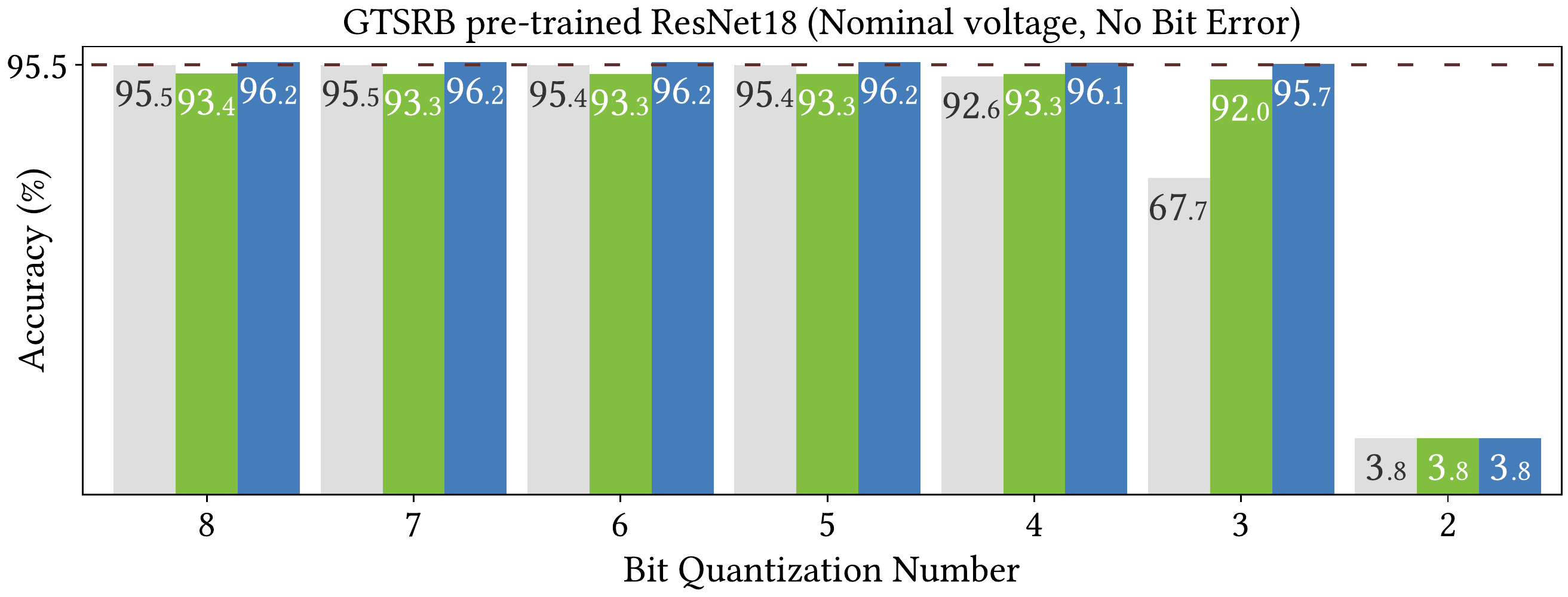}
         \includegraphics[width=\textwidth, trim=-0.35cm -0.1cm 0cm -0.1cm, clip]{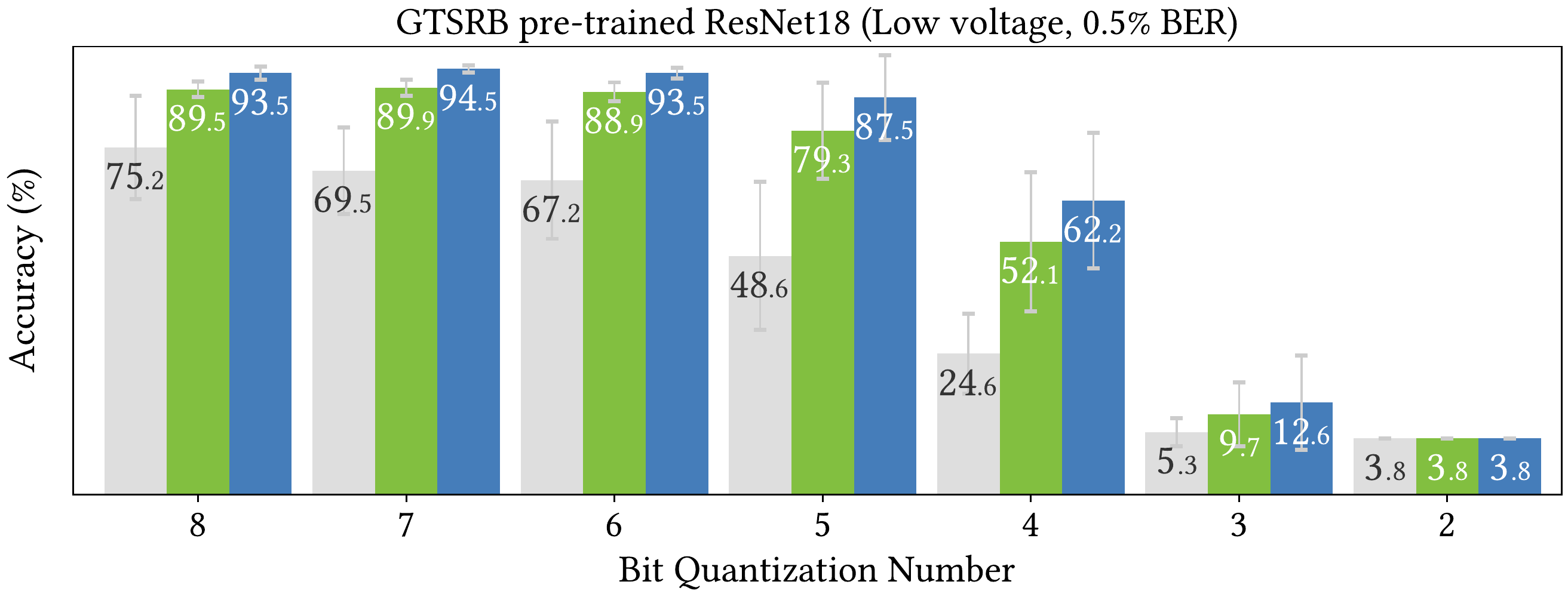}\vspace{-2mm}
    \captionof{figure}{Reduced-precision accuracy \label{fig:low_bit_precision}}
  \end{minipage}\vspace{-5mm}
\end{figure*}

%% file: sections/5_conclusions.tex
\section{Conclusion}\label{sec:conclusion}
In this paper, we have proposed NeuralFuse, the first non-intrusive post hoc module that protects model inference against bit errors induced by low voltage. NeuralFuse is particularly well suited to practical machine-deployment cases in which access to the base model is limited or relaxed. The design of NeuralFuse includes a novel loss function and a new optimizer, EOPM, that enable it to handle simulated randomness in perturbed models. Our comprehensive experimental results and analysis show that NeuralFuse can recover test accuracy by up to $57\%$ while simultaneously enjoying an up to $24\%$ reduction in memory-access energy requirements. Moreover, NeuralFuse demonstrates high transferability to access-constrained models and high versatility, e.g., robustness to low-precision quantization. In short, NeuralFuse is a notable advancement in mitigating neural-network inference’s energy/accuracy tradeoff in low-voltage regimes, and points the way to greener future AI technology. Our future work will include extending this study to other neural-network architectures and modalities, such as transformer-based language models.

\textbf{Limitations.}~ We acknowledge the challenge of achieving significant power savings without accuracy loss and view NeuralFuse as a foundational, proof-of-concept step toward this goal. Future research could enhance this approach by optimizing the pre-processing module to adapt to specific error characteristics of low-voltage SRAM or by integrating lightweight hardware modifications to further improve the energy-accuracy trade-off.

\textbf{Broader Impacts.}~ We see no ethical or immediate negative societal consequence of our work, and it holds the potential for positive social impacts, from environmental benefits to improved access to technology and enhanced safety in critical applications.

%% file: sections/6_supplementary.tex
\section*{Appendix}
The appendix provides more implementation details for our method, experimental results on more datasets and settings, ablation studies, and qualitative analysis. The appendices cover the following:

\begin{itemize}[leftmargin=*]
  \setlength{\itemsep}{0pt}%
    \item \textbf{Implementation Details:} NeuralFuse Training Algorithm (Sec. \ref{sec:appendix_training_algorithm}), NeuralFuse Generator (Sec. \ref{sec:appendix_NeuralFuse_implementation}), Energy/Accuracy Tradeoff (Sec. \ref{sec:appendix_energy_accuracy_tradeoff}), Latency Reports (Sec. \ref{sec:appendix_latency})
    \item \textbf{Experimental Results:} Ablation Studies (Sec. \ref{sec:appendix_ablation}), Relaxed Access (Sec. \ref{sec:appendix_relaxed_access}), Restricted Access (Sec. \ref{sec:appendix_transferability}), Reduced-precision Quantization (Sec. \ref{sec:appendix_low_precision_quantization}), Adversarial Training (Sec. \ref{sec:appendix_AdvTraining}), Adversarial Weight Perturbation (Sec. \ref{sec:appendix_robust_model_neuralfuse})
    \item \textbf{Qualitative Studies:} 
    Data Embeddings Visualization (Sec. \ref{sec:visualization}), Transformed Inputs Visualization (Sec. \ref{sec:visual_transformed})
\end{itemize}

\section{Training Algorithm of NeuralFuse} \label{sec:appendix_training_algorithm}

\input{assets/TrainingAlgo.tex}

\section{Implementation Details of NeuralFuse Generator}\label{sec:appendix_NeuralFuse_implementation}
We consider two main goals in designing the NeuralFuse Generator: 1) efficiency (so the overall energy overhead is decreased) and 2) robustness (so that it can generate robust patterns on the input image and overcome the random bit flipping in subsequent models). Accordingly, we choose to utilize an encode-decoder architecture in implementing the generator. The design of ConvL is inspired by \citet{InputAware}, in which the authors utilize a similar architecture to design an input-aware trigger generator, and have demonstrated its efficiency and effectiveness. Furthermore, we attempted to enhance it by replacing the \textit{Upsampling} layer with a \textit{Deconvolution} layer, leading to the creation of DeConvL. The UNetL-based NeuralFuse draws inspiration from \citet{UNet}, known for its robust performance in image segmentation, and thus, we incorporated it as one of our architectures. Lastly, ConvS, DeConvS, and UNetS are \textit{scaled-down} versions of the model designed to reduce computational costs and total parameters. The architectures of Convolutional-based and Deconvolutional-based are shown in Table \ref{table:GeneratorArchConDeCon}, and the architecture of UNet-based generators is in Table \ref{table:GeneratorArchUNet}.
For the abbreviation used in the table, ConvBlock means the Convolution block, Conv means a single Convolution layer, DeConvBlock means the Deconvolution block, DeConv means a single Deconvolution layer, and BN means a Batch Normalization layer. We use $\text{learning rate}=0.001$, $\lambda=5$, and Adam optimizer. For CIFAR-10, GTSRB, and CIFAR-100, we set batch size $b=25$ for each base model. For ImageNet-10, we set $b=64$ for ResNet18, ResNet50 and VGG11, and $b=32$ for both VGG16 and VGG19.

\input{assets/_tables/_tables_Appendix_GeneratorArchConDeCon}
\input{assets/_tables/_tables_Appendix_GeneratorArchUNet}
\newpage
\section{NeuralFuse's Energy/Accuracy Tradeoff}\label{sec:appendix_energy_accuracy_tradeoff}
\paragraph{SCALE-SIM.}
SCALE-SIM \citep{ScaleSim} is a systolic array-based CNN simulator that can calculate the number of memory accesses and the total time in execution cycles by giving the specific model architecture and accelerator architectural configuration as inputs. In this paper, we use SCALE-SIM to calculate the weights memory access of 5 based models (ResNet18, ResNet50, VGG11, VGG16, VGG19), and 6 generators (ConvL, ConvS, DeConvL, DeConvS, UNetL, UNetS). While SCALE-SIM supports both Convolutional and Linear layers, it does not yet support Deconvolution layers. Instead, we try to approximate the memory costs of Deconvolution layers by Convolution layers. We change the input and output from Deconvolution into the output and input of the Convolution layers. Besides, we also change the stride into 1 when we approximate it. We also add padding for the convolution layers while generating input files for SCALE-SIM. In this paper, we only consider the energy saving on weights accesses, so we only take the value ``SRAM Filter Reads'' from the output of SCALE-SIM as the \textit{total weights memory accesses} (T.W.M.A.) for further energy calculation.

In Table \ref{table:Total_Mem_E}, we report the total weight memory access (T.W.M.A.) using SCALE-SIM. We then showed the energy/accuracy tradeoff between all of the combinations of NeuralFuse and base models under a 1\% of bit error rate in Figure \ref{fig:WholeModel}.

\input{assets/_tables/_tables_Appendix_Total_Mem_E}
\input{assets/_figures/tradeoffWholeModels.tex}

\paragraph{Parameters and MACs Calculation.} In addition to T.W.M.A., the number of parameters and MACs (multiply–accumulate operations) are also common metrics in measuring the energy consumption of machine learning models. \citet{EnergyAwarePuring} have shown that the \textit{energy consumption of computation} and \textit{memory accesses} are both proportional to MACs, allowing us to take computation energy consumption into account.

Here, we use the open-source package \texttt{ptflops} \citep{ptflops} to calculate the parameters and MAC values of all the base models and the NeuralFuse generators, in the same units as \citet{QUALCOMM-Batch-shaping} used. The results are shown in Table \ref{table:Param_MACs}. Note that we modified the base model architectures for ImageNet-10, as it has larger input sizes. For example, we use a larger $\text{kernel size}=7$ instead of $3$ in the first Convolution layer in ResNet-based models to enhance the learning abilities. Therefore, the parameters of base models are different between different datasets. For NeuralFuse generators, we utilize the same architectures for implementation (including ImageNet-10). As a result, our proposed NeuralFuse generators are generally smaller than base models, either on total model parameters or MAC values.
\input{assets/_tables/_tables_Appendix_Param_MACs}

\paragraph{MACs-based Energy Consumption.}
We can then use the MAC values to further approximate the end-to-end energy consumption of the whole model. Assume that all values are stored on SRAM and that a MAC represents single memory access. The corresponding MACs-based energy saving percentage (MAC-ES, \%) can be derived from Eq. \ref{eqn:mac-es} (c.f. Sec. \ref{subsec:energy_acc_tradeoff}), and results can be found in Table \ref{table:EnergySavingMACs}. We can observe that most combinations can save a large amount of energy, except that VGG11 with two larger NeuralFuse (ConvL and DeConvL) may increase the total energy. These results are consistent with the results reported in Table \ref{table:EnergySaving}. In addition, we also showed the MACs-based energy/accuracy tradeoff between all of the combinations of NeuralFuse and base models under a 1\% of bit error rate in Figure \ref{fig:WholeModelMACsBased}.
\begin{equation}
\scalebox{0.9}{
    $\text{MAC-ES} = \frac{\text{MACs}_{\text{base model}} \cdot \text{Energy}_{\text{nominal voltage}} - \big(\text{MACs}_{\text{base model}} \cdot \text{Energy}_{\text{low-voltage-regime}} + \text{MACs}_{\text{NeuralFuse}} \cdot \text{Energy}_{\text{NeuralFuse at nominal voltage}}\big)}{\text{MACs}_{\text{base model}} \cdot \text{Energy}_{\text{nominal voltage}}}\times{100\%}\\
    {~~~}$}
    \label{eqn:mac-es}
\end{equation}

\input{assets/_tables/_tables_Appendix_EnergySavingBasedOnMACs.tex}
\input{assets/_figures/tradeoffWholeModelsMACs.tex}

Although using ConvL or DeConvL along with base model VGG11 for CIFAR-10 implies an increase in energy consumption, using other smaller-scale generators, we can still save the overall energy and recover the base model’s accuracy. That said, developers can always choose smaller generators (with orders of magnitude fewer MAC operations than the original network) to restore model accuracy, further demonstrating the flexibility of choosing NeuralFuse generators of different sizes.

\section{Inference Latency of NeuralFuse}\label{sec:appendix_latency}
In Table \ref{table:latency}, we report the latency (batch\_size=1, CIFAR-10/ImageNet-10 testing dataset) of utilizing the different NeuralFuse generators with two different base models, ResNet18 and VGG19. 
While NeuralFuse contributes some additional latency, we consider this an unavoidable tradeoff necessary to achieve reduced energy consumption within our framework. Although the primary focus of this paper is not on latency, we acknowledge its importance. Future research could explore designing a more lightweight version of the NeuralFuse module or applying model compression techniques to minimize latency. Additionally, we recognize that running NeuralFuse on a general-purpose CPU could lead to different latency and energy consumption figures due to various influencing factors like CPU architecture and manufacturing processes.

\input{assets/_tables/_tables_Appendix_AppendixLatencyNeuralFuse}

\section{Ablation Studies}\label{sec:appendix_ablation}
\paragraph{Study for $N$ in EOPM.} Here, we study the effect of $N$ used in EOPM (Eq. \ref{eqn:gradient_calculation}). In Figure \ref{fig:TestN}, we report the results for ConvL and ConvS on CIFAR-10 pre-trained ResNet18, under a 1\% bit error rate (BER). The results demonstrate that if we apply larger $N$, the performance increases until convergence. Specifically, for ConvL (Figure \ref{fig:TestN_ConvL}), larger $N$ empirically has a smaller standard deviation; this means larger $N$ gives better stability but at the cost of time-consuming training. In contrast, for the small generator ConvS (Figure \ref{fig:TestN_ConvS}), we can find that the standard deviation is still large even trained by larger $N$; the reason might be that small generators are not as capable of learning representations as larger ones. Therefore, there exists a tradeoff between the \textit{stability} of the generator performance and the total training time. In our implementation, choosing $N=5$ or $10$ is a good balance.
\input{assets/_figures/error_bar.tex}

\paragraph{Tradeoff Between Clean Accuracy (CA) and Perturbed Accuracy (PA).} We conducted an experiment to study the effect of different $\lambda$ values, which balance the ratio of clean accuracy and perturbed accuracy. In Table \ref{table:AppendixLambdaCIFAR10}, the experimental results showed that a smaller $\lambda$ can preserve clean accuracy, but result in poor perturbed accuracy. On the contrary, larger $\lambda$ can deliver higher perturbed accuracy, but with more clean accuracy drop. This phenomenon has also been observed in adversarial training \citep{RobustnessAccTradeoff}.
\input{assets/_tables/_tables_Appendix_AppendixlambdaCifar10.tex}

\paragraph{Comparison to Universal Input Perturbation (\textbf{UIP}).} \citet{universal} has shown that there exists a universal adversarial perturbation to the input data such that the model will make wrong predictions on a majority of the perturbed images. 
In our NeuralFuse framework, the universal perturbation is a special case when we set $\mathcal{G}(\mathbf{x}) = \tanh{(\text{\textbf{UIP}})}$ for any data sample $\mathbf{x}$. The transformed data sample then becomes $\mathbf{x}_{t}=\text{clip}_{[-1,1]}\big(\mathbf{x}+\tanh{(\text{\textbf{UIP}})}\big)$, where $\mathbf{x}_{t} \in [-1,1]^d$ and \textbf{UIP} is a trainable universal input perturbation that has the same size as the input data. The experimental results with the universal input perturbation are shown in Table \ref{table:Mainuniversalp1CIFAR10}. We observe that its performance is much worse than our proposed NeuralFuse. This result validates the necessity of adopting input-aware transformation for learning error-resistant data representations in low-voltage scenarios.

\input{assets/_tables/_tables_experiments_universal}
\clearpage

\section{Additional Experimental Results on Relaxed Access Settings}\label{sec:appendix_relaxed_access}

We conducted more experiments on \textit{Relaxed Access} settings to show that our NeuralFuse can protect the models under different BER The results can be found in Sec. \ref{sec:appendix_relaxed_access_cifar10} (CIFAR-10), Sec. \ref{sec:appendix_relaxed_access_gtsrb} (GTSRB), Sec. \ref{sec:appendix_relaxed_access_imagenet10} (ImageNet-10), and Sec. \ref{sec:appendix_relaxed_access_cifar100}  (CIFAR-100). For comparison, we also visualize the experimental results in the figures below each table.

\subsection{CIFAR-10}\label{sec:appendix_relaxed_access_cifar10}
\input{assets/_tables/relaxed_access/_tables_Appendix_CIFAR10}
\input{assets/_figures/appendix_relaxed_access/cifar10}
\clearpage
\subsection{GTSRB}\label{sec:appendix_relaxed_access_gtsrb}
\input{assets/_tables/relaxed_access/_tables_Appendix_GTSRB}
\input{assets/_figures/appendix_relaxed_access/gtsrb}
\clearpage
\subsection{ImageNet-10}\label{sec:appendix_relaxed_access_imagenet10}
\input{assets/_tables/relaxed_access/_tables_Appendix_ImageNet10}
\input{assets/_figures/appendix_relaxed_access/imagenet10}
\clearpage
\subsection{CIFAR-100}\label{sec:appendix_relaxed_access_cifar100}
As mentioned in the previous section, larger generators like ConvL, DeConvL, and UNetL have better performance than small generators. For CIFAR-100, we find that the gains of utilizing NeuralFuse are less compared to the other datasets. We believe this is because CIFAR-100 is a more challenging dataset (more classes) for the generators to learn to protect the base models. Nevertheless, NeuralFuse can still function to restore some degraded accuracy; these results also demonstrate that our NeuralFuse is applicable to different datasets.
In addition, although the recover percentage is less obvious on CIFAR-100 (the more difficult dataset), we can still conclude that our NeuralFuse is applicable to different datasets.
\input{assets/_tables/relaxed_access/_tables_Appendix_CIFAR100}
\input{assets/_figures/appendix_relaxed_access/cifar100}

\section{Additional Experimental Results on Restricted Access Settings (Transferability)}\label{sec:appendix_transferability}

We conduct more experiments with \textit{Restricted Access} settings to show that our NeuralFuse can be transferred to protect various black-box models. The experimental results are shown in Sec. \ref{sec:appendix_restricted_access_cifar10} (CIFAR-10), Sec. \ref{sec:appendix_restricted_access_gtsrb} (GTSRB), and Sec. \ref{sec:appendix_restricted_access_cifar100} (CIFAR-100).

We find that using VGG19 as a white-box surrogate model has better \textit{transferability} than ResNet18 for all datasets. In addition, we can observe that some NeuralFuse generators have \textit{downward applicability} if base models have a similar architecture. In other words, if we try to transfer a generator trained on a large BER (e.g., 1\%) to a model with a small BER (e.g., 0.5\%), the performance will be better than that of a generator trained with the original BER (e.g., 0.5\%).
For example, in Table \ref{table:Appendixtransferabilityp1CIFAR10}, we could find that if we use VGG19 as a source model to train the generator ConvL (1\%), the generator could deliver better performance (in terms of PA (NF)) when applied to similar base models (e.g., VGG11, VGG16, or VGG19) under a 0.5\% BER, compared to using itself as a source model (shown in Table \ref{table:Appendix_relaxed_access_cifar10}). We conjecture that this is because the generators trained on a larger BER can also cover the error patterns of a smaller BER, and thus they have better generalizability across smaller B.E.Rs.

To further improve the transferability to cross-architecture target models, we also conduct an experiment in Sec. \ref{sec:appendix_restricted_access_generator_ensemble} to show that using ensemble-based training can help the generator to achieve this feature.

\subsection{CIFAR-10}\label{sec:appendix_restricted_access_cifar10}
The results of CIFAR-10 in which NeuralFuse is trained at 1\% BER are shown in Table \ref{table:Appendixtransferabilityp1CIFAR10}.
\input{assets/_tables/_tables_Appendix_Appendixtransferabilityp1CIFAR10}
\clearpage
\subsection{GTSRB}\label{sec:appendix_restricted_access_gtsrb}
In Tables \ref{table:Appendixtransferabilityp15GTSRB} and \ref{table:Appendixtransferabilityp1GTSRB}, we show the results on GTSRB in which NeuralFuse is trained at 1.5\% and 1\% BER, respectively.
\input{assets/_tables/_tables_Appendix_Appendixtransferabilityp15GTSRB}
\input{assets/_tables/_tables_Appendix_Appendixtransferabilityp1GTSRB}
\clearpage
\subsection{CIFAR-100}\label{sec:appendix_restricted_access_cifar100}
In Tables \ref{table:Appendixtransferabilityp1CIFAR100} and \ref{table:Appendixtransferabilityp05CIFAR100}, we show results on CIFAR-100 with NeuralFuse trained at 1\% and 0.5\% BER, respectively.
\input{assets/_tables/_tables_Appendix_Appendixtransferabilityp1CIFAR100}
\input{assets/_tables/_tables_Appendix_Appendixtransferabilityp05CIFAR100}

\clearpage
\subsection{Generator Ensembling}\label{sec:appendix_restricted_access_generator_ensemble}
To improve the transferability performance on cross-architecture cases (e.g., using ResNet-based models as surrogate models to train NeuralFuse and then transfer NeuralFuse to VGG-based target models), we try to adopt ensemble surrogate models to train our NeuralFuse. The experimental results are shown in Table \ref{table:AppendixtransferabilityDoublep15CIFAR10}. We use the same experimental settings mentioned in Table \ref{table:Maintransferabilityp1} but change one source model (e.g., ResNet18 or VGG19) into two (ResNet18 with VGG19) for training. The results show that the overall performance is better than the results shown in Table \ref{table:Maintransferabilityp1}, which means ensemble-based training can easily solve the performance degradation on cross-architecture target models.
\input{assets/_tables/_tables_Appendix_AppendixtransferabilityDoublep15CIFAR10.tex}

\vspace{-5mm}\section{NeuralFuse on Reduced-precision Quantization and Random Bit Errors}\label{sec:appendix_low_precision_quantization}
As mentioned in Sec. \ref{sec:appendix_quant_precision_loss_neuralfuse}, we explore the robustness of NeuralFuse to low-precision quantization of model weights and consider the case of random bit errors. Here, we demonstrate that NeuralFuse can recover not only the accuracy drop due to reduced precision, but also the drop caused by low-voltage-induced bit errors (0.5\% BER) under low precision. We selected two NeuralFuse generators (ConvL and UNetL) for our experiments, and these generators were trained with the corresponding base models (ResNet18 and VGG19) at 1\% BER (CIFAR-10, GTSRB) and 0.5\% BER (ImageNet-10). The experimental results are shown as follows: CIFAR-10 (Sec. \ref{sec:appendix_low_precision_cifar10}), GTSRB (Sec. \ref{sec:appendix_low_precision_gtsrb}), and ImageNet-10 (Sec. \ref{sec:appendix_low_precision_imagenet10}).  Similarly, for ease of comparison, we visualize the experimental results in the figures below each table. Our results show that NeuralFuse can consistently perform well in low-precision regimes as well as recover the low-voltage-induced accuracy drop.
\subsection{CIFAR-10}\label{sec:appendix_low_precision_cifar10}

\input{assets/_tables/low_precision_bit_errors/cifar10}
\newpage

\input{assets/_figures/low_precision_quantization/cifar10}

\subsection{GTSRB}\label{sec:appendix_low_precision_gtsrb}
\input{assets/_tables/low_precision_bit_errors/gtsrb}
\input{assets/_figures/low_precision_quantization/gtsrb}

\clearpage

\subsection{ImageNet-10}\label{sec:appendix_low_precision_imagenet10}
\input{assets/_tables/low_precision_bit_errors/imagenet10}
\input{assets/_figures/low_precision_quantization/imagenet10}

\section{Additional Experiments on Adversarial Training}\label{sec:appendix_AdvTraining}
Adversarial training is a common strategy to derive a robust neural network against certain perturbations. By training the generator using adversarial training proposed in \citet{ErrorAwareTraining}, we report its performance against low voltage-induced bit errors.
We use ConvL as the generator and ResNet18 as the base model, trained on CIFAR-10. Furthermore, we explore different $K$ flip bits as the perturbation on weights of the base model during adversarial training, and then for evaluation, the trained-generator will be applied against 1\% of bit errors rate on the base model. The results are shown in Table \ref{table:AppendixAdvTrainingp1CIFAR10}.  After careful tuning of hyperparameters, we find that we are not able to obtain satisfactory recovery when adopting adversarial training. Empirically, we argue that adversarial training may not be suitable for training generator-based methods.
\input{assets/_tables/_tables_Appendix_AdversarialTraining.tex}

\section{Additional Experiments on Robust Model Trained with Adversarial Weight Perturbation with NeuralFuse}\label{sec:appendix_robust_model_neuralfuse}
Previously, \citet{awp} proposed that one could obtain a more robust model via adversarial weight perturbation. To seek whether such models could also be robust to random bit errors, we conducted an experiment on CIFAR-10 with the proposed adversarially trained PreAct ResNet18. The experimental results are shown in Table \ref{table:AppendixRobustCifar10Table}. We find that the average perturbed accuracy is 23\% and 63.2\% for PreAct ResNet18 under 1\% and 0.5\% BER, respectively. This result is lower than 38.9\% and 70.1\% from ResNet18 in Table \ref{table:Appendix_relaxed_access_cifar10}, indicating their poor generalization ability against random bit errors. Nevertheless, when equipped  NeuralFuse on the perturbed model, we could still witness a significant recover percentage under both 1\% and 0.5\% BER This result further demonstrates that NeuralFuse could be adapted to various models (i.e., trained in different learning algorithms).

\input{assets/_tables/_tables_Appendix_AppendixRobustModelPreResnet18CIFAR10}

\section{Data Embeddings Visualization}\label{sec:visualization}
To further understand how our proposed NeuralFuse works, we visualize the output distribution from the final linear layer of the base models and project the results onto the 2D space using t-SNE \citep{TSNE}. Figure \ref{fig:TSNE} shows the output distribution from ResNet18 (trained on CIFAR-10) under a 1\% bit error rate. We chose two generators that have similar architecture: ConvL and ConvS, for this experiment. We can observe that: (a) The output distribution of the clean model without NeuralFuse can be grouped into 10 classes denoted by different colors. (b) The output distribution of the perturbed model under a 1\% bit error rate without NeuralFuse shows mixed representations and therefore degraded accuracy. (c) The output distribution of the clean model with ConvL shows that applying NeuralFuse will not hurt the prediction of the clean model too much (i.e., it retains high accuracy in the regular voltage setting). (d) The output distribution of the perturbed model with ConvL shows high separability (and therefore high perturbed accuracy) as opposed to (b).
(e)/(f) shows the output distribution of the clean/perturbed model with ConvS.
For both (e) and (f), we can see nosier clustering when compared to (c) and (d), which means the degraded performance of ConvS compared to ConvL. 
The visualization validates that NeuralFuse can help retain good data representations under random bit errors and that larger generators in NeuralFuse have better performance than smaller ones.

\input{assets/_figures/visualization}

\section{Qualitative Analysis of Transformed Inputs}\label{sec:visual_transformed}
In this section, we conduct a qualitative study to visualize the images transformed by NeuralFuse and present some properties and observations of these images. We utilize six different architectures of NeuralFuse generators trained with ResNet18 under a $1\%$ bit error rate.

Figure \ref{fig:VisualizationImages} (a) showcases several images from the \textit{truck} class in CIFAR-10. Notably, images of the same class, when transformed by the same NeuralFuse, exhibit similar patterns, such as circles symbolizing the wheels of the trucks.

In Figures \ref{fig:VisualizationImages} (b) and \ref{fig:VisualizationImages} (c), we observe analogous phenomena in the GTSRB and CIFAR-100 datasets. Transformed images of the same class using the same generator consistently display patterns. On GTSRB, NeuralFuse-generated patterns highlight the sign's shape with a green background, even if the original images have a dark background and are under different lighting conditions. These results further underscore the efficacy and efficiency of NeuralFuse.

Figure \ref{fig:VisualizationImagesALL} presents more images from different classes in (a) CIFAR-10, (b) GTSRB, and (c) CIFAR-100. The transformed images exhibit distinct patterns for each class, suggesting that NeuralFuse effectively transforms images into class-specific patterns, making associated features robust to random bit errors and easily recognizable by the base model in low-voltage settings.

\input{assets/_figures/visualization_transformed}
\input{assets/_figures/visualization_transformed_all}

%% file: assets/TrainingAlgo.tex
\begin{algorithm}[h]
\caption{Training steps for NeuralFuse}
\label{alg:algorithm}
\textbf{Input}: Base model $M_0$; Generator $\mathcal{G}$; Training data samples $\mathcal{X}$; Distribution of the perturbed models $\mathcal{M}_p$; Number of perturbed models $N$; Total training iterations $T$ \\
\textbf{Output}: Optimized parameters $\mathcal{W}_{\mathcal{G}}$ for the Generator $\mathcal{G}$
\begin{algorithmic}[1]
\FOR {$t=0,...,T-1$}
    \FORALL {mini-batches $\{\mathbf{x}, y \}^{B}_{b=1} \sim \mathcal{X}$}
    \STATE Create transformed inputs 
    $\mathbf{x}_{t}=\mathcal{F}(\mathbf{x})=\text{clip}_{[-1,1]}\big(\mathbf{x}+\mathcal{G}(\mathbf{x})\big)$.
        \STATE Sample $N$ perturbed models $\{M_{p_{1}},..., M_{p_{N}}\}$ from  $\mathcal{M}_p$ under $p\%$ random bit errors.
        \FORALL {$M_{p_i} \sim \{M_{p_{1}},..., M_{p_{N}}\}$}
        \STATE Calculate the loss $\mathcal{L}_{p_i}$ based on the output of the perturbed model $M_{p_i}$. Then calculate the gradients $g_{p_i}$ for $\mathcal{W}_{\mathcal{G}}$ based on $\mathcal{L}_{p_i}$.
        \ENDFOR
        \STATE Calculate the loss $\mathcal{L}_{0}$ based on the output of the clean model $M_{0}$. Then calculate the gradients $g_{0}$ for $\mathcal{W}_{\mathcal{G}}$ based on $\mathcal{L}_{0}$.
        \STATE Calculate the final gradient $g_{final}$ using (\ref{eqn:gradient_calculation}) based on $g_{0}$ and $g_{p_1}, ..., g_{p_N}$. 
        \STATE Update $\mathcal{W}_{\mathcal{G}}$ using $g_{final}$.
    \ENDFOR
\ENDFOR
\end{algorithmic}
\label{alg:EOPM_Algo}
\end{algorithm}

%% file: assets/_tables/_tables_Appendix_GeneratorArchConDeCon.tex
\begin{table*}[ht]
\setlength{\tabcolsep}{2pt}
\renewcommand{\arraystretch}{1.15}
\vspace{-1mm}
\centering
\caption{Model architecture for both Convolution-based and Deconvolution-based generators. Each ConvBlock consists of a Convolution ($\text{kernel}=3\times3$, $\text{padding}=1$, $\text{stride}=1$), a Batch Normalization, and a ReLU layer. Each DeConvBlock consists of a Deconvolution ($\text{kernel}=4\times4$, $\text{padding}=1$, $\text{stride}=2$), a Batch Normalization, and a ReLU layer.}
\label{table:GeneratorArchConDeCon}
\begin{adjustbox}{max width=\linewidth}
\begin{threeparttable}
\begin{tabular}{ cc|cc|cc|cc }

\toprule[2pt]
\multicolumn{2}{c|}{ConvL} & \multicolumn{2}{c|}{ConvS} & \multicolumn{2}{c|}{DeConvL} & \multicolumn{2}{c}{DeConvS} \\
Layers & \#CHs & Layers & \#CHs & Layers & \#CHs & Layers & \#CHs  \\
\midrule
(ConvBlock)$\times$2, MaxPool & 32 & ConvBlock, Maxpool & 32 & (ConvBlock)$\times$2, MaxPool & 32 & ConvBlock, Maxpool & 32  \\
(ConvBlock)$\times$2, MaxPool & 64 & ConvBlock, Maxpool & 64 & (ConvBlock)$\times$2, MaxPool & 64 & ConvBlock, Maxpool & 64 \\
(ConvBlock)$\times$2, MaxPool & 128 & ConvBlock, Maxpool & 64 & (ConvBlock)$\times$2, MaxPool, & 128 & ConvBlock, Maxpool & 64 \\
ConvBlock, UpSample, ConvBlock & 128 & ConvBlock, UpSample & 64 & ConvBlock & 128 & DeConvBlock & 64 \\
ConvBlock, UpSample, ConvBlock & 64 & ConvBlock, UpSample & 32 & DeConvBlock, ConvBlock & 64 & DeConvBlock & 32 \\
ConvBlock, UpSample, ConvBlock& 32 & ConvBlock, UpSample & 3 & DeConvBlock, ConvBlock & 32 & DeConv, BN, Tanh & 3 \\
Conv, BN, Tanh & 32 & Conv, BN, Tanh & 3 & Conv, BN, Tanh & 3 & & \\
\bottomrule[2pt]

\end{tabular}
    \begin{tablenotes}
      \large
      \item {[Note] \#CHs: \textit{number of channels}.}
    \end{tablenotes}
\end{threeparttable}
\end{adjustbox}
\vspace{-2mm}
\end{table*}

%% file: assets/_tables/_tables_Appendix_GeneratorArchUNet.tex
\begin{table*}[ht]
\renewcommand{\arraystretch}{1.05}
\centering
\caption{Model architecture for UNet-based generators. Each ConvBlock consists of a Convolution ($\text{kernel}=3\times3$, $\text{padding}=1$, $\text{stride}=1$), a Batch Normalization, and a ReLU layer. Other layers, such as the Deconvolutional layer ($\text{kernel}=2\times2$, $\text{padding}=1$, $\text{stride}=2$), are used in UNet-based models. For the final Convolution layer, the kernel size is set to 1.}
\label{table:GeneratorArchUNet}
\begin{adjustbox}{max width=.75\linewidth}
\begin{tabular}{ cc|cc }

\toprule[1.5pt]
\multicolumn{2}{c}{UNetL} & \multicolumn{2}{|c}{UNetS} \\
Layers & \#Channels & Layers & \#Channels \\
\midrule
L1: (ConvBlock)$\times$2 & 16 &  L1: (ConvBlock)$\times$2 & 8\\
L2: Maxpool, (ConvBlock)$\times$2 & 32 & L2: Maxpool, (ConvBlock)$\times$2 & 16 \\
L3: Maxpool, (ConvBlock)$\times$2 & 64 & L3: Maxpool, (ConvBlock)$\times$2 & 32 \\
L4: Maxpool, (ConvBlock)$\times$2 & 128 & L4: Maxpool, (ConvBlock)$\times$2 & 64\\
L5: DeConv & 64 & L5: DeConv & 32 \\
L6: Concat[L3, L5] & 128 & L6: Concat[L3, L5] & 64\\ 
L7: (ConvBlock)$\times$2 & 64 & L7: (ConvBlock)$\times$2 & 32 \\
L8: DeConv & 32 & L8: DeConv & 16 \\
L9: Concat[L2, L8] & 64 & L9: Concat[L2, L8] &  32\\
L10: (ConvBlock)$\times$2 & 32 & L10: (ConvBlock)$\times$2 & 16 \\
L11: DeConv & 16 & L11: DeConv & 8\\
L12: Concat[L1, L11] & 32 & L12: Concat[L1, L11] & 16 \\
L13: (ConvBlock)$\times$2 & 16 & L13: (ConvBlock)$\times$2 & 8  \\
L14: Conv & 3 & L14: Conv & 3\\
\bottomrule[1.5pt]
\end{tabular}
\end{adjustbox}
\end{table*}

%% file: assets/_tables/_tables_Appendix_Total_Mem_E.tex
\begin{table*}[ht]
\vspace{-3mm}
\setlength\doublerulesep{1pt}
\centering
\begin{threeparttable}
\caption{The total weights memory access calculated by SCALE-SIM.}
\label{table:Total_Mem_E}
\begin{tabular}{c|cccccc}
\toprule
Base Model & ResNet18 & ResNet50 & VGG11 & VGG16 & VGG19 & -  \\
\midrule
T.W.M.A. & 2,755,968 & 6,182,144 & 1,334,656 & 2,366,848 & 3,104,128 & - \\
\midrule\midrule
NeuralFuse & ConvL & ConvS & DeConvL & DeConvS & UNetL & UNetS \\
\midrule
T.W.M.A. & 320,256 & 41,508 & 259,264 & 86,208 & 180,894 & 45,711 \\
\bottomrule
\end{tabular}
    \begin{tablenotes}
      \small
      \item {[Note] T.W.M.A.: \textit{total weight memory access}.}
    \end{tablenotes}
\end{threeparttable}
\end{table*}

%% file: assets/_figures/tradeoffWholeModels.tex
\begin{figure}[ht]
\vspace{-3mm}
    \centering
    \includegraphics[width=.85\linewidth]{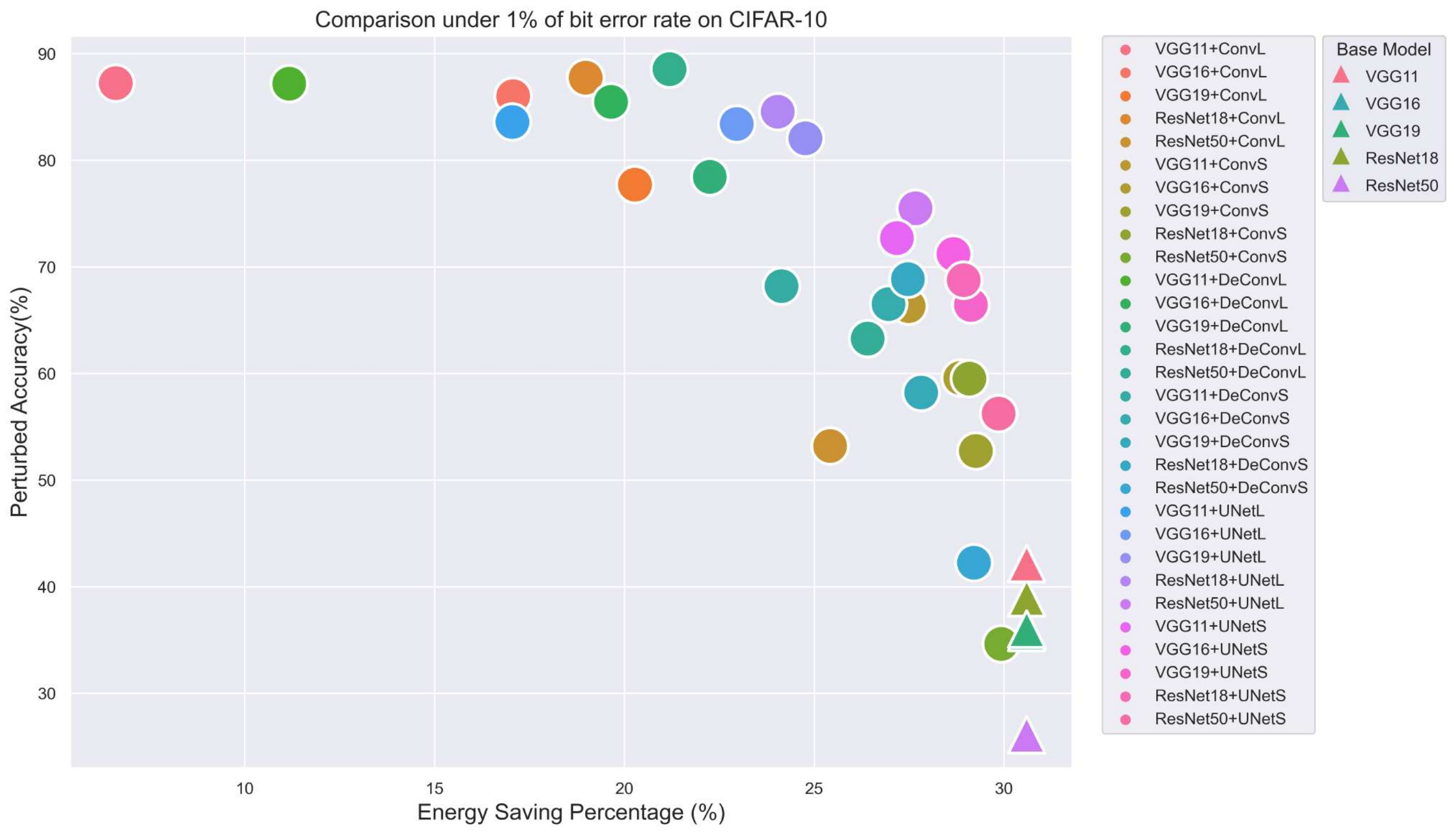}
    \caption{The energy/accuracy tradeoff of different NeuralFuse implementations with all CIFAR-10 pre-trained based models. The X-axis represents the percentage reduction in dynamic memory access energy at low-voltage settings (base model protected by NeuralFuse) compared to the bit-error-free (nominal) voltage; the Y-axis represents the perturbed accuracy (evaluated at low voltage) with a 1\% bit error rate.}
    \label{fig:WholeModel}\vspace{-2mm}
\end{figure}

%% file: assets/_tables/_tables_Appendix_Param_MACs.tex
\begin{table*}[ht]
\centering
\caption{Parameter counts and MACs for all base models and generators in this paper.}

\begin{adjustbox}{max width=.9\linewidth}
\begin{tabular}{ cccccccc }
\toprule
& \multirow{2}{*}{} & \multicolumn{6}{c}{Base Model}  \\
 &  & ResNet18 & ResNet50 & VGG11 & VGG16 & VGG19 & -  \\
\midrule
\multirow{2}{*}{Parameter} & CIFAR-10 & 11,173,962 & 23,520,842 & 9,231,114 & 14,728,266 & 20,040,522 & \multirow{2}{*}{-} \\
 & ImageNet-10 & 11,181,642 & 23,528,522 & 128,812,810 & 134,309,962 & 139,622,218 &  \\
\midrule
\multirow{2}{*}{MACs} & CIFAR-10 & 557.14M & 1.31G & 153.5M & 314.43M & 399.47M & \multirow{2}{*}{-} \\
& ImageNet-10 & 1.82G & 4.12G & 7.64G & 15.53G & 19.69G &  \\

\midrule[0.1pt]\midrule[0.1pt]
& \multirow{2}{*}{} & \multicolumn{6}{c}{NeuralFuse}  \\
 &  & ConvL & ConvS & DeConvL & DeConvS & UNetL & UNetS \\
\midrule
Parameter & \begin{tabular}{@{}c@{}}CIFAR-10\\ImageNet-10\end{tabular}  & 723,273 & 113,187 & 647,785 & 156,777 & 482,771 & 121,195 \\
\midrule
\multirow{2}{*}{MACs} & CIFAR-10 & 80.5M & 10.34M & 64.69M & 22.44M & 41.41M & 10.58M \\
& ImageNet-10 & 3.94G & 506.78M & 3.17G & 1.1G & 2.03G & 518.47M \\
\bottomrule
\end{tabular}
\end{adjustbox}
\label{table:Param_MACs}
\end{table*}

%% file: assets/_tables/_tables_Appendix_EnergySavingBasedOnMACs.tex
\begin{table*}[ht]
\centering
\caption{The MACs-Based energy saving percentage ($\%$) for different combinations of base models and NeuralFuse.}
\label{table:EnergySavingMACs}
\begin{adjustbox}{max width=.7\linewidth}
\begin{tabular}{c|rrrrrr}

\hline
 & ConvL & ConvS & DeConvL & DeConvS & UNetL & UNetS \\
\hline
ResNet18 & 16.2 & 28.7 & 19.0 & 26.6 & 23.2 & 28.7 \\
ResNet50 & 24.5 & 29.8 & 25.7 & 28.9 & 27.4 & 29.8 \\
VGG11 & -21.8 & 23.9 & -11.5 & 16.0 & 3.6 & 23.7 \\
VGG16 & 5.0 & 27.3 & 10.0 & 23.5 & 17.4 & 27.2 \\
VGG19 & 10.4 & 28.0 & 14.4 & 25.0 & 20.2 & 28.0 \\
\hline
\end{tabular}
\end{adjustbox}
\end{table*}

%% file: assets/_figures/tradeoffWholeModelsMACs.tex
\begin{figure}[h]
    \centering
    \includegraphics[width=.9\linewidth]{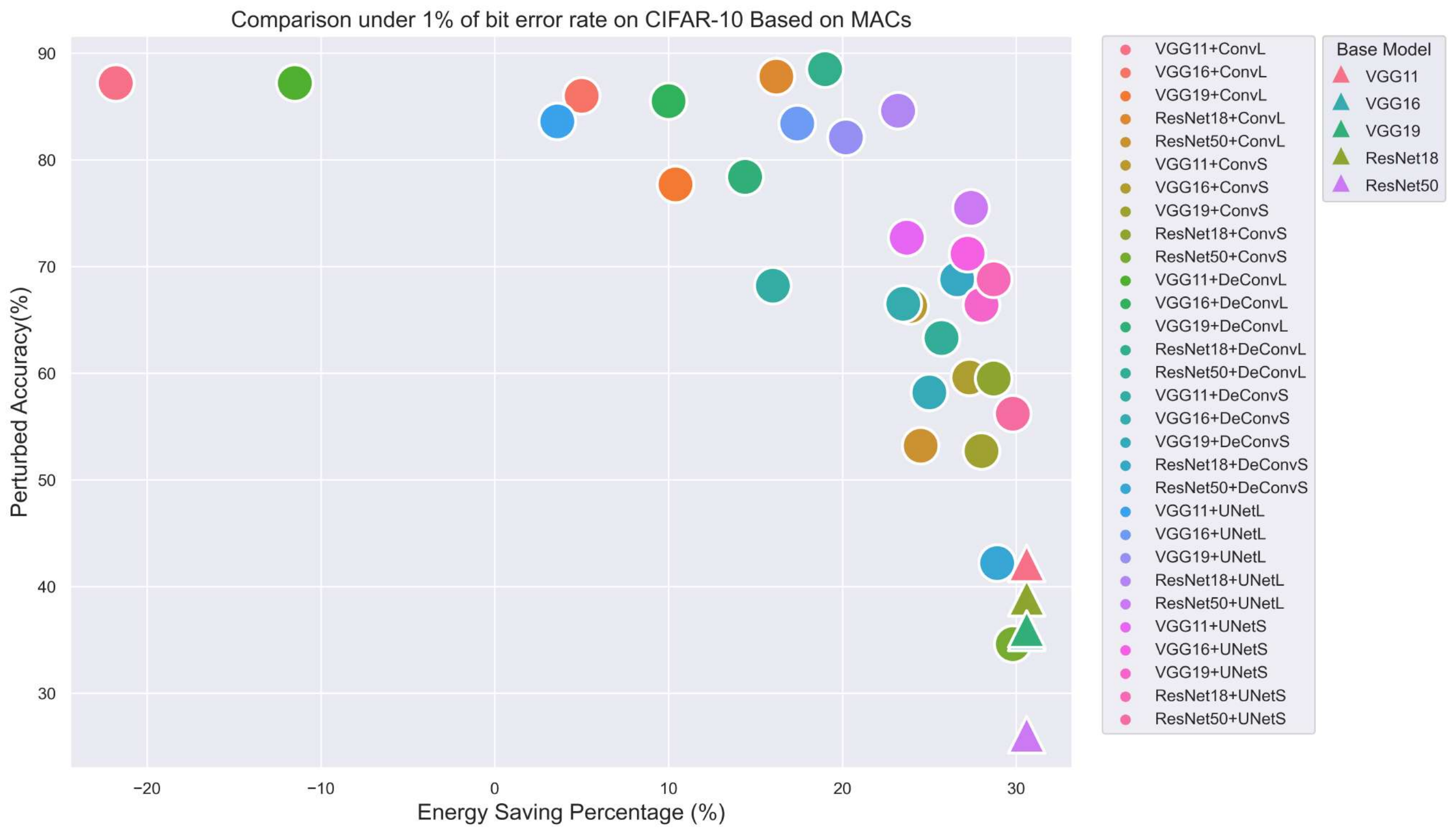}
    \caption{The Mac-based energy/accuracy tradeoff of different NeuralFuse implementations with all CIFAR-10 pre-trained based models. The X-axis represents the percentage reduction in dynamic memory access energy at low-voltage settings (base model protected by NeuralFuse), compared to the bit-error-free (nominal) voltage; the Y-axis represents the perturbed accuracy (evaluated at low voltage) with a 1\% bit error rate.}
    \label{fig:WholeModelMACsBased}
\end{figure}

%% file: assets/_tables/_tables_Appendix_AppendixLatencyNeuralFuse.tex
\begin{table*}[ht]
\centering
\caption{The Inference Latency of base model and base model with NeuralFuse.}
\label{table:latency}
\begin{adjustbox}{max width=.9\linewidth}
\begin{tabular}{c|rrrr}

\toprule
 & ResNet18 (CIFAR-10) & VGG19 (CIFAR-10) & ResNet18 (ImageNet-10) & VGG19 (ImageNet-10) \\
\midrule
Base Model & 5.84 ms & 5.32 ms & 6.21 ms & 14.34 ms \\
\midrule

\multicolumn{1}{r|}{+~ConvL} & 9.37 ms (+3.53) & 8.96 ms (+3.64) & 10.51 ms (+4.3) & 17.66 ms (+3.32) \\
\multicolumn{1}{r|}{+~ConvS} & 7.86 ms (+2.02) & 7.40 ms (+2.08) & 8.28 ms (+2.07) & 16.72 ms (+2.38) \\
\multicolumn{1}{r|}{+~DeConvL} & 9.18 ms (+3.34) & 8.59 ms (+3.27) & 10.07 ms (+3.86) & 17.24 ms (+2.90) \\
\multicolumn{1}{r|}{+~DeConvS} & 7.49 ms (+1.65) & 7.04 ms (+1.72) & 7.79 ms (+1.58) & 15.67 ms (+1.33) \\
\multicolumn{1}{r|}{+~UNetL} & 10.69 ms (+4.85) & 10.06 ms (+4.74) & 11.14 ms (+4.93) & 18.54 ms (+4.20) \\
\multicolumn{1}{r|}{+~UNetS} & 10.63 ms (+4.79) & 10.13 ms (+4.81) & 11.36 ms (+5.15) & 18.60 ms (+4.26) \\
\bottomrule
\end{tabular}
\end{adjustbox}
\vspace{-1mm}
\end{table*}

%% file: assets/_figures/error_bar.tex
\begin{figure}[ht]
    \centering
    \begin{subfigure}[b]{0.4\textwidth}
        \centering
        \includegraphics[width=.95\textwidth]{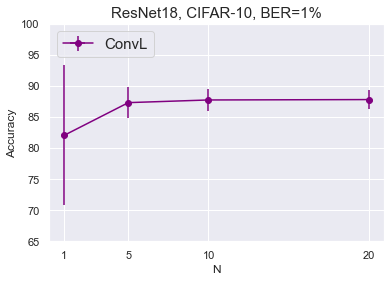}
        \caption{Using ConvL}\label{fig:TestN_ConvL}
    \end{subfigure}
    \hspace{8mm}
    \begin{subfigure}[b]{0.4\textwidth}
        \centering
        \includegraphics[width=.95\textwidth]{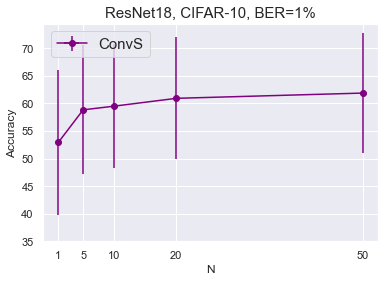}
        \caption{Using ConvS}\label{fig:TestN_ConvS}
    \end{subfigure}
       \caption{The experimental results on using different sizes of $N$ for EOPM.}
       \label{fig:TestN}
\end{figure}

%% file: assets/_tables/_tables_Appendix_AppendixlambdaCifar10.tex
\begin{table*}[h!t]
\centering
\begin{threeparttable}
\caption{Experimental results based on $\lambda$ value choosing. The results show that $\lambda=5$ can balance the tradeoff between clean accuracy and perturbed accuracy.}
\label{table:AppendixLambdaCIFAR10}
\begin{tabular}{ c|c|cc|crr }
\toprule[1.5pt]
\multirow{2}{*}{\begin{tabular}{@{}c@{}}Base \\ Model\end{tabular}} & \multirow{2}{*}{$\lambda$} & \multirow{2}{*}{CA} & \multirow{2}{*}{PA} & \multicolumn{3}{c}{ConvL} \\ 
& & & & CA (NF) & PA (NF) & RP \\
\midrule

\multirow{5}{*}{ResNet18} & 10 & \multirow{5}{*}{92.6} & \multirow{5}{*}{38.9 $\pm$ 12.4} & 90.1 & 88.0 $\pm$ 1.7 & 49.1 \\
& 5 & & & 89.8 & 87.8 $\pm$ 1.7 & 48.8\\
& 1 & & & 90.0 & 84.2 $\pm$ 3.8 & 45.3\\
& 0.1 & & & 91.6 & 65.7 $\pm$ 9.3 & 26.8\\
& 0.01 & & & 92.2 & 43.6 $\pm$ ~13 & 4.7\\
\midrule

\multirow{5}{*}{VGG19} & 10 & \multirow{5}{*}{90.5} & \multirow{5}{*}{36.0 $\pm$ 12.0} & 89.6 & 77.9 $\pm$ ~19 & 41.9 \\
& 5 & & & 89.8 & 77.7 $\pm$ ~19 & 41.7\\
& 1 & & & 89.9 & 73.1 $\pm$ ~19 & 37.1\\
& 0.1 & & & 89.1 & 51.2 $\pm$ ~16 & 15.2\\
& 0.01 & & & 90.2 & 36.8 $\pm$ ~12 & 0.8\\
\bottomrule[1.5pt]

\end{tabular}
    \begin{tablenotes}
      \small
      \item {\textit{Note.} CA (\%): clean accuracy; PA (\%): perturbed accuracy; NF: NeuralFuse; and RP: total recover percentage of PA (NF) vs. PA}
    \end{tablenotes}
\end{threeparttable}
\end{table*}

%% file: assets/_tables/_tables_experiments_universal.tex
\begin{table*}[ht]
\setlength{\tabcolsep}{12pt}
\centering
\begin{threeparttable}
\caption{Performance of the universal input perturbation (UIP) trained by EOPM on CIFAR-10 pre-trained ResNet18.}
\label{table:Mainuniversalp1CIFAR10}
    \begin{tabular}{ c|c|cc|ccr }
        \toprule[1.5pt]
        Base Model & BER & CA & PA & CA (\textbf{UIP}) & PA (\textbf{UIP}) & RP \\ 
        \midrule
       \multirow{2}{*}{ResNet18} & 1\% & \multirow{2}{*}{92.6} & 38.9 $\pm$ 12.4 & 91.8 & 37.9 $\pm$ ~11 & -1.0 \\ 
        & 0.5\% & & 70.1 $\pm$ 11.6 & 92.5 & 70.6 $\pm$ ~11 & 0.5 \\ 
        \midrule
        \multirow{2}{*}{ResNet50} & 1\% & \multirow{2}{*}{92.6} & 26.1 $\pm$ ~~9.4 & 80.7 & 21.0 $\pm$ 5.9 & -5.1 \\
        & 0.5\% & & 61.0 $\pm$ 10.3 & 91.9 & 62.4 $\pm$ ~12 & 1.4 \\ 
        \midrule
        \multirow{2}{*}{VGG11} & 1\% & \multirow{2}{*}{88.4} & 42.2 $\pm$ 11.6 & 86.9 & 43.0 $\pm$ ~11 & 0.8 \\ 
        & 0.5\% & & 63.6 $\pm$ ~~9.3 & 88.2 & 64.2 $\pm$ 8.8 & 0.6 \\ 
        \midrule
        \multirow{2}{*}{VGG16} & 1\% & \multirow{2}{*}{90.3} & 35.7 $\pm$ ~~7.9 & 90.1 & 37.1 $\pm$ 8.5 & 1.4 \\
        & 0.5\% & & 66.6 $\pm$ ~~8.1& 90.4 & 67.3 $\pm$ 8.1 & 0.7 \\ 
        \midrule
        \multirow{2}{*}{VGG19} & 1\% & \multirow{2}{*}{90.5} & 36.0 $\pm$ 12.0 & 89.9 & 35.3 $\pm$ ~12 & -0.7 \\
        & 0.5\% & & 64.2 $\pm$ 12.4 & 90.1 & 64.4 $\pm$ ~12 & 0.2 \\ 
        \bottomrule[1.5pt]
    \end{tabular}
    \begin{tablenotes}
      \small
      \item {\textit{Note}. BER: the bit-error rate of the base model; CA ($\%$): clean accuracy; PA ($\%$): perturbed accuracy; UIP: universal input transformation parameter;  and RP: total recovery percentage of PA (UIP) vs. PA}
    \end{tablenotes}
  \end{threeparttable}
\end{table*}

%% file: assets/_tables/relaxed_access/_tables_Appendix_CIFAR10.tex
\begin{table*}[ht]
\centering
\caption{Testing accuracy (\%) under 1\% and 0.5\% of random bit error rate on CIFAR-10.}
\label{table:Appendix_relaxed_access_cifar10}
\begin{adjustbox}{max width=\linewidth}
\begin{threeparttable}
\begin{tabular}{ c|c|c|ccrr|ccrr}

\toprule[1.5pt]
\multirow{2}{*}{\begin{tabular}{@{}c@{}}Base \\ Model\end{tabular}} & \multirow{2}{*}{NF} & \multirow{2}{*}{CA} &\multicolumn{4}{c|}{\textbf{1\% BER}} & \multicolumn{4}{c}{\textbf{0.5\% BER}}\\
& & & PA & CA (NF) & PA (NF) & RP 
    & PA & CA (NF) & PA (NF) & RP \\
\midrule
\multirow{6}{*}{ResNet18} & ConvL &
\multirow{6}{*}{92.6} & & 89.8 & 87.8 $\pm$ 1.7 & 48.8& 
& 90.4 & 87.9 $\pm$ 2.2 & 17.8 \\ 
& ConvS &
& & 88.2 & 59.5 $\pm$ ~11 & 20.6 & 
& 91.7 & 78.4 $\pm$ 8.3 & 8.3 \\ 
& DeConvL &
& 38.9 & 89.6 & 88.5 $\pm$ 0.8 & 49.6 & 
70.1 & 90.2 & 90.0 $\pm$ 0.2 & 19.9 \\ 
& DeConvS & 
& $\pm$ 12.4 & 82.9 & 68.8 $\pm$ 6.4 & 29.9 & 
$\pm$ 11.6 & 84.1 & 79.9 $\pm$ 3.6 & 9.8 \\ 
& UNetL &
& & 86.6 & 84.6 $\pm$ 0.8 & 45.6 & 
& 89.7 & 86.3 $\pm$ 2.4 & 16.2 \\ 
& UNetS &
& & 84.4 & 68.8 $\pm$ 6.0 & 29.8 & 
& 90.9 & 80.7 $\pm$ 5.8 & 10.7 \\ 
\midrule

\multirow{6}{*}{ResNet50} & ConvL &
\multirow{6}{*}{92.6} & & 85.5 & 53.2 $\pm$ ~22 & 27.1 & 
& 90.3 & 86.5 $\pm$ 3.2 & 25.5 \\ 
& ConvS &
& & 85.2 & 34.6 $\pm$ ~14 & 8.5 & 
& 90.8 & 73.3 $\pm$ 8.7 & 12.3 \\ 
& DeConvL &
& 26.1 & 87.4  & 63.3 $\pm$ ~21 & 37.2 & 
61.0 & 89.5  & 87.2 $\pm$ 2.5 & 26.2 \\ 
& DeConvS &
& $\pm$ 9.4 & 82.4  & 42.2 $\pm$ ~17 & 16.1 & 
$\pm$ 10.3 & 90.3 & 75.5 $\pm$ 8.1 & 14.5 \\ 
& UNetL &
& & 86.2 & 75.5 $\pm$ ~12 & 49.4 & 
& 89.9 & 83.9 $\pm$ 3.6 & 22.9 \\ 
& UNetS & 
& & 77.3 & 56.2 $\pm$ ~19 & 30.1 & 
& 89.7 & 76.1 $\pm$ 7.2 & 15.1 \\ 
\midrule

\multirow{6}{*}{VGG11} & ConvL & 
\multirow{6}{*}{88.4} & & 89.6 & 87.2 $\pm$ 2.9 & 45.1 & 
& 89.8 & 87.0 $\pm$ 1.3 & 23.3 \\ 
& ConvS & 
& & 84.9 & 66.3 $\pm$ 7.5 & 24.1 & 
& 88.2 & 74.5 $\pm$ 5.7 & 10.9 \\ 
& DeConvL & 
& 42.2 & 89.3  & 87.2 $\pm$ 2.6 & 45.0 & 
63.6 & 89.6  & 86.9 $\pm$ 1.1 & 23.2 \\ 
& DeConvS &
& $\pm$ 11.6 & 85.6  & 68.2 $\pm$ 7.1 & 26.0 & 
$\pm$ 9.3 & 88.3  & 75.7 $\pm$ 4.6 & 12.1 \\ 
& UNetL & 
& & 87.1 & 83.6 $\pm$ 1.3 & 41.4 & 
& 88.0 & 82.4 $\pm$ 1.8 & 18.8 \\ 
& UNetS & 
& & 85.5 & 72.7 $\pm$ 4.6 & 30.5 & 
& 88.1 & 75.8 $\pm$ 4.3 & 12.2 \\ 
\midrule

\multirow{6}{*}{VGG16} & ConvL &
\multirow{6}{*}{90.3} & & 90.1 & 86.0 $\pm$ 6.2 & 50.3 & 
& 90.2 & 88.5 $\pm$ 0.9 & 21.9  \\ 
& ConvS &
& & 87.4 & 59.6 $\pm$ ~12 & 23.9 & 
& 89.9 & 77.8 $\pm$ 4.8 & 11.1 \\ 
& DeConvL & 
& 35.7 & 89.7  & 85.5 $\pm$ 6.8 & 49.8 & 
66.6 & 89.7 & 88.2 $\pm$ 1.0 & 21.4 \\ 
& DeConvS & 
& $\pm$ 7.9 & 86.8  & 66.5 $\pm$ ~11 & 30.8 & 
$\pm$ 8.1 & 90.0  & 78.4 $\pm$ 4.7 & 11.8 \\ 
& UNetL &
& & 87.4 & 83.4 $\pm$ 4.4 & 47.7 & 
& 89.0 & 86.2 $\pm$ 1.5 & 19.6 \\ 
& UNetS & 
& & 87.4 & 71.2 $\pm$ 8.2 & 35.5 & 
& 89.0 & 80.2 $\pm$ 3.5 & 13.7 \\ 
\midrule

\multirow{6}{*}{VGG19} & ConvL &
\multirow{6}{*}{90.5} & & 89.8 & 77.7 $\pm$ ~19 & 41.7 & 
& 90.4 & 88.1 $\pm$ 1.8 & 23.9 \\ 
& ConvS & 
& & 87.3 & 52.7 $\pm$ ~17 & 16.7 & 
& 89.6 & 74.5 $\pm$ 9.0 & 10.3 \\ 
& DeConvL &
& 36.0 & 86.3  & 78.4 $\pm$ ~18 & 42.4 & 
64.2 & 90.4  & 88.5 $\pm$ 1.4 & 24.3 \\ 
& DeConvS &
& $\pm$ ~12.0 & 86.5  & 58.2 $\pm$ ~18 & 22.2 & 
$\pm$ 12.4 & 89.7  & 75.2 $\pm$ 8.6 & 11.0 \\ 
& UNetL & 
& & 86.3 & 82.1 $\pm$ 4.8 & 46.0 & 
& 89.1 & 85.0 $\pm$ 2.7 & 20.8 \\ 
& UNetS &
& & 86.3 & 66.4 $\pm$ ~13 & 30.4 & 
& 89.2 & 77.1 $\pm$ 7.3 & 12.9 \\ 
\bottomrule[1.5pt]
\end{tabular}
    \begin{tablenotes}
      \small
      \item {\textit{Note}. CA ($\%$): clean accuracy; PA ($\%$): perturbed accuracy; NF: NeuralFuse; and RP: total recovery percentage of PA (NF) vs. PA}
    \end{tablenotes}
\end{threeparttable}
\end{adjustbox}
\end{table*}

%% file: assets/_figures/appendix_relaxed_access/cifar10.tex
\begin{figure}[ht]
     \vspace{-2mm}
     \centering
     \begin{subfigure}[b]{\textwidth}
         \centering
         \includegraphics[width=\textwidth, trim=-0.2cm -0.1cm 0.2cm 0.1cm, clip]{assets/figures_paths/exp_results/legend.pdf}
     \end{subfigure}
     \hfill
    \begin{adjustbox}{max width=\linewidth}
     \begin{subfigure}[b]{0.195\textwidth}
         \centering
         \includegraphics[width=\textwidth]{assets/figures_paths/exp_results/cifar10_resnet18.pdf}
     \end{subfigure}
     \hfill
     \begin{subfigure}[b]{0.195\textwidth}
         \centering
         \includegraphics[width=\textwidth]{assets/figures_paths/exp_results/cifar10_resnet50.pdf}
     \end{subfigure}
     \hfill
     \begin{subfigure}[b]{0.195\textwidth}
         \centering
         \includegraphics[width=\textwidth]{assets/figures_paths/exp_results/cifar10_vgg11.pdf}
     \end{subfigure}
     \hfill
     \begin{subfigure}[b]{0.195\textwidth}
         \centering
         \includegraphics[width=\textwidth]{assets/figures_paths/exp_results/cifar10_vgg16.pdf}
     \end{subfigure}
     \hfill
     \begin{subfigure}[b]{0.195\textwidth}
         \centering
         \includegraphics[width=\textwidth]{assets/figures_paths/exp_results/cifar10_vgg19.pdf}
     \end{subfigure}
     \end{adjustbox}\\
     {\small (a) CIFAR-10, $1\%$ Bit Error Rate\\}\vspace{2mm}
    \begin{adjustbox}{max width=\linewidth}
     \begin{subfigure}[b]{0.195\textwidth}
         \centering
         \includegraphics[width=\textwidth]{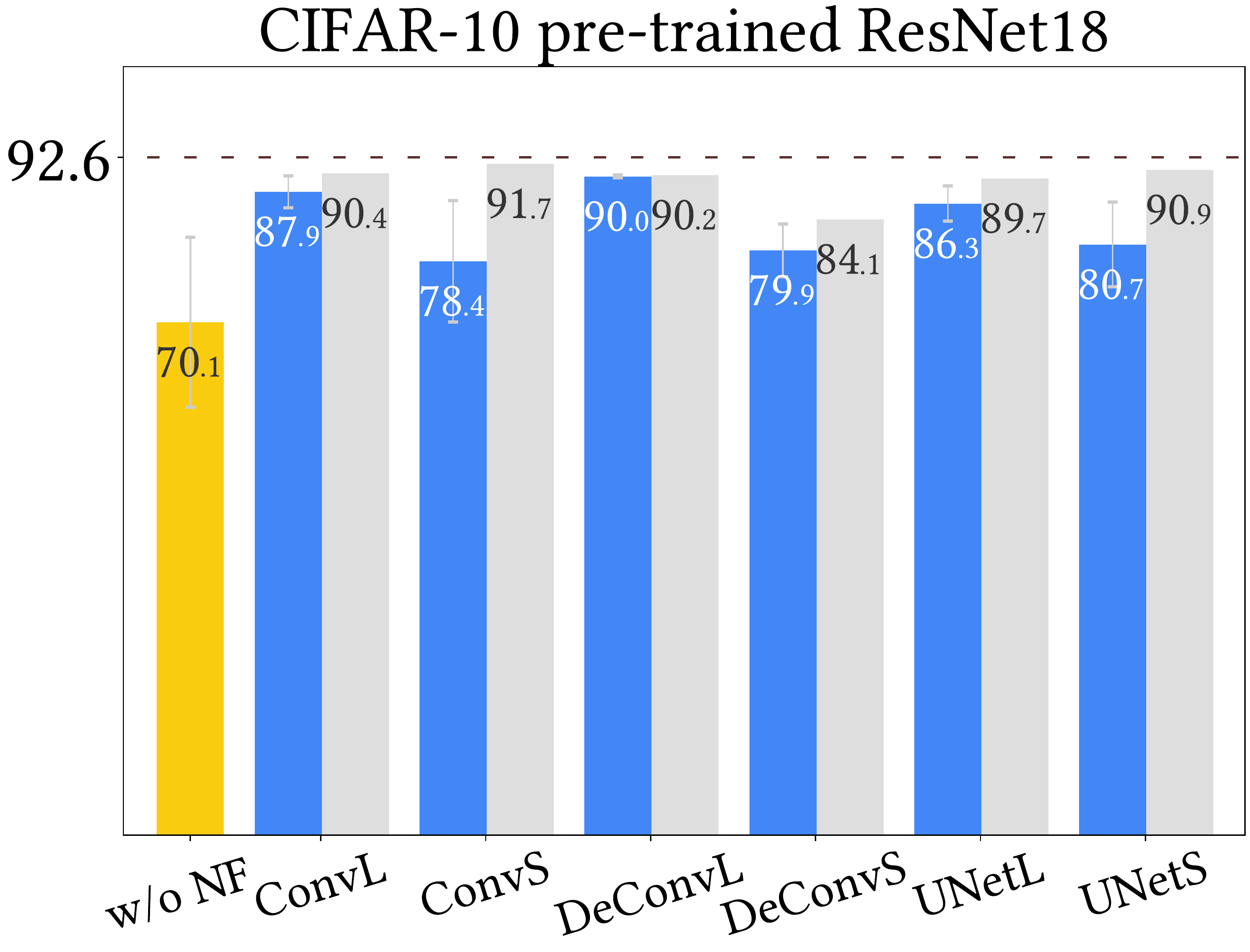}
     \end{subfigure}
     \hfill
     \begin{subfigure}[b]{0.195\textwidth}
         \centering
         \includegraphics[width=\textwidth]{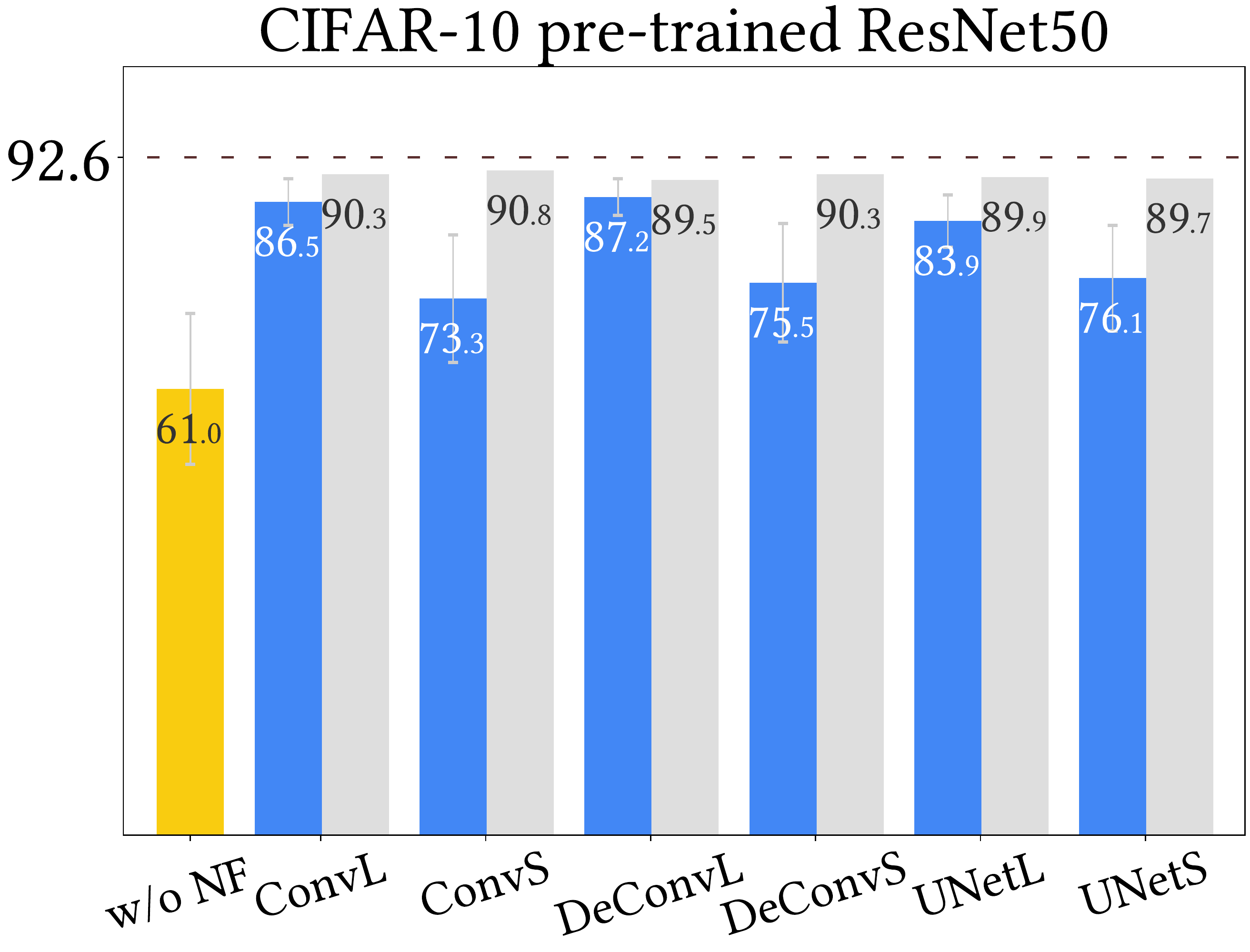}
     \end{subfigure}
     \hfill
     \begin{subfigure}[b]{0.195\textwidth}
         \centering
         \includegraphics[width=\textwidth]{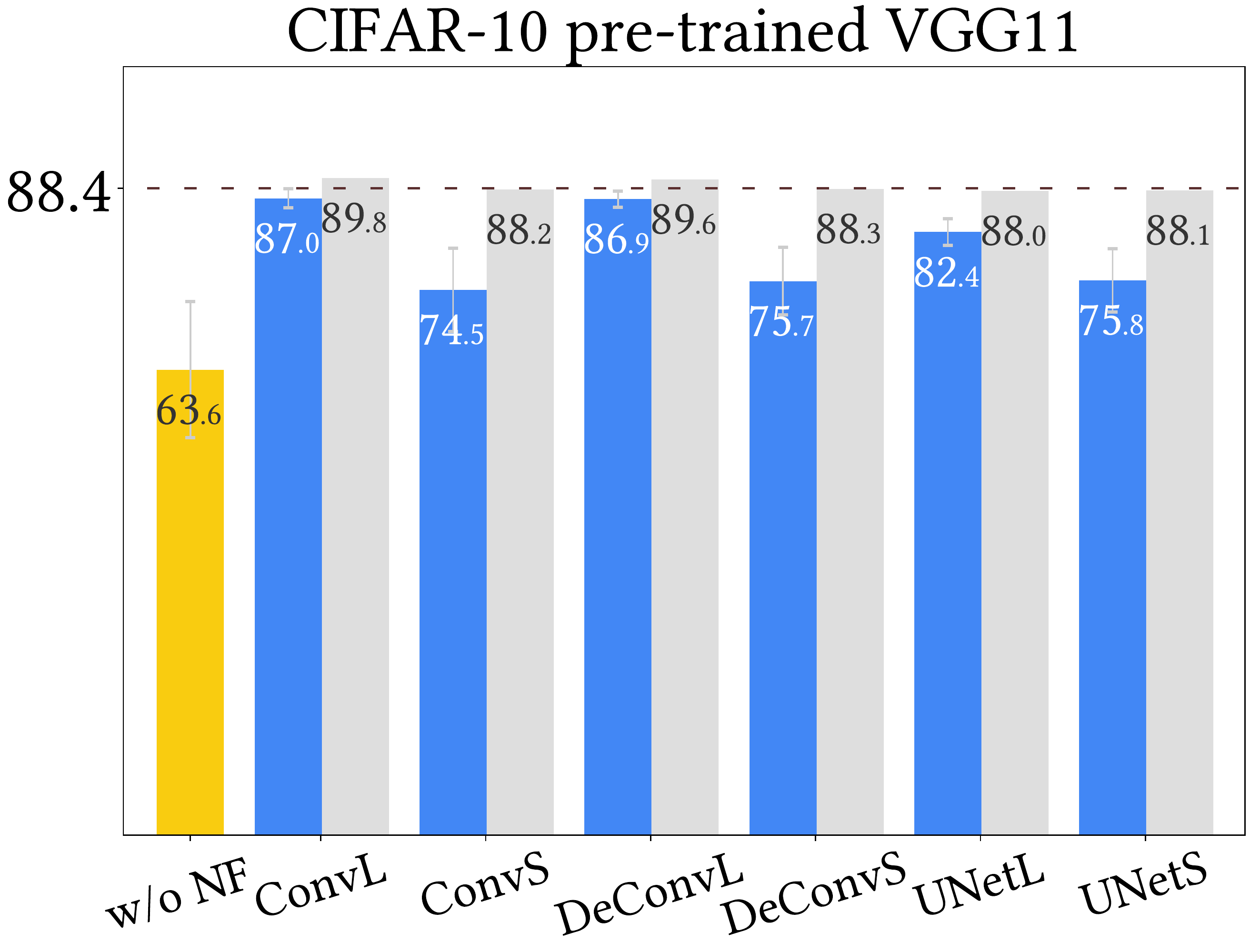}
     \end{subfigure}
     \hfill
     \begin{subfigure}[b]{0.195\textwidth}
         \centering
         \includegraphics[width=\textwidth]{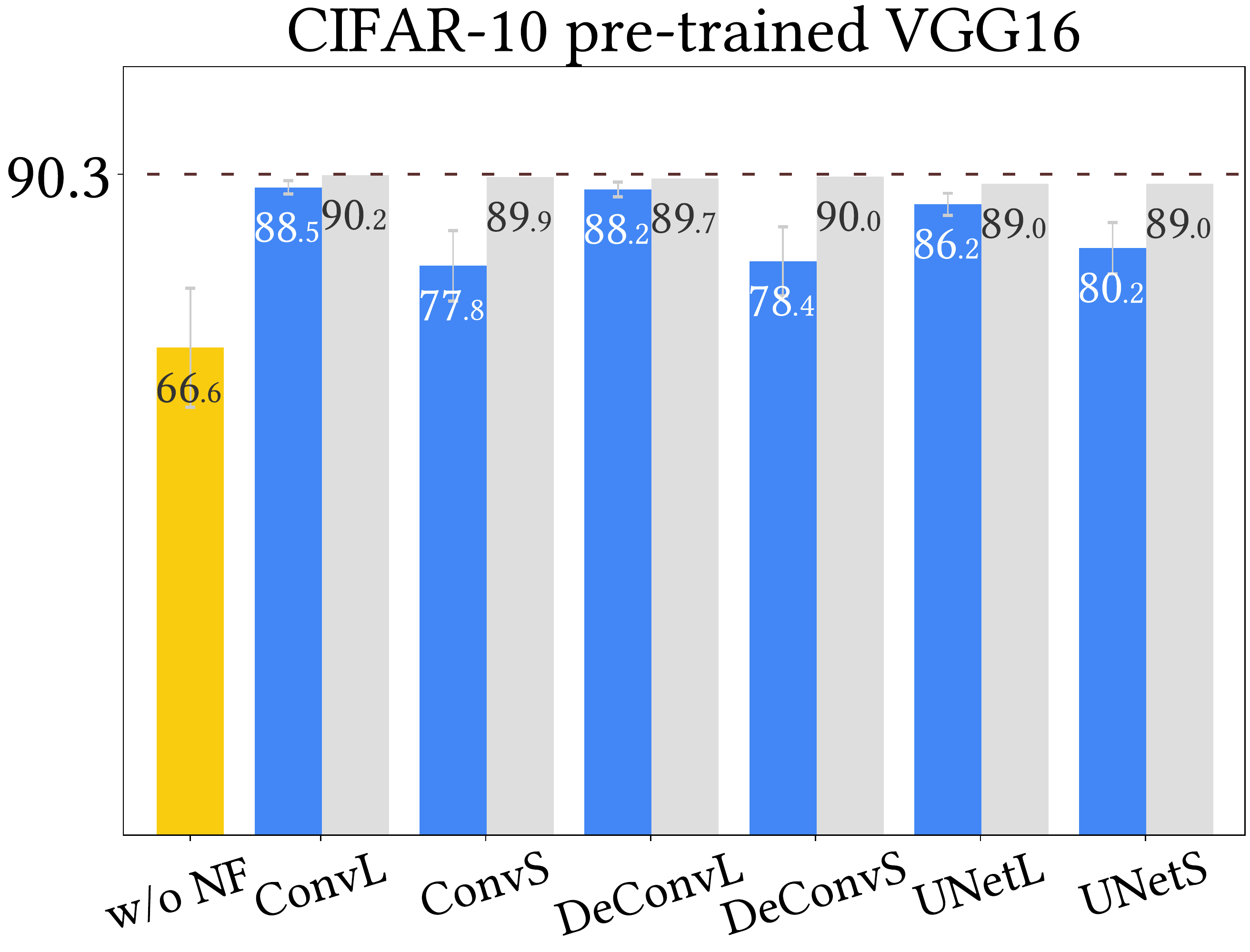}
     \end{subfigure}
     \hfill
     \begin{subfigure}[b]{0.195\textwidth}
         \centering
         \includegraphics[width=\textwidth]{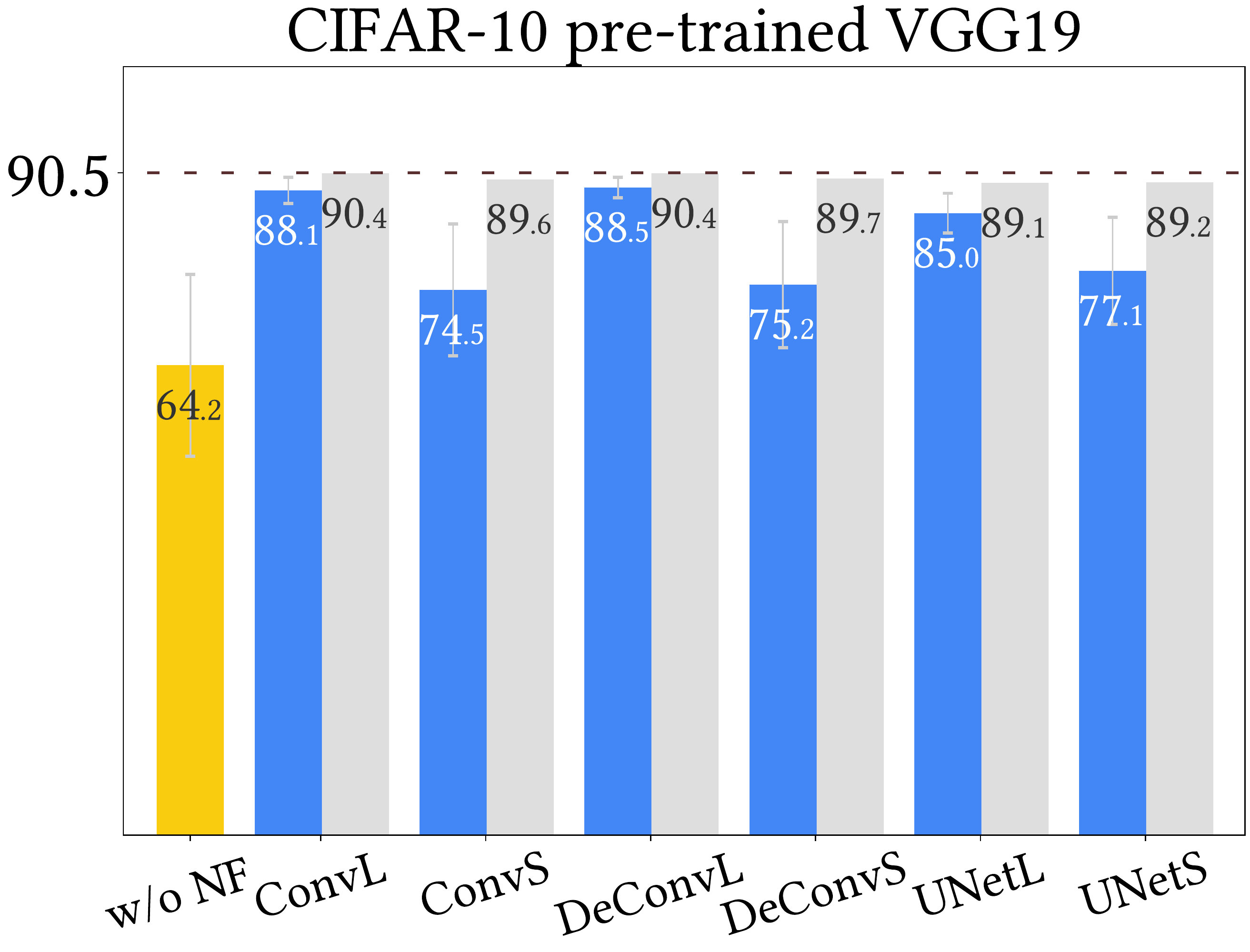}
     \end{subfigure}
     \end{adjustbox}\\
     {\small (b) CIFAR-10, $0.5\%$ Bit Error Rate\\}
        \vspace{-2mm}
        \caption{Experimental results on CIFAR-10}
        \label{fig:cifar10_experimental_results_appendix}
        \vspace{-3.5cm}
\end{figure}

%% file: assets/_tables/relaxed_access/_tables_Appendix_GTSRB.tex
\begin{table*}[ht]
\centering
\caption{Testing accuracy (\%) under 1\% and 0.5\% of random bit error rate on GTSRB.}
\label{table:Appendix_relaxed_access_gtsrb}
\begin{adjustbox}{max width=\linewidth}
\begin{threeparttable}
\begin{tabular}{ c|c|c|ccrr|ccrr }

\toprule[1.5pt]
\multirow{2}{*}{\begin{tabular}{@{}c@{}}Base \\ Model\end{tabular}} & \multirow{2}{*}{NF} & \multirow{2}{*}{CA} &\multicolumn{4}{c|}{\textbf{1\% BER}} & \multicolumn{4}{c}{\textbf{0.5\% BER}}\\
& & & PA & CA (NF) & PA (NF) & RP 
    & PA & CA (NF) & PA (NF) & RP \\
\midrule
\multirow{6}{*}{ResNet18} & ConvL &
\multirow{6}{*}{95.5} & & 95.7 & 91.1 $\pm$ 4.7 & 54.2 & 
& 93.4 & 89.5 $\pm$ 1.9 & 14.3 \\ 
& ConvS &
& & 94.4 & 68.6 $\pm$ ~12 & 31.7 & 
& 94.8  & 87.7 $\pm$ 4.2 & 12.4 \\ 
& DeConvL &
& 36.9 & 95.6 & 91.3 $\pm$ 4.3 & 54.4 & 
75.2 & 95.4 & 93.4 $\pm$ 1.1 & 18.1 \\ 
& DeConvS & 
& $\pm$ 16.0 & 95.7 & 78.1 $\pm$ 9.1 & 41.2 & 
$\pm$ 12.7 & 95.8 & 90.1 $\pm$ 3.3 & 14.9 \\ 
& UNetL &
& & 96.2 & 93.8 $\pm$ 1.0 & 56.9 & 
& 96.2  & 93.5 $\pm$ 1.6 & 18.3 \\ 
& UNetS &
& & 95.9 & 85.1 $\pm$ 6.9 & 48.2 & 
& 95.5 & 91.4 $\pm$ 2.8 & 16.2 \\ 
\midrule

\multirow{6}{*}{ResNet50} & ConvL &
\multirow{6}{*}{95.0} & & 95.6 & 71.6 $\pm$ ~20 & 42.1 & 
& 94.6 & 90.6 $\pm$ 3.7 & 16.6 \\ 
& ConvS &
& & 94.8 & 50.5 $\pm$ ~22 & 21.0 & 
& 95.4 & 84.5 $\pm$ 8.5 & 10.5 \\ 
& DeConvL &
& 29.5 & 94.9  & 71.6 $\pm$ ~21 & 42.0 & 
74.0 & 94.7  & 91.6 $\pm$ 2.9 & 17.6 \\ 
& DeConvS &
& $\pm$ 16.9 & 93.0  & 56.4 $\pm$ ~17 & 26.9 & 
$\pm$ 13.0 & 94.6 & 87.4 $\pm$ 5.9 & 13.5 \\ 
& UNetL &
& & 94.5 & 80.6 $\pm$ ~15 & 51.1 & 
& 96.5 & 93.7 $\pm$ 2.3 & 19.7 \\ 
& UNetS & 
& & 94.7 & 64.7 $\pm$ ~22 & 35.2 & 
& 95.9 & 90.6 $\pm$ 4.8 & 16.7 \\ 
\midrule

\multirow{6}{*}{VGG11} & ConvL & 
\multirow{6}{*}{91.9} & & 94.8 & 85.7 $\pm$ 7.2 & 50.9 & 
& 93.9 & 92.6 $\pm$ 0.7 & 27.7 \\ 
& ConvS & 
& & 91.1 & 62.2 $\pm$ ~11 & 27.3 & 
& 90.9 & 80.5 $\pm$ 3.5 & 15.7 \\ 
& DeConvL & 
& 34.9 & 95.0  & 84.6 $\pm$ 7.6 & 49.7 & 
64.9 & 93.6  & 91.9 $\pm$ 0.6 & 27.1 \\ 
& DeConvS &
& $\pm$ 12.4 & 92.4  & 67.5 $\pm$ ~11 & 32.6 & 
$\pm$ 10.8 & 92.3  & 83.1 $\pm$ 3.7 & 18.2 \\ 
& UNetL & 
& & 92.2 & 83.2 $\pm$ 6.0 & 48.3 & 
& 94.8 & 90.6 $\pm$ 1.7 & 25.7 \\ 
& UNetS & 
& & 94.7 & 73.4 $\pm$ ~10 & 38.5 & 
& 94.6 & 88.9 $\pm$ 2.2 & 24.1 \\ 
\midrule

\multirow{6}{*}{VGG16} & ConvL &
\multirow{6}{*}{95.2} & & 96.3 & 72.4 $\pm$ ~12 & 57.3 & 
& 95.6 & 93.2 $\pm$ 1.8 & 34.4 \\ 
& ConvS &
& & 94.1 & 39.8 $\pm$ ~13 & 24.6 & 
& 94.3 & 82.2 $\pm$ 6.2 & 23.4 \\ 
& DeConvL & 
& 15.1 & 96.4  & 72.0 $\pm$ ~12 & 56.9 & 
58.8 & 95.6 & 93.1 $\pm$ 2.0 & 34.3 \\ 
& DeConvS & 
& $\pm$ 6.8 & 93.8  & 50.9 $\pm$ ~13 & 35.8 & 
$\pm$ 8.9 & 95.1  & 84.0 $\pm$ 5.3 & 25.2 \\ 
& UNetL &
& & 95.8 & 78.6 $\pm$ ~11 & 63.5 & 
& 96.0 & 92.8 $\pm$ 2.0 & 34.0 \\ 
& UNetS & 
& & 94.3 & 63.3 $\pm$ ~14 & 48.1 & 
& 95.4 & 87.8 $\pm$ 3.6 & 29.0 \\ 
\midrule

\multirow{6}{*}{VGG19} & ConvL &
\multirow{6}{*}{95.5} & & 96.0 & 88.3 $\pm$ 7.2 & 51.7 & 
& 95.6 & 93.4 $\pm$ 2.1 & 24.2 \\ 
& ConvS & 
& & 93.8 & 69.0 $\pm$ ~14 & 32.4 & 
& 94.9 & 87.0 $\pm$ 4.4 & 17.8 \\ 
& DeConvL &
& 36.6 & 95.4  & 87.2 $\pm$ 7.5 & 50.6 & 
69.1 & 95.5  & 92.4 $\pm$ 2.2 & 23.3 \\ 
& DeConvS &
& $\pm$ 6.8 & 94.5  & 73.1 $\pm$ ~12 & 36.5 & 
$\pm$ 11.1 & 95.5  & 88.8 $\pm$ 3.7 & 19.7 \\ 
& UNetL & 
& & 95.4 & 88.2 $\pm$ 6.7 & 51.7 & 
& 94.9 & 91.7 $\pm$ 2.5 & 22.6 \\ 
& UNetS &
& & 94.6 & 80.6 $\pm$ 9.0 & 44.1 & 
& 96.5 & 90.8 $\pm$ 3.4 & 21.6 \\ 
\bottomrule[1.5pt]
\end{tabular}
    \begin{tablenotes}
      \small
      \item {\textit{Note}. CA ($\%$): clean accuracy; PA ($\%$): perturbed accuracy; NF: NeuralFuse; and RP: total recovery percentage of PA (NF) vs. PA}
    \end{tablenotes}
\end{threeparttable}
\end{adjustbox}
\end{table*}

%% file: assets/_figures/appendix_relaxed_access/gtsrb.tex
\begin{figure}[ht]
     \centering
     \hfill
     \begin{subfigure}[b]{\textwidth}
         \centering
         \includegraphics[width=\textwidth, trim=-0.2cm -0.1cm 0.2cm 0.1cm, clip]{assets/figures_paths/exp_results/legend.pdf}
     \end{subfigure}
     \hfill
     \begin{subfigure}[b]{0.195\textwidth}
         \centering
         \includegraphics[width=\textwidth]{assets/figures_paths/exp_results/gtsrb_resnet18.pdf}
     \end{subfigure}
     \hfill
     \begin{subfigure}[b]{0.195\textwidth}
         \centering
         \includegraphics[width=\textwidth]{assets/figures_paths/exp_results/gtsrb_resnet50.pdf}
     \end{subfigure}
     \hfill
     \begin{subfigure}[b]{0.195\textwidth}
         \centering
         \includegraphics[width=\textwidth]{assets/figures_paths/exp_results/gtsrb_vgg11.pdf}
     \end{subfigure}
     \hfill
     \begin{subfigure}[b]{0.195\textwidth}
         \centering
         \includegraphics[width=\textwidth]{assets/figures_paths/exp_results/gtsrb_vgg16.pdf}
     \end{subfigure}
     \hfill
     \begin{subfigure}[b]{0.195\textwidth}
         \centering
         \includegraphics[width=\textwidth]{assets/figures_paths/exp_results/gtsrb_vgg19.pdf}
     \end{subfigure}
     {\small (a) GTSRB, $1\%$ Bit Error Rate\\}\vspace{2mm}
     \begin{subfigure}[b]{0.195\textwidth}
         \centering
         \includegraphics[width=\textwidth]{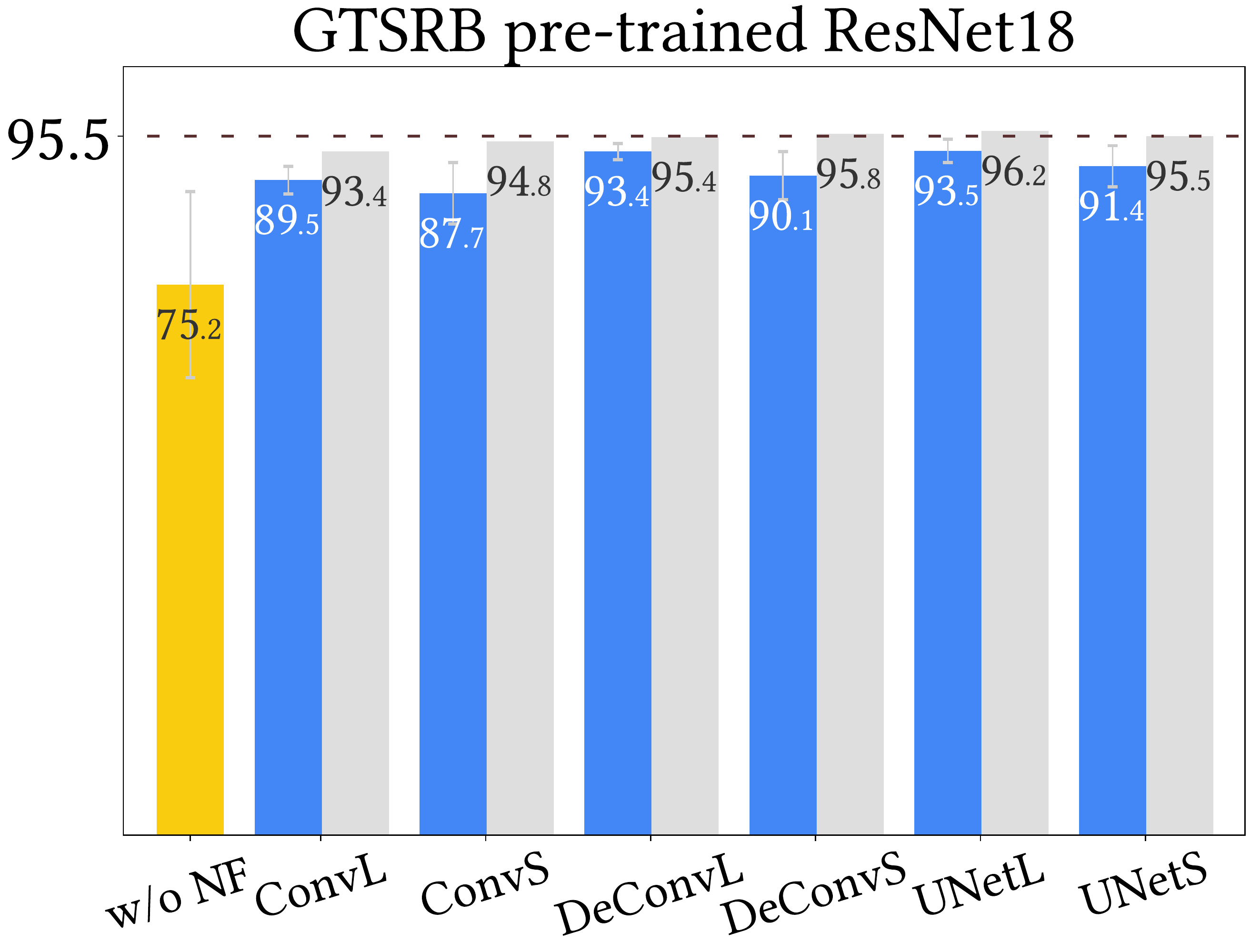}
     \end{subfigure}
     \hfill
     \begin{subfigure}[b]{0.195\textwidth}
         \centering
         \includegraphics[width=\textwidth]{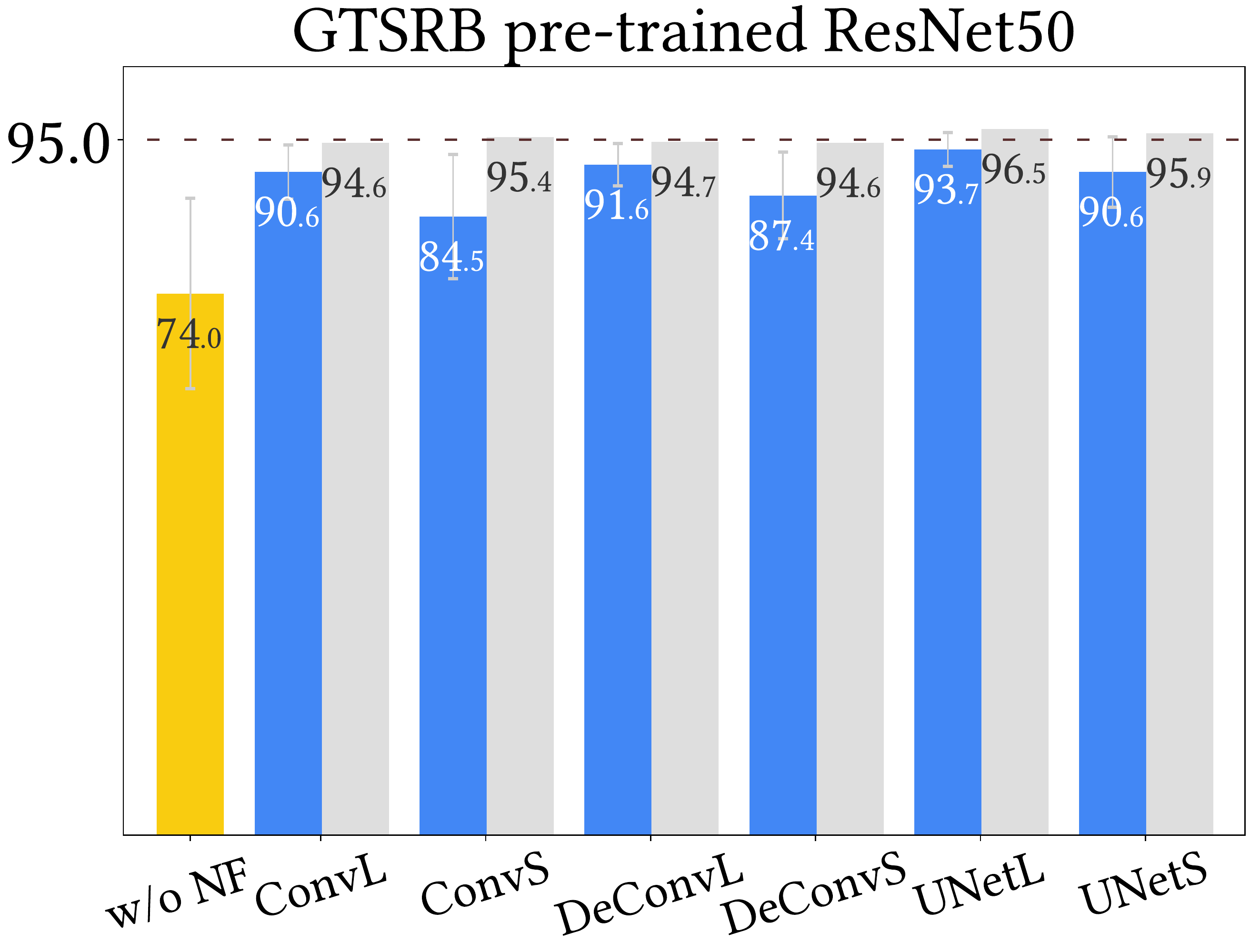}
     \end{subfigure}
     \hfill
     \begin{subfigure}[b]{0.195\textwidth}
         \centering
         \includegraphics[width=\textwidth]{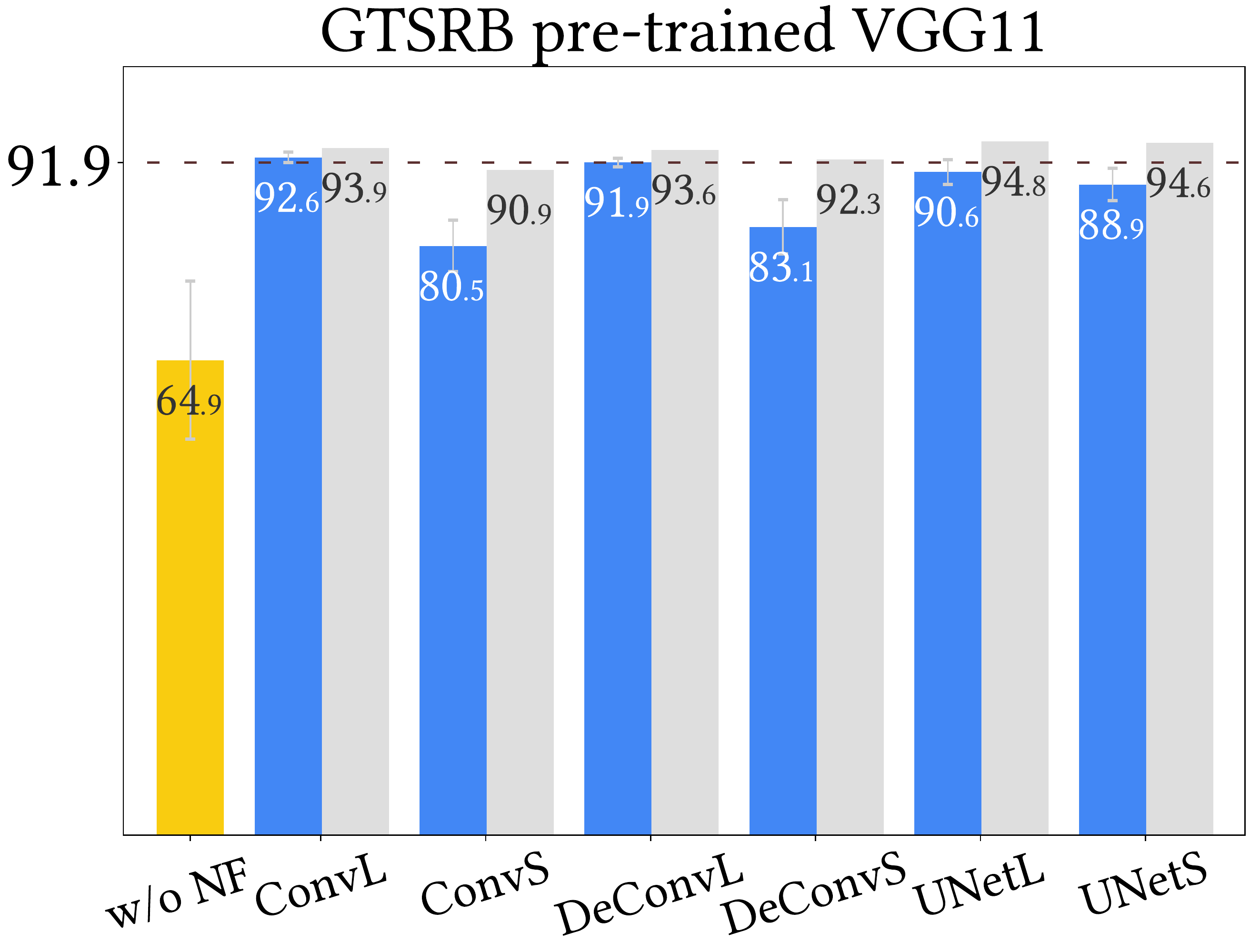}
     \end{subfigure}
     \hfill
     \begin{subfigure}[b]{0.195\textwidth}
         \centering
         \includegraphics[width=\textwidth]{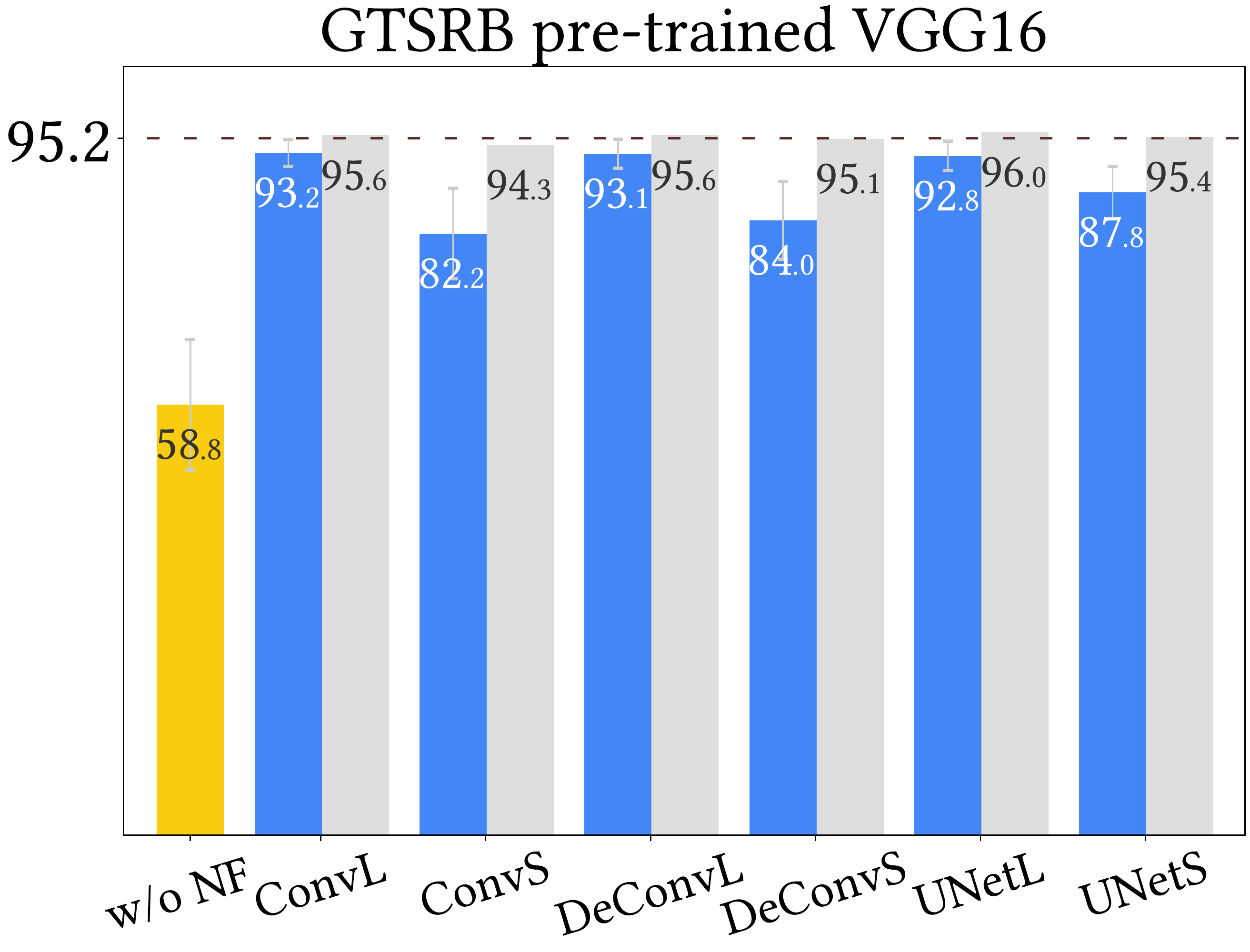}
     \end{subfigure}
     \hfill
     \begin{subfigure}[b]{0.195\textwidth}
         \centering
         \includegraphics[width=\textwidth]{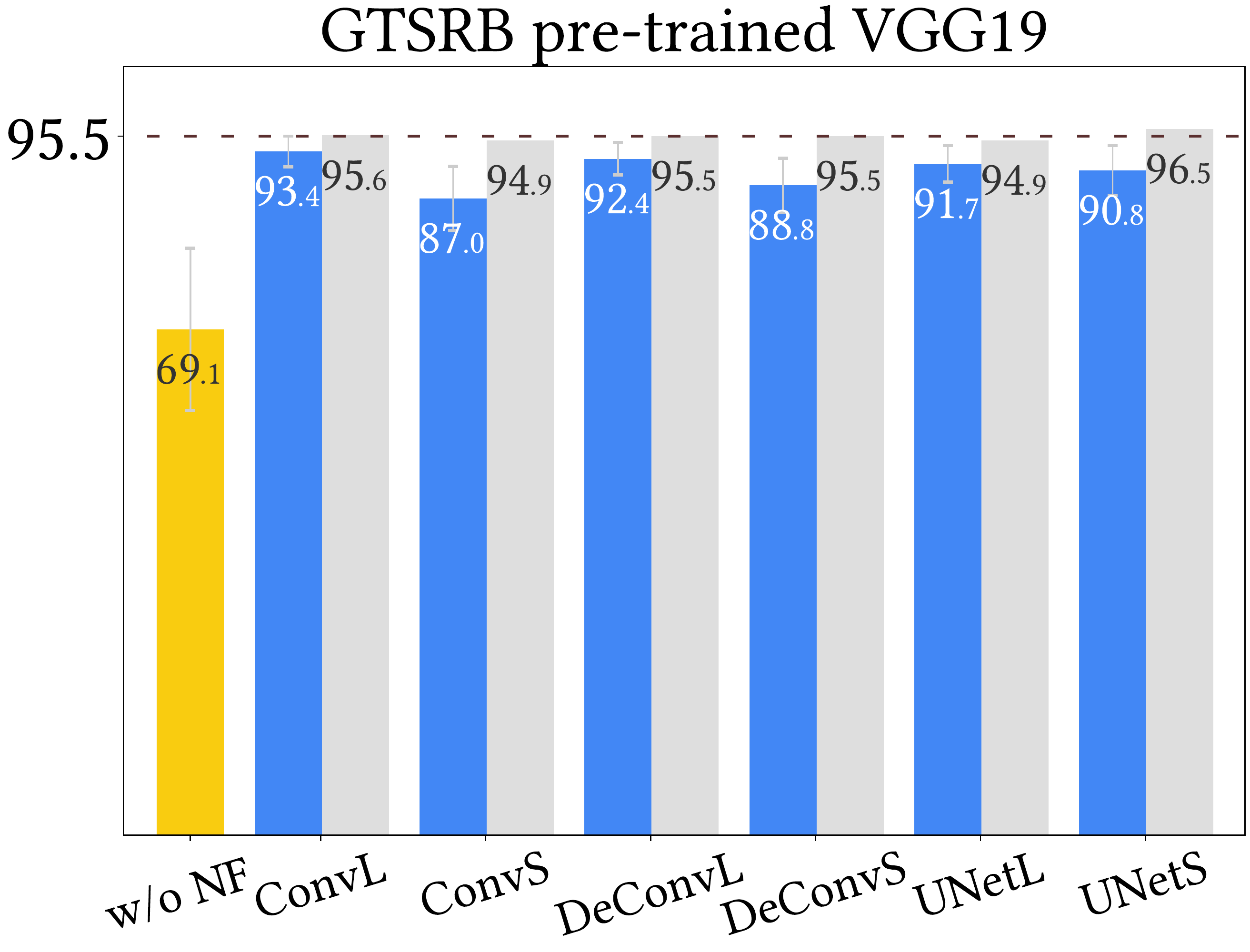}
     \end{subfigure}
     {\small (b) GTSRB, $0.5\%$ Bit Error Rate\\}
        \caption{Experimental results on GTSRB.}
        \label{fig:gtsrb_experimental_results_appendix}
        \vspace{-4.5cm}
\end{figure}

%% file: assets/_tables/relaxed_access/_tables_Appendix_ImageNet10.tex
\begin{table*}[ht]
\centering
\begin{threeparttable}
\caption{Testing accuracy under 0.5\% of random bit error rate on ImageNet-10.}
\label{table:Appendix_relaxed_access_imagenet10}

\begin{tabular}{ c|c|c|ccrr }
\toprule[1.5pt] 
\multirow{2}{*}{\begin{tabular}{@{}c@{}}Base \\ Model\end{tabular}} & \multirow{2}{*}{NF} & \multirow{2}{*}{CA} &\multicolumn{4}{c}{\textbf{0.5\% BER}}  \\
& & & PA & CA (NF) & PA (NF) & RP  \\
\midrule 

\multirow{6}{*}{ResNet18} & ConvL & \multirow{6}{*}{92.2}  & \multirow{6}{*}{72.3 $\pm$ 7.0} & 94.0  & 88.0 $\pm$ 2.0 & 15.7 \\
& ConvS & 
& & 91.8 & 83.6 $\pm$ 4.1 & 11.3 \\
& DeConvL & 
& & 94.0 & 89.2 $\pm$ 1.3 & 16.9 \\
& DeConvS & 
& & 92.8 & 87.5 $\pm$ 2.3 & 15.2 \\
& UNetL & 
& & 94.0 & 88.1 $\pm$ 1.4 & 15.8 \\
& UNetS & 
& & 93.2 & 86.4 $\pm$ 2.2 & 14.1 \\
\midrule

\multirow{6}{*}{ResNet50}
& ConvL & 
\multirow{6}{*}{89.8} & \multirow{6}{*}{39.4 $\pm$ ~11} & 92.2 & 80.0 $\pm$ 5.8 & 40.6 \\
& ConvS & 
& & 91.8 & 65.0 $\pm$ ~11 & 25.6 \\
& DeConvL & 
& & 93.0 & 79.4 $\pm$ 5.9 & 40.0 \\
& DeConvS & 
& & 93.2 & 70.9 $\pm$ 9.1 & 31.5 \\
& UNetL & 
& & 92.2 & 80.5 $\pm$ 5.8 & 41.1 \\
& UNetS & 
& & 92.4 & 73.6 $\pm$ 8.9 & 34.2 \\
\midrule

\multirow{6}{*}{VGG11}
& ConvL & 
\multirow{6}{*}{91.6} & \multirow{6}{*}{47.8 $\pm$ ~13} & 92.0 & 86.1 $\pm$ 3.7 & 38.3 \\
& ConvS & 
& & 89.4 & 66.4 $\pm$ 7.1 & 18.6 \\
& DeConvL & 
& & 91.0  & 86.0 $\pm$ 3.0 & 38.2 \\
& DeConvS & 
& & 89.0  & 72.5 $\pm$ 7.8 & 24.7 \\
& UNetL & 
& & 92.4 & 83.0 $\pm$ 3.5 & 35.2 \\
& UNetS & 
& & 86.2 & 73.5 $\pm$ 6.0 & 25.7 \\
\midrule

\multirow{6}{*}{VGG16}
& ConvL & 
\multirow{6}{*}{94.6} & \multirow{6}{*}{38.4 $\pm$ ~17} & 90.8 & 77.1 $\pm$ ~11 & 38.7 \\
& ConvS & 
& & 90.2 & 60.2 $\pm$ ~14 & 21.8 \\
& DeConvL & 
& & 91.2 & 77.2 $\pm$ ~11 & 38.8 \\
& DeConvS & 
& & 90.0 & 62.3 $\pm$ ~14 & 23.9 \\
& UNetL & 
& & 90.6 & 81.1 $\pm$ 5.9 & 42.7 \\
& UNetS & 
& & 86.4 & 72.3 $\pm$ 8.8 & 33.9 \\
\midrule

\multirow{6}{*}{VGG19}
& ConvL & 
\multirow{6}{*}{92.4} & \multirow{6}{*}{37.2 $\pm$ ~11} & 91.4 & 75.5 $\pm$ 8.8 & 38.3 \\
& ConvS & 
& & 88.8  & 56.5 $\pm$ ~13 & 19.3 \\
& DeConvL & 
& & 91.0  & 75.9 $\pm$ 8.9 & 38.7 \\
& DeConvS &
& & 88.8  & 64.0 $\pm$ ~11 & 26.8 \\
& UNetL & 
& & 89.4  & 77.9 $\pm$ 6.1 & 40.7 \\
& UNetS & 
& & 87.6  & 65.9 $\pm$ ~10 & 28.7 \\
\bottomrule[1.5pt]
\end{tabular}
    \begin{tablenotes}
      \small
      \item {\textit{Note}. CA ($\%$): clean accuracy; PA ($\%$): perturbed accuracy; NF: NeuralFuse; and RP: total recovery percentage of PA (NF) vs. PA}
    \end{tablenotes}
\end{threeparttable}
\end{table*}

%% file: assets/_figures/appendix_relaxed_access/imagenet10.tex
\begin{figure}[ht]
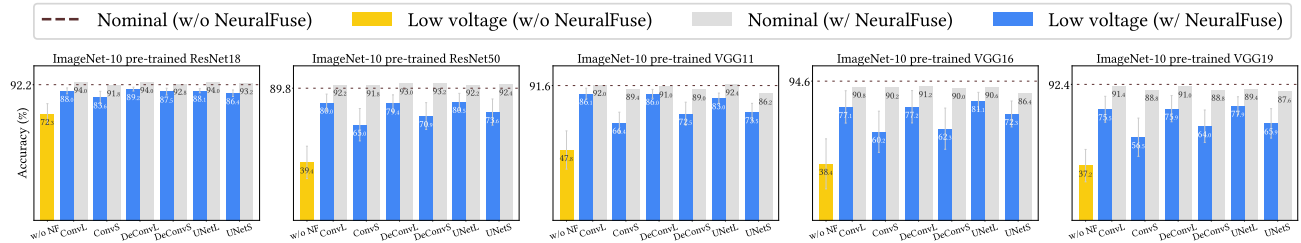

     \centering
     \hfill
     \begin{subfigure}[b]{\textwidth}
         \centering
         \includegraphics[width=\textwidth, trim=-0.2cm -0.1cm 0.2cm 0.1cm, clip]{assets/figures_paths/exp_results/legend.pdf}
     \end{subfigure}
     \hfill
     \begin{subfigure}[b]{0.195\textwidth}
         \centering
         \includegraphics[width=\textwidth]{assets/figures_paths/exp_results/imagenet10_resnet18.pdf}
     \end{subfigure}
     \hfill
     \begin{subfigure}[b]{0.195\textwidth}
         \centering
         \includegraphics[width=\textwidth]{assets/figures_paths/exp_results/imagenet10_resnet50.pdf}
     \end{subfigure}
     \hfill
     \begin{subfigure}[b]{0.195\textwidth}
         \centering
         \includegraphics[width=\textwidth]{assets/figures_paths/exp_results/imagenet10_vgg11.pdf}
     \end{subfigure}
     \hfill
     \begin{subfigure}[b]{0.195\textwidth}
         \centering
         \includegraphics[width=\textwidth]{assets/figures_paths/exp_results/imagenet10_vgg16.pdf}
     \end{subfigure}
     \hfill
     \begin{subfigure}[b]{0.195\textwidth}
         \centering
         \includegraphics[width=\textwidth]{assets/figures_paths/exp_results/imagenet10_vgg19.pdf}
     \end{subfigure}
        \caption{Experimental results on ImageNet-10, $0.5\%$ Bit Error Rate.}
        \label{fig:imagenet10_experimental_results_appendix}
        \vspace{-4.5cm}
\end{figure}

%% file: assets/_tables/relaxed_access/_tables_Appendix_CIFAR100.tex
\begin{table*}[ht]
\centering
\caption{Testing accuracy (\%) under 1\%, 0.5\% and 0.35\% of random bit error rate on CIFAR-100.}
\label{table:Appendix_relaxed_access_cifar100}
\begin{adjustbox}{max width=\linewidth}
\begin{threeparttable}
\begin{tabular}{ c|c|c|ccrr|ccrr|ccrr }

\toprule[1.5pt]
\multirow{2}{*}{\begin{tabular}{@{}c@{}}Base \\ Model\end{tabular}} & \multirow{2}{*}{NF} & \multirow{2}{*}{C.A.} & \multicolumn{4}{c|}{\textbf{1\% BER}} & \multicolumn{4}{c|}{\textbf{0.5\% BER}} & \multicolumn{4}{c}{\textbf{0.35\% BER}}\\
& & & PA & CA (NF) & PA (NF) & RP
    & PA & CA (NF) & PA (NF) & RP
    & PA & CA (NF) & PA (NF) & RP \\
\midrule
\multirow{6}{*}{ResNet18} & ConvL &
\multirow{6}{*}{73.7} & & 54.8 & 11.0 $\pm$ 7.7 & 6.4 & 
& 65.2 & 39.0 $\pm$ 7.1 & 18.1 & 
& 69.4  & 42.9 $\pm$ 6.2 & 11.4 \\ 
& ConvS &
& & 49.7 & 4.2 $\pm$ 2.2 & -0.4 & 
& 70.0  & 24.5 $\pm$ 7.6 & 3.6 & 
& 72.1 & 35.1 $\pm$ 7.3 & 3.7 \\ 
& DeConvL &
& 4.6 & 55.2 & 11.9 $\pm$ 8.2 & 7.3 & 
20.9 & 66.3 & 38.2 $\pm$ 6.9 & 17.3 & 
31.4 & 69.2 & 42.9 $\pm$ 5.5 & 11.4 \\ 
& DeConvS & 
& $\pm$ 2.9 & 32.7 & 4.0 $\pm$ 2.2 & -0.6 & 
$\pm$ 7.4 & 68.2  & 25.9 $\pm$ 6.8 & 5 & 
$\pm$ 7.6 & 71.6 & 35.8 $\pm$ 5.5 & 4.4 \\ 
& UNetL &
& & 50.6 & 14.5 $\pm$ 8.9 & 10.0 & 
& 66.2  & 40.1 $\pm$ 6.4 & 19.2 & 
& 70.3 & 46.3 $\pm$ 5.5 & 14.9 \\ 
& UNetS &
& & 26.8 & 4.6 $\pm$ 2.5 & -0.0 & 
& 67.1 & 28.8 $\pm$ 6.8 & 7.9 & 
& 70.9 & 38.3 $\pm$ 6.4 & 6.9 \\ 
\midrule

\multirow{6}{*}{ResNet50} & ConvL &
\multirow{6}{*}{73.5} & & 63.5 & 3.2 $\pm$ 1.7 & 0.1 & 
& 68.4 & 28.8 $\pm$ 6.7 & 7.6 & 
& 72.0 & 40.8 $\pm$ 7.5 & 5.1 \\ 
& ConvS &
& & 65.5  & 3.2 $\pm$ 1.6 & 0.1 & 
& 71.9 & 23.1 $\pm$ 6.9 & 1.9 & 
& 73.0 & 37.4 $\pm$ 8.0 & 1.7 \\ 
& DeConvL &
& 3.0 & 59.6  & 3.2 $\pm$ 1.7 & 0.2 & 
21.3 & 68.1  & 28.6 $\pm$ 7.0 & 7.4 & 
35.7 & 71.7 & 41.7 $\pm$ 7.7 & 6.1 \\ 
& DeConvS &
& $\pm$ 1.8 & 61.1  & 3.2 $\pm$ 1.7 & 0.1 & 
$\pm$ 7.0 & 70.3  & 25.0 $\pm$ 6.7 & 3.7 & 
$\pm$ 8.6 & 72.8 & 38.9 $\pm$ 7.9 & 3.3 \\ 
& UNetL &
& & 39.0 & 5.0 $\pm$ 1.7 & 1.9 & 
& 66.6 & 36.5 $\pm$ 6.2 & 15.3 & 
& 70.8 & 45.3 $\pm$ 6.7 & 9.6 \\ 
& UNetS & 
& & 47.7 & 3.4 $\pm$ 1.8 & 0.3 & 
& 69.1 & 26.1 $\pm$ 6.6 & 4.8 & 
& 72.6 & 39.6 $\pm$ 7.8 & 3.9 \\ 
\midrule

\multirow{6}{*}{VGG11} & ConvL & 
\multirow{6}{*}{64.8} & & 58.3 & 19.7 $\pm$ ~11 & 11.5 & 
& 63.1 & 38.8 $\pm$ 9.3 & 15.0 & 
& 63.9 & 42.4 $\pm$ 9.0 & 11.1 \\ 
& ConvS & 
& & 56.6 & 10.4 $\pm$ 7.4 & 2.2 & 
& 62.7 & 27.9 $\pm$ ~10 & 4.0 & 
& 63.9 & 41.8 $\pm$ 8.3 & 10.5 \\ 
& DeConvL & 
& 8.2 & 60.3  & 21.2 $\pm$ ~11 & 13.0 & 
23.9 & 63.9  & 40.0 $\pm$ 9.0 & 16.2 & 
31.3 & 64.0  & 42.8 $\pm$ 9.1 & 11.5 \\ 
& DeConvS &
& $\pm$ 5.7 & 58.3  & 11.8 $\pm$ 7.9 & 3.5 & 
$\pm$ 9.4 & 61.9  & 29.8 $\pm$ 9.9 & 5.9 & 
$\pm$ ~10 & 63.5  & 36.1 $\pm$ ~10 & 4.8 \\ 
& UNetL & 
& & 51.1 & 22.1 $\pm$ 8.2 & 13.9 & 
& 61.8 & 37.8 $\pm$ 9.0 & 13.9 & 
& 63.5 & 40.9 $\pm$ 9.3 & 9.6 \\ 
& UNetS & 
& & 51.9 & 13.1 $\pm$ 7.9 & 4.9 & 
& 61.7 & 29.8 $\pm$ 9.7 & 6.0 & 
& 63.8 & 35.7 $\pm$ 9.9 & 4.5 \\ 
\midrule

\multirow{6}{*}{VGG16} & ConvL &
\multirow{6}{*}{67.8} & & 51.4 & 19.2 $\pm$ 6.0 & 12.6 & 
& 61.8 & 41.1 $\pm$ 5.6 & 18.7 & 
& 64.9 & 44.9 $\pm$ 5.3 & 13.8 \\ 
& ConvS &
& & 44.3 & 6.7 $\pm$ 2.3 & 0.1 & 
& 63.8 & 27.5 $\pm$ 6.8 & 5.1 & 
& 66.0 & 36.3 $\pm$ 6.1 & 5.1 \\ 
& DeConvL & 
& 7.0 & 53.1  & 20.8 $\pm$ 6.2 & 14.2 & 
22.4 & 62.8 & 42.1 $\pm$ 5.5 & 19.8 & 
31.1 & 65.0 & 46.6 $\pm$ 5.2 & 15.5 \\ 
& DeConvS & 
& $\pm$ 3.5 & 23.5  & 4.8 $\pm$ 1.7 & -1.8 & 
$\pm$ 7.0 & 62.1  & 29.9 $\pm$ 6.7 & 7.5 & 
$\pm$ 7.2 & 64.9 & 38.1 $\pm$ 6.3 & 7.0 \\ 
& UNetL &
& & 50.2 & 25.3 $\pm$ 1.7 & 18.7 & 
& 61.7 & 41.3 $\pm$ 5.0 & 18.9 & 
& 64.8 & 46.8 $\pm$ 4.6 & 15.7 \\ 
& UNetS & 
& & 27.7 & 9.9 $\pm$ 2.1 & 3.3 & 
& 61.6 & 31.3 $\pm$ 6.3 & 8.9 & 
& 65.0 & 39.8 $\pm$ 5.9 & 8.7 \\ 
\midrule

\multirow{6}{*}{VGG19} & ConvL &
\multirow{6}{*}{67.8} & & 59.4 & 29.2 $\pm$ 8.1 & 18.6 & 
& 65.6 & 46.5 $\pm$ 6.8 & 12.5 & 
& 66.9 & 49.2 $\pm$ 7.4 & 7.0 \\ 
& ConvS & 
& & 63.7 & 14.4 $\pm$ 5.1 & 3.8 & 
& 66.6 & 38.3 $\pm$ 6.8 & 4.2 & 
& 67.7  & 45.3 $\pm$ 8.5 & 3.2 \\ 
& DeConvL &
& 10.6 & 60.1  & 29.6 $\pm$ 8.5 & 19.0 & 
34.0 & 65.7  & 46.9 $\pm$ 7.1 & 12.9 & 
42.1 & 67.3  & 49.8 $\pm$ 7.6 & 7.6 \\ 
& DeConvS &
& $\pm$ 4.3 & 60.9  & 16.1 $\pm$ 6.0 & 5.6 & 
$\pm$ 9.6 & 66.5  & 39.0 $\pm$ 3.7 & 5.0 & 
$\pm$ 9.4 & 67.7  & 45.7 $\pm$ 8.4 & 3.6 \\ 
& UNetL & 
& & 58.7 & 30.2 $\pm$ 8.2 & 19.6 & 
& 65.5  & 46.9 $\pm$ 6.5 & 12.9 & 
& 67.4  & 50.0 $\pm$ 7.5 & 7.9 \\ 
& UNetS &
& & 59.1 & 18.0 $\pm$ 6.2 & 7.4 & 
& 66.3 & 40.1 $\pm$ 8.0 & 6.1 & 
& 67.5  & 46.6 $\pm$ 8.4 & 4.5 \\ 
\bottomrule[1.5pt]
\end{tabular}
    \begin{tablenotes}
      \item {\textit{Note}. CA ($\%$): clean accuracy; PA ($\%$): perturbed accuracy; NF: NeuralFuse; and RP: total recovery percentage of PA (NF) vs. PA}
    \end{tablenotes}
\end{threeparttable}
\end{adjustbox}
\end{table*}

%% file: assets/_figures/appendix_relaxed_access/cifar100.tex
\begin{figure}[ht]
     \vspace{-5mm}
     \centering
     \hfill
     \begin{subfigure}[b]{\textwidth}
         \centering
         \includegraphics[width=\textwidth, trim=-0.2cm -0.1cm 0.2cm 0.1cm, clip]{assets/figures_paths/exp_results/legend.pdf}
     \end{subfigure}
     \hfill
     \begin{adjustbox}{max width=.96\linewidth}
     \begin{subfigure}[b]{0.195\textwidth}
         \centering
         \includegraphics[width=\textwidth]{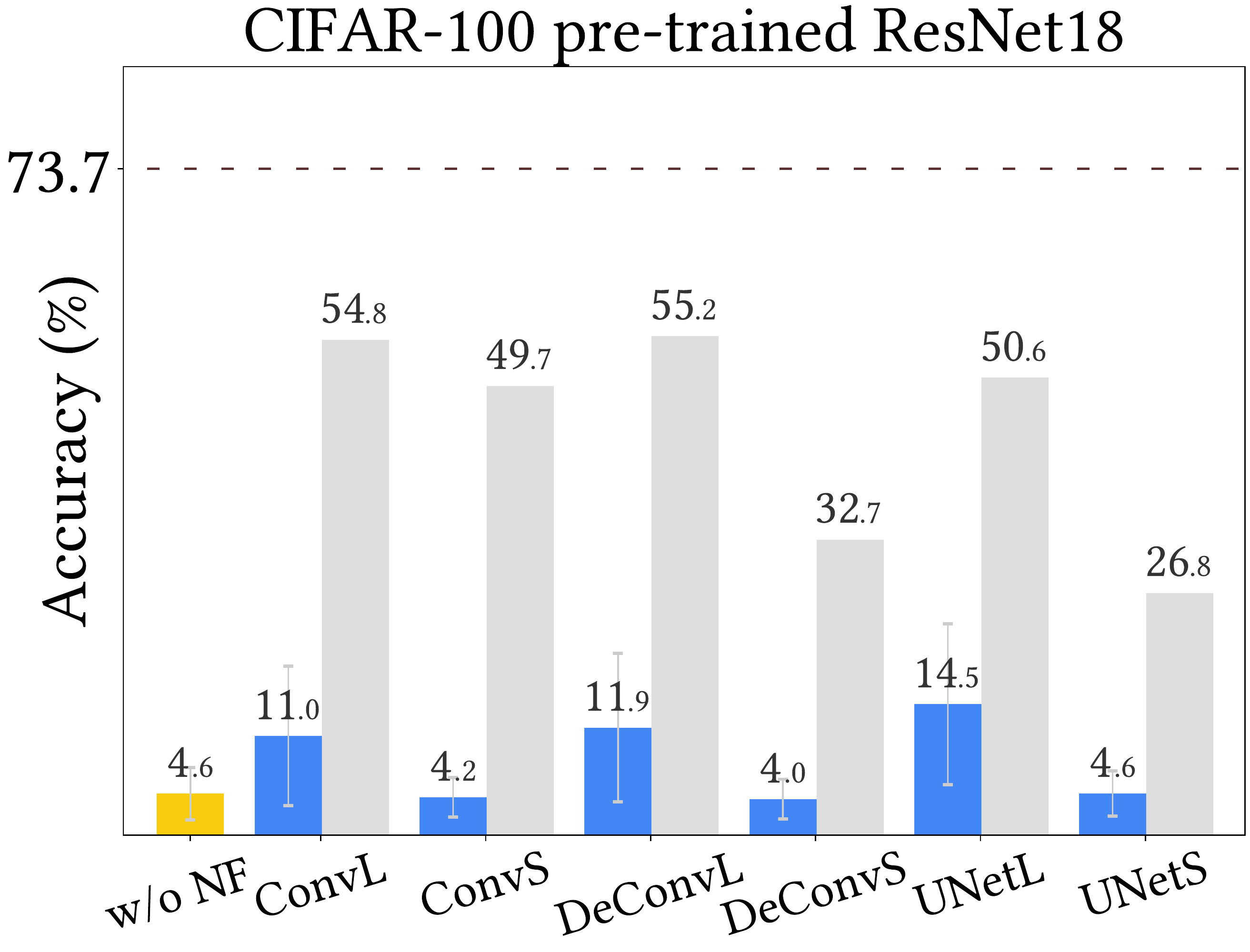}
     \end{subfigure}
     \hfill
     \begin{subfigure}[b]{0.195\textwidth}
         \centering
         \includegraphics[width=\textwidth]{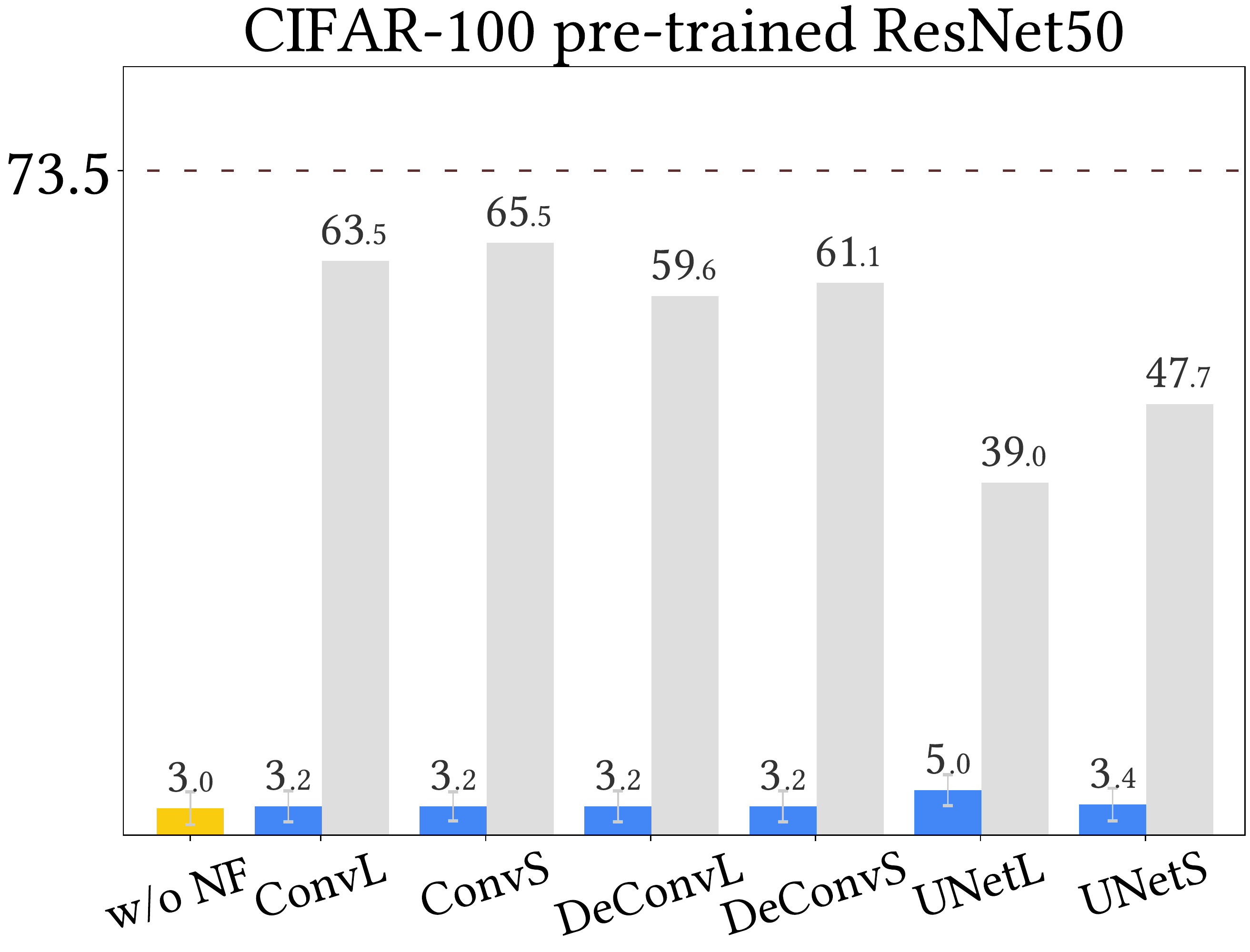}
     \end{subfigure}
     \hfill
     \begin{subfigure}[b]{0.195\textwidth}
         \centering
         \includegraphics[width=\textwidth]{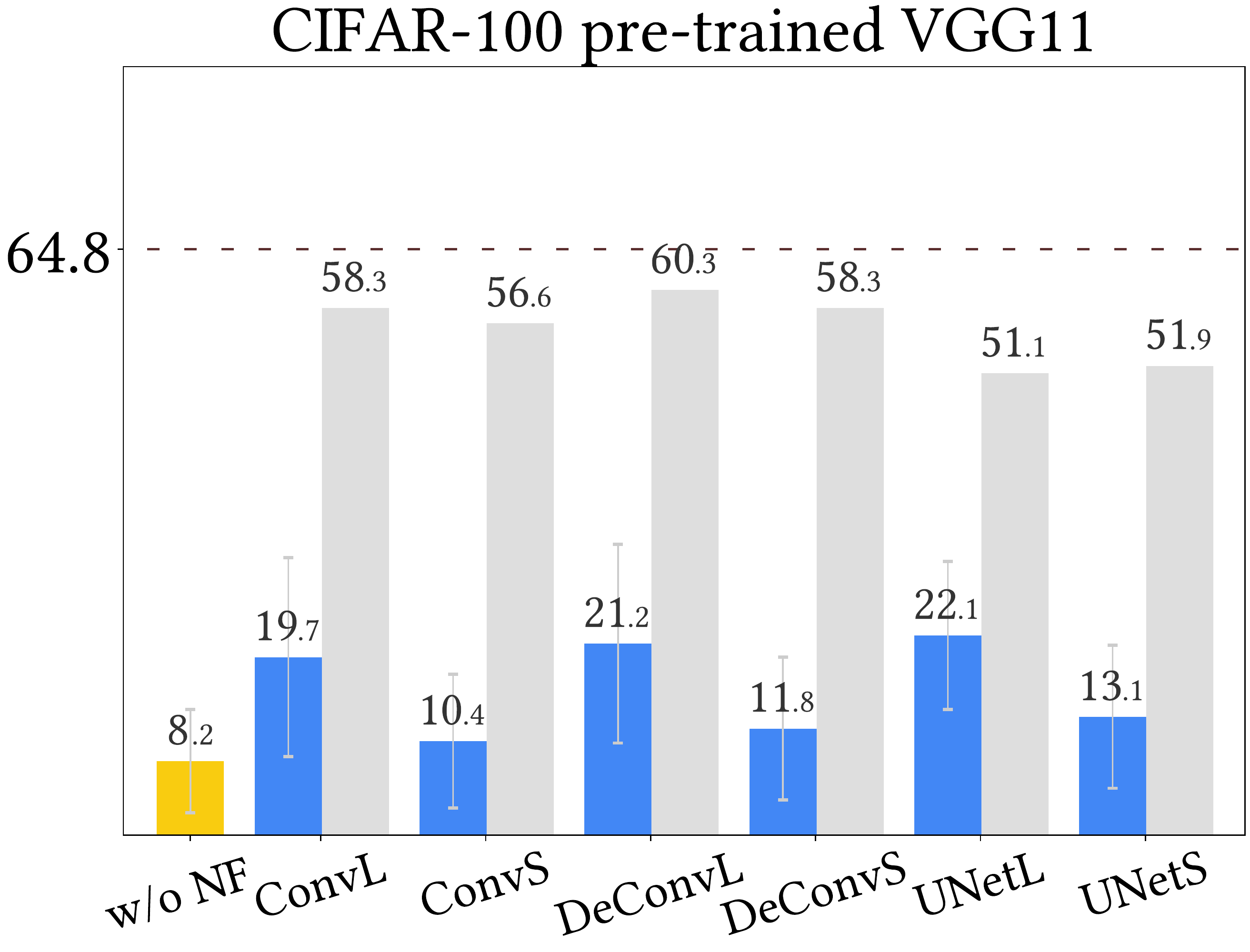}
     \end{subfigure}
     \hfill
     \begin{subfigure}[b]{0.195\textwidth}
         \centering
         \includegraphics[width=\textwidth]{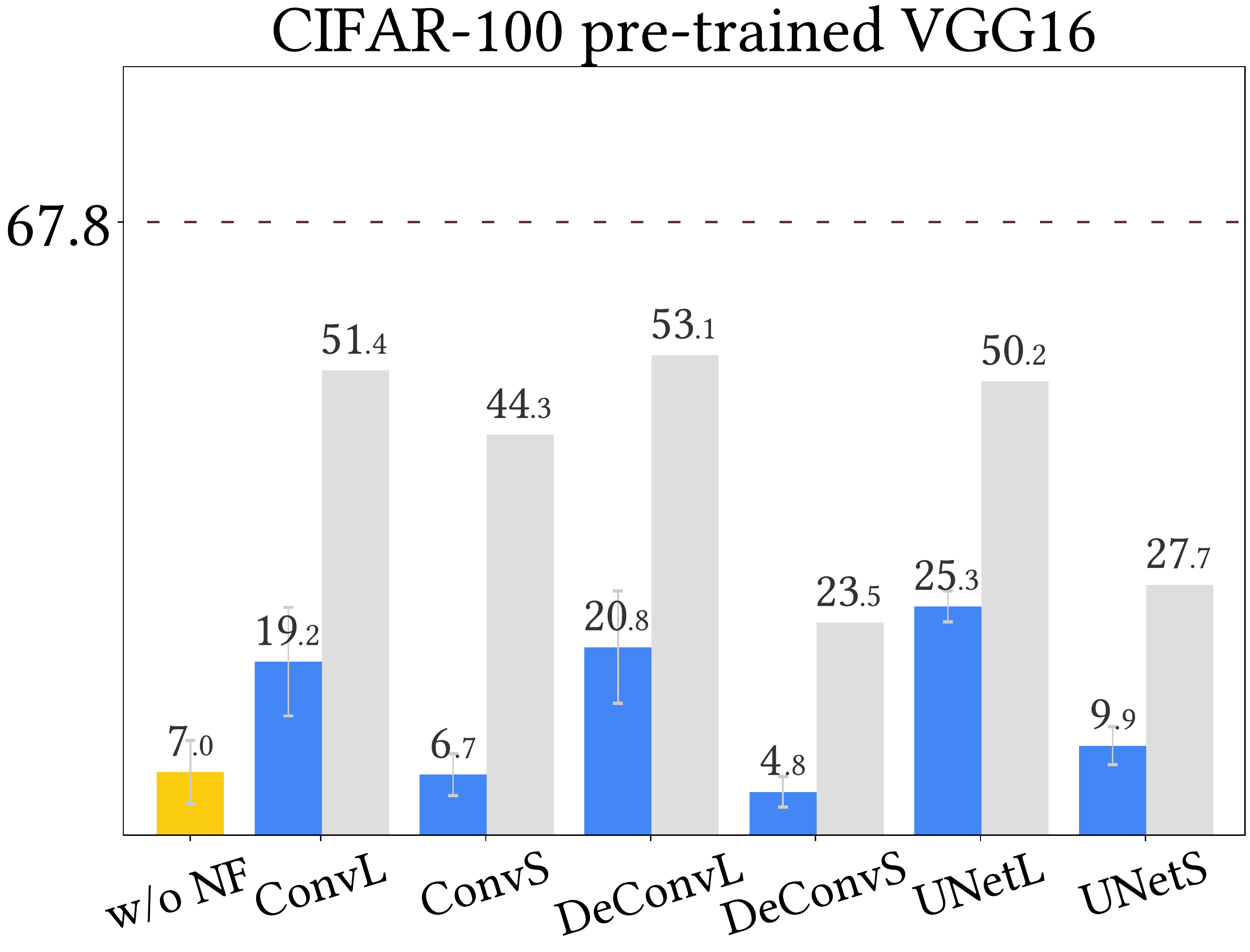}
     \end{subfigure}
     \hfill
     \begin{subfigure}[b]{0.195\textwidth}
         \centering
         \includegraphics[width=\textwidth]{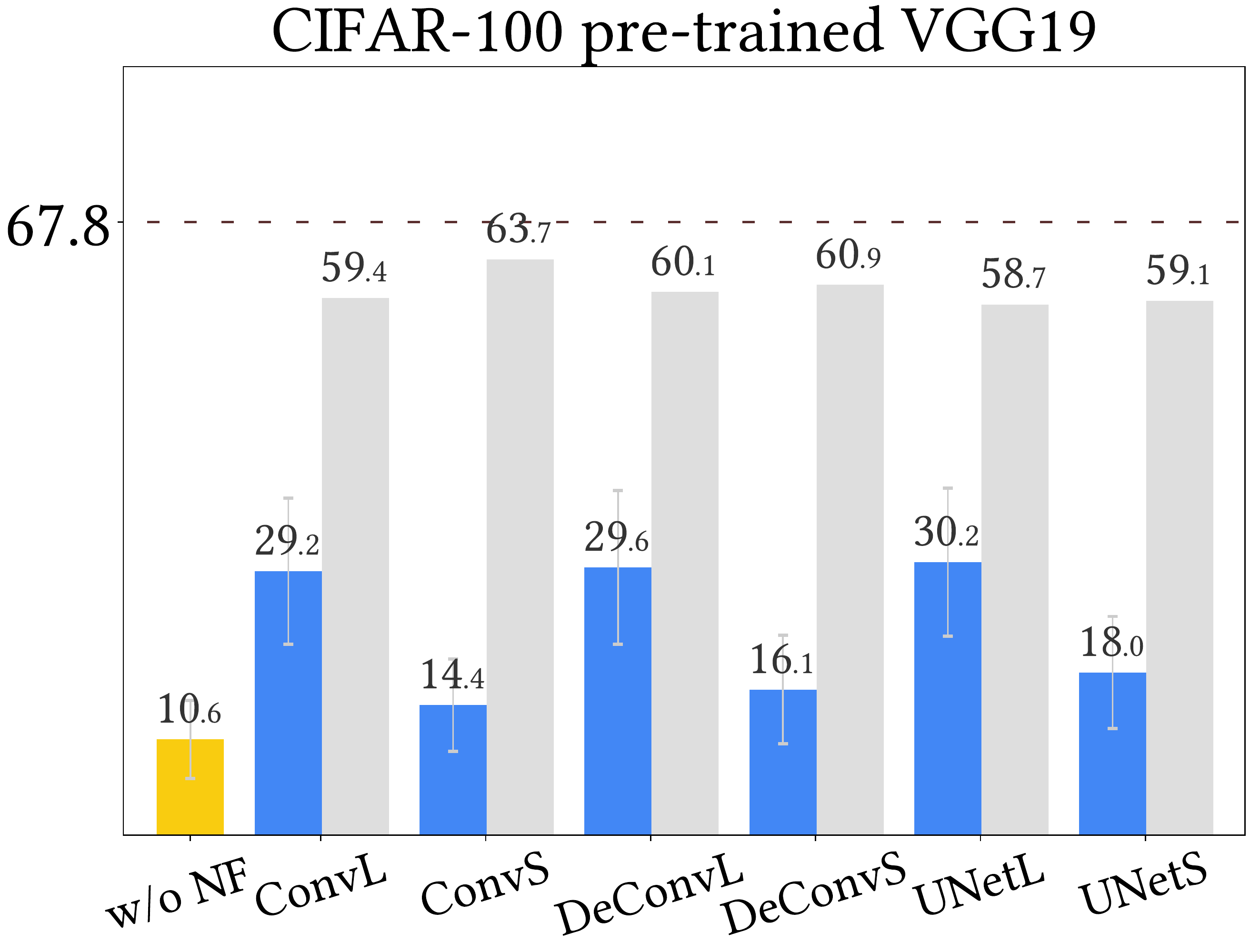}
     \end{subfigure}
     \end{adjustbox}\\
     {\small (a) CIFAR-100, $1\%$ Bit Error Rate\\}\vspace{2mm}

     \begin{adjustbox}{max width=.96\linewidth}
     \begin{subfigure}[b]{0.195\textwidth}
         \centering
         \includegraphics[width=\textwidth]{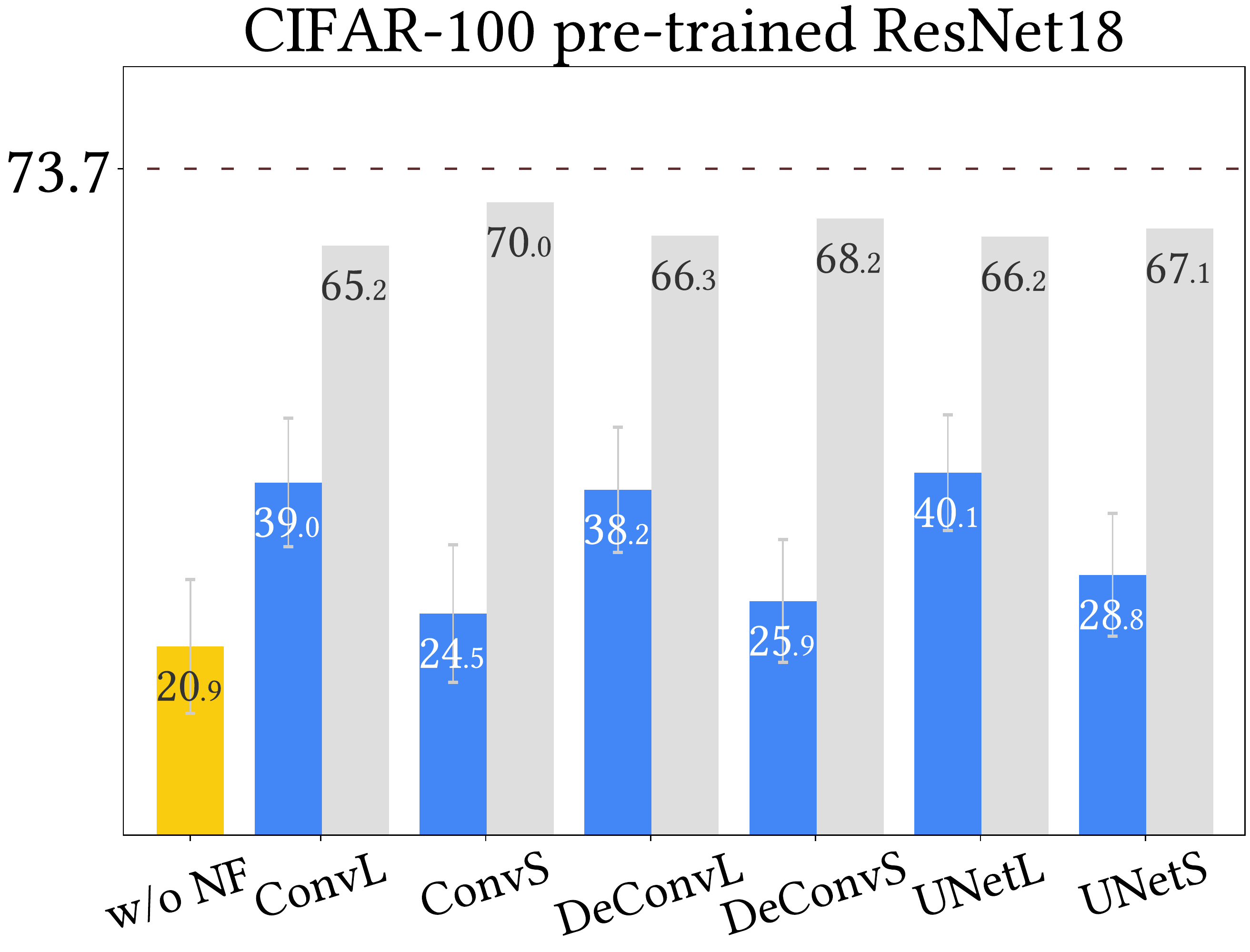}
     \end{subfigure}
     \hfill
     \begin{subfigure}[b]{0.195\textwidth}
         \centering
         \includegraphics[width=\textwidth]{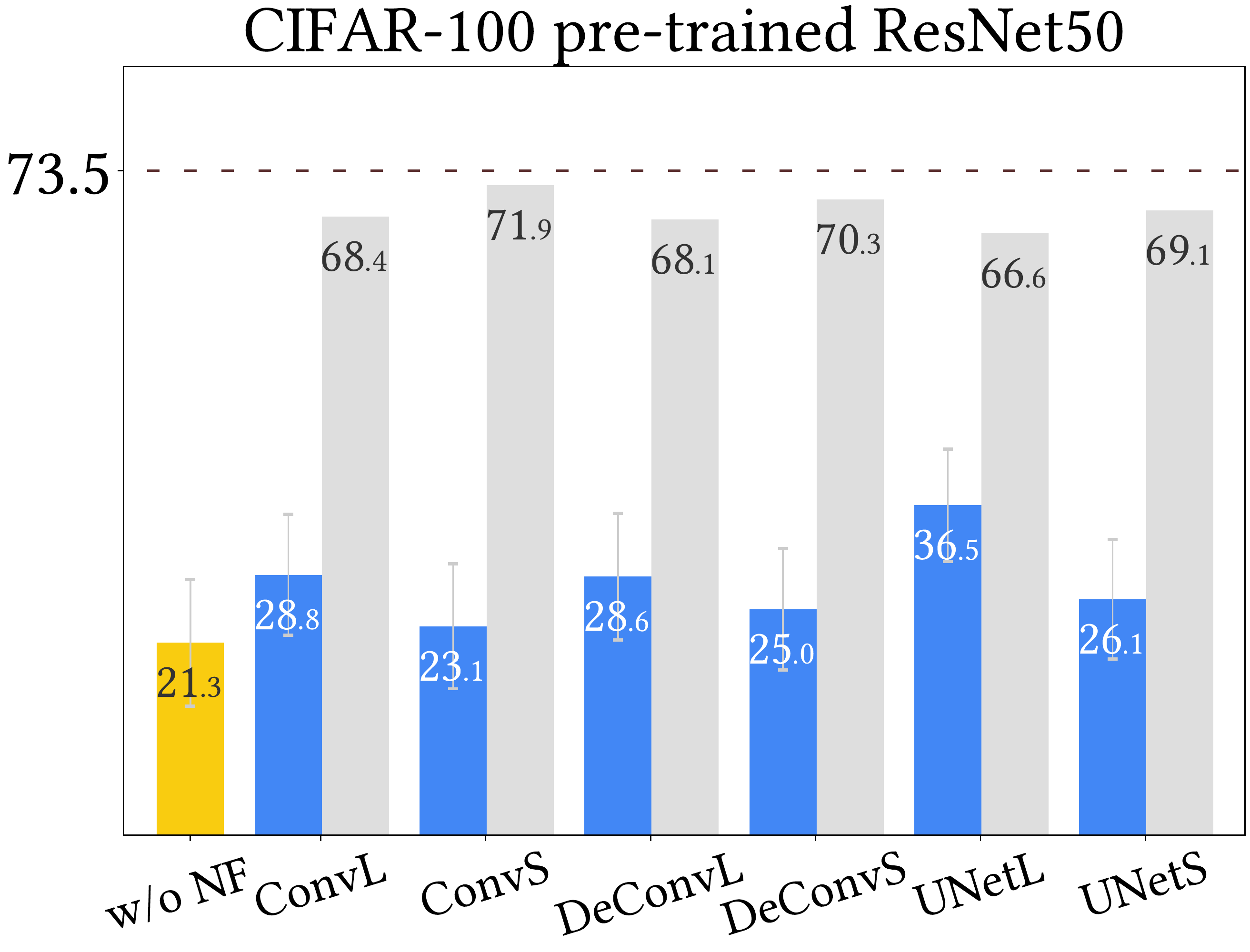}
     \end{subfigure}
     \hfill
     \begin{subfigure}[b]{0.195\textwidth}
         \centering
         \includegraphics[width=\textwidth]{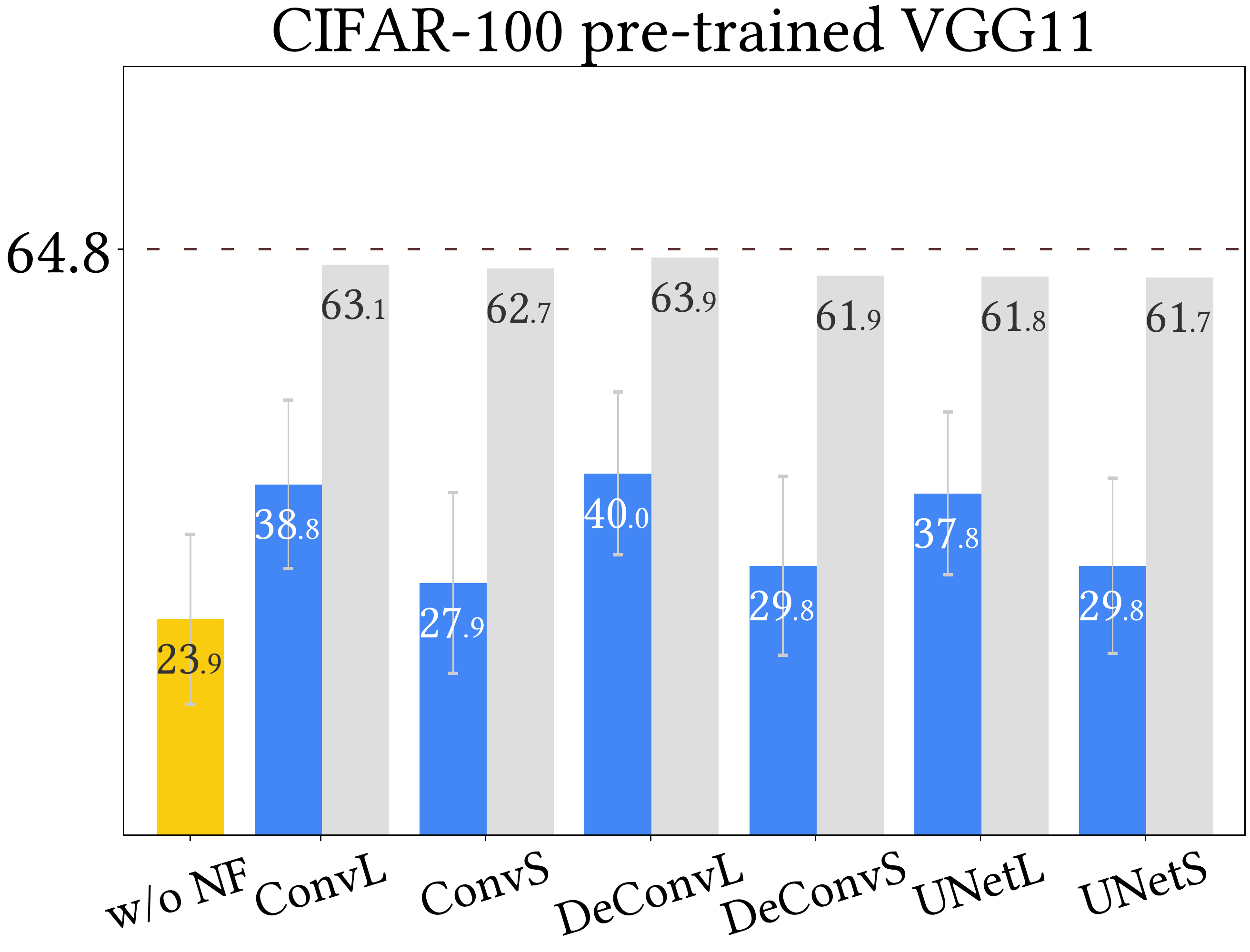}
     \end{subfigure}
     \hfill
     \begin{subfigure}[b]{0.195\textwidth}
         \centering
         \includegraphics[width=\textwidth]{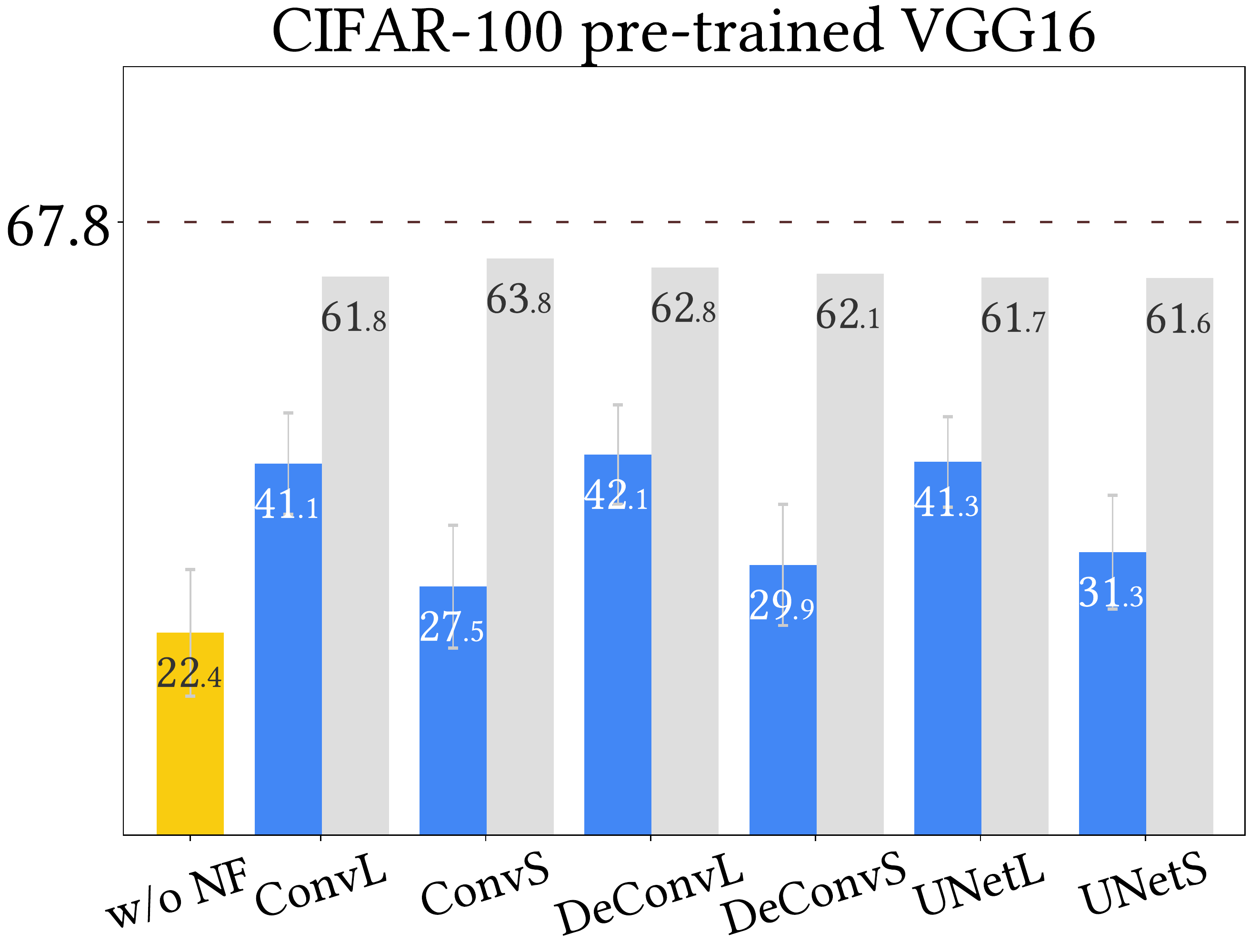}
     \end{subfigure}
     \hfill
     \begin{subfigure}[b]{0.195\textwidth}
         \centering
         \includegraphics[width=\textwidth]{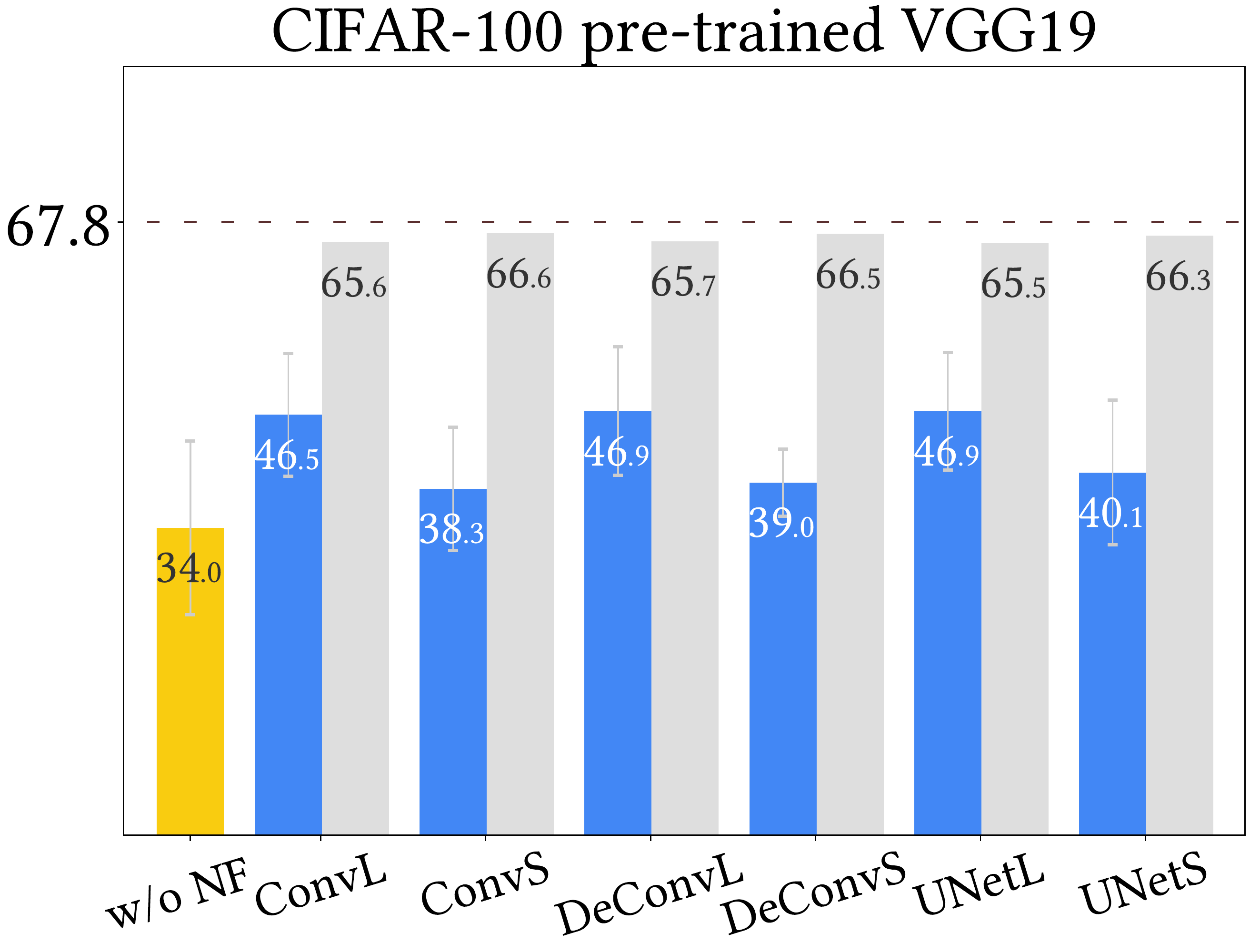}
     \end{subfigure}
     \end{adjustbox}\\
     {\small (b) CIFAR-100, $0.5\%$ Bit Error Rate\\}\vspace{2mm}

     \begin{adjustbox}{max width=.96\linewidth}
     \begin{subfigure}[b]{0.195\textwidth}
         \centering
         \includegraphics[width=\textwidth]{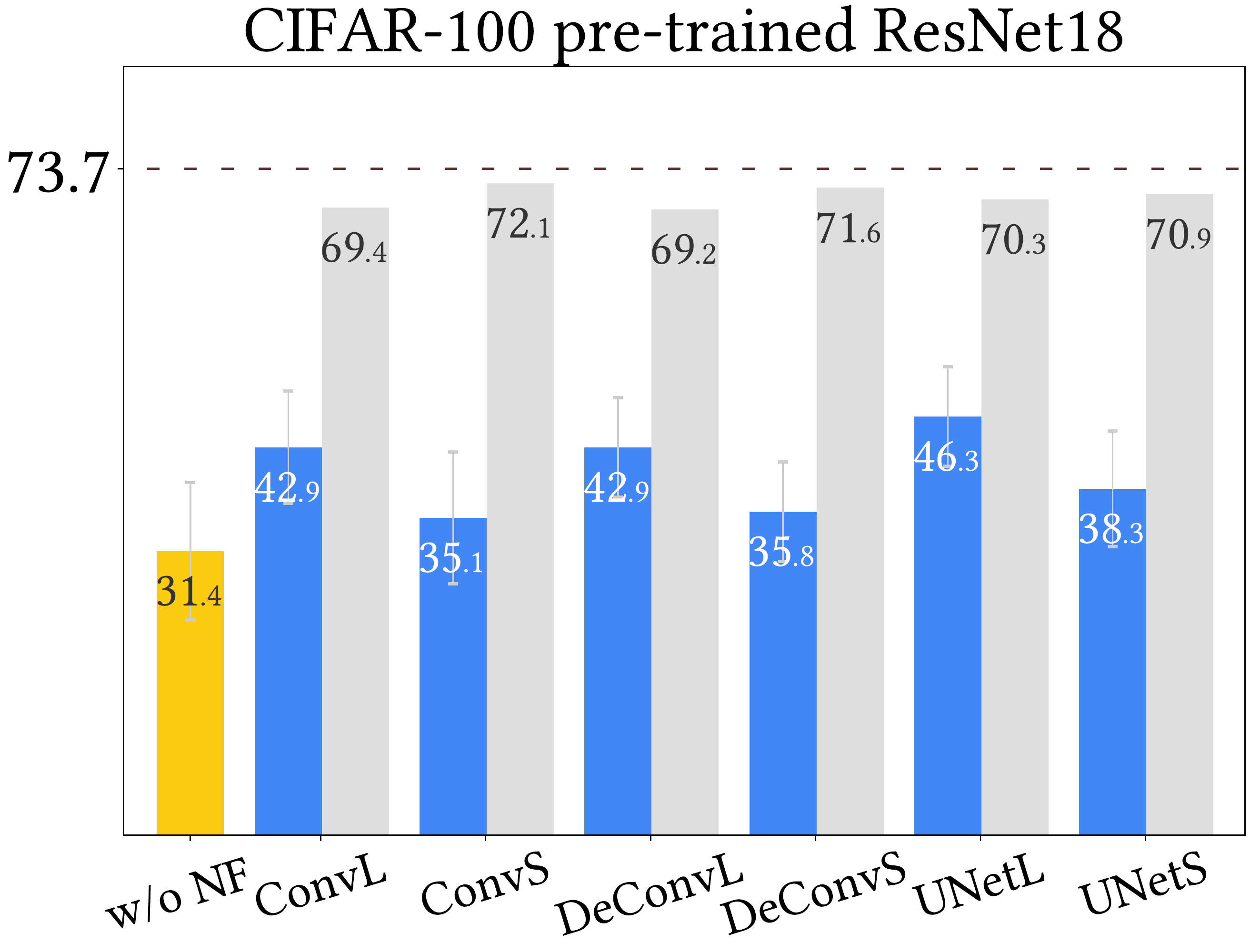}
     \end{subfigure}
     \hfill
     \begin{subfigure}[b]{0.195\textwidth}
         \centering
         \includegraphics[width=\textwidth]{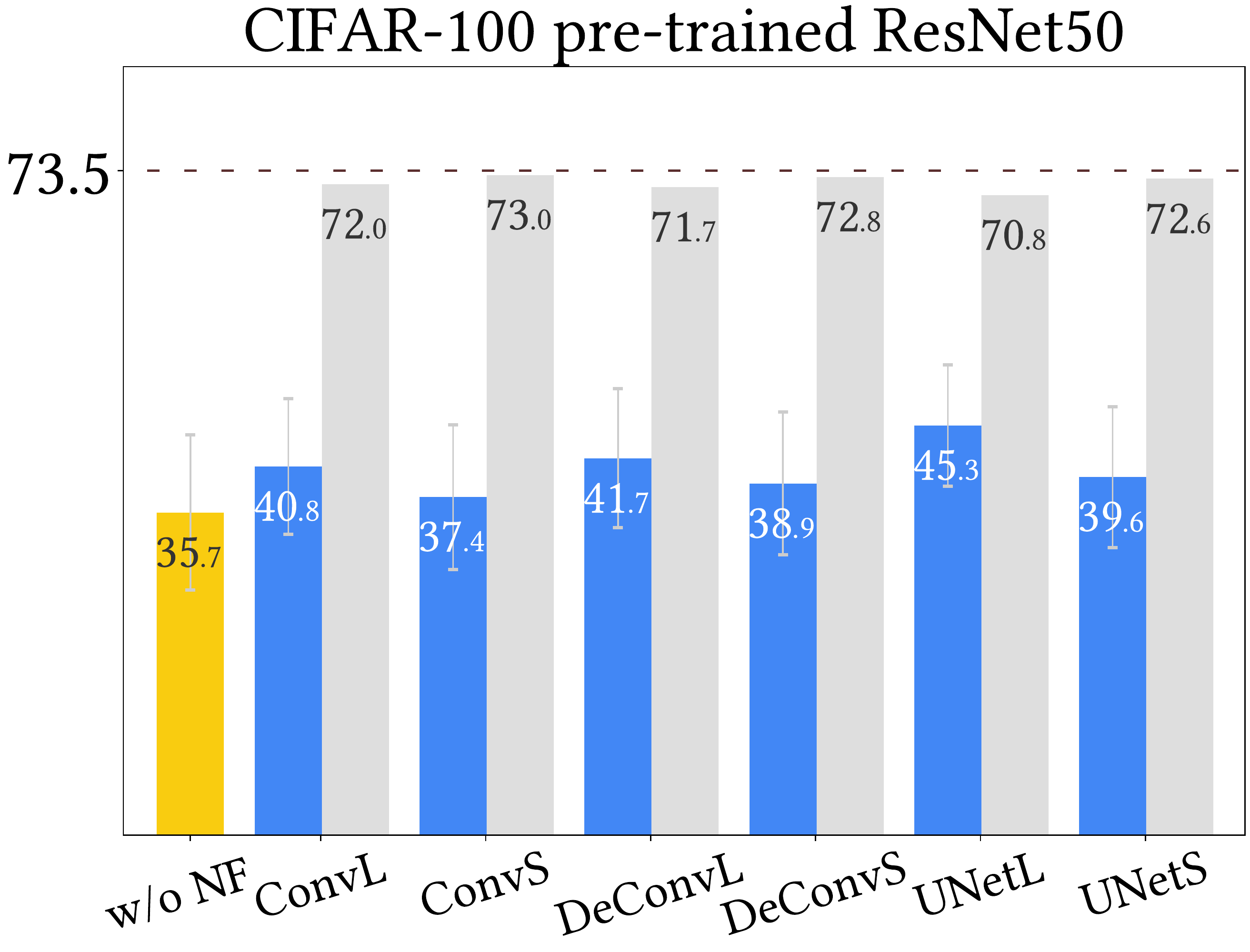}
     \end{subfigure}
     \hfill
     \begin{subfigure}[b]{0.195\textwidth}
         \centering
         \includegraphics[width=\textwidth]{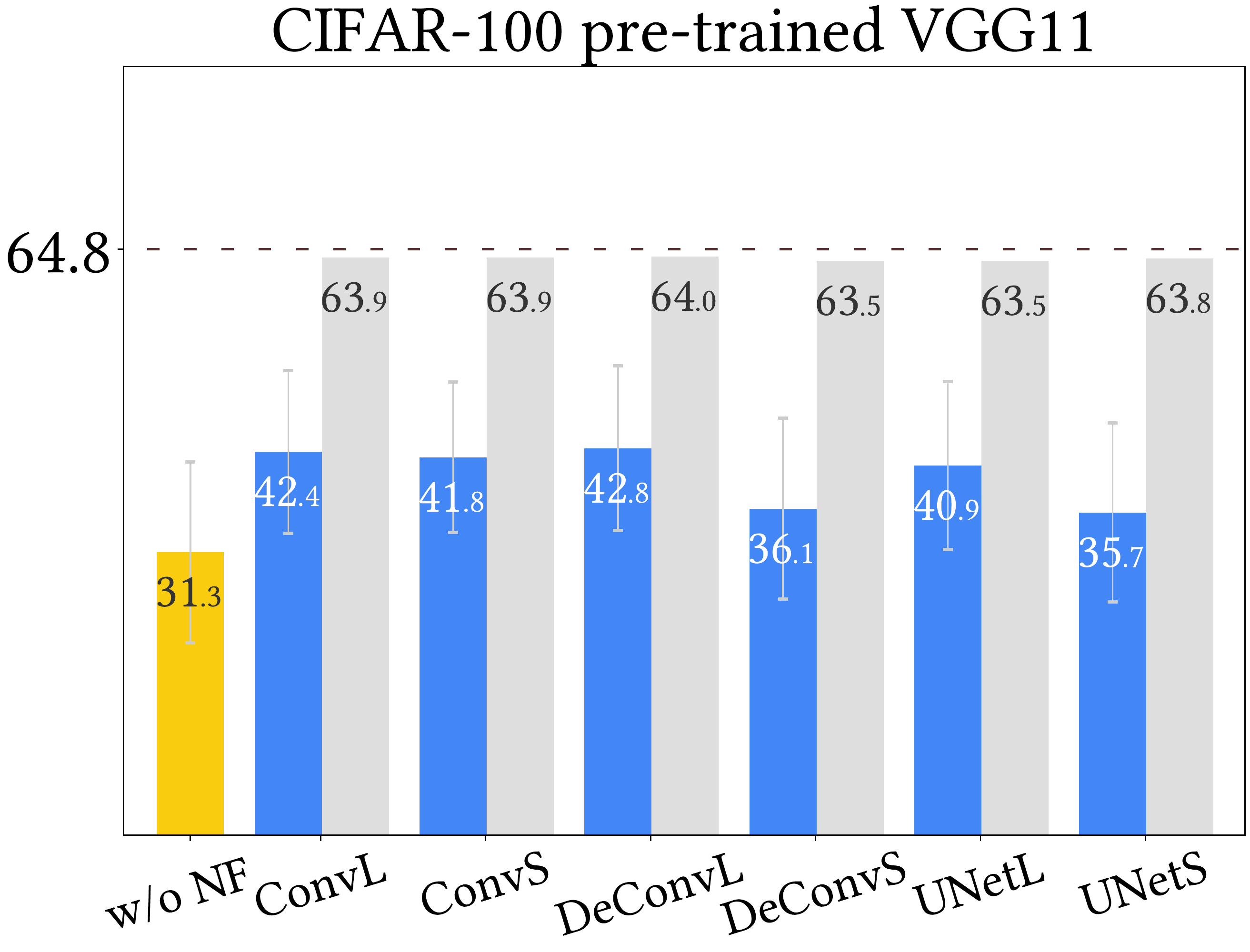}
     \end{subfigure}
     \hfill
     \begin{subfigure}[b]{0.195\textwidth}
         \centering
         \includegraphics[width=\textwidth]{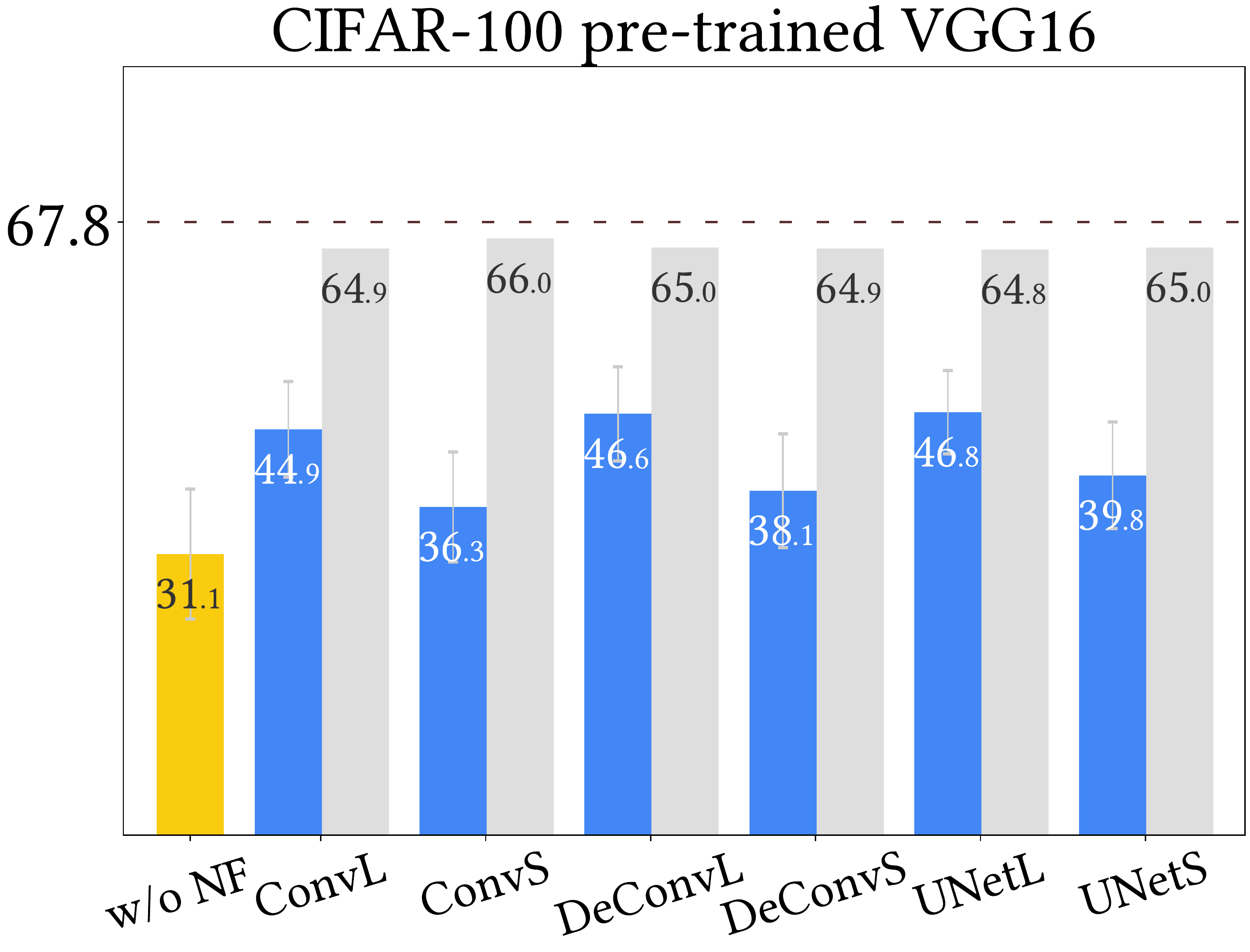}
     \end{subfigure}
     \hfill
     \begin{subfigure}[b]{0.195\textwidth}
         \centering
         \includegraphics[width=\textwidth]{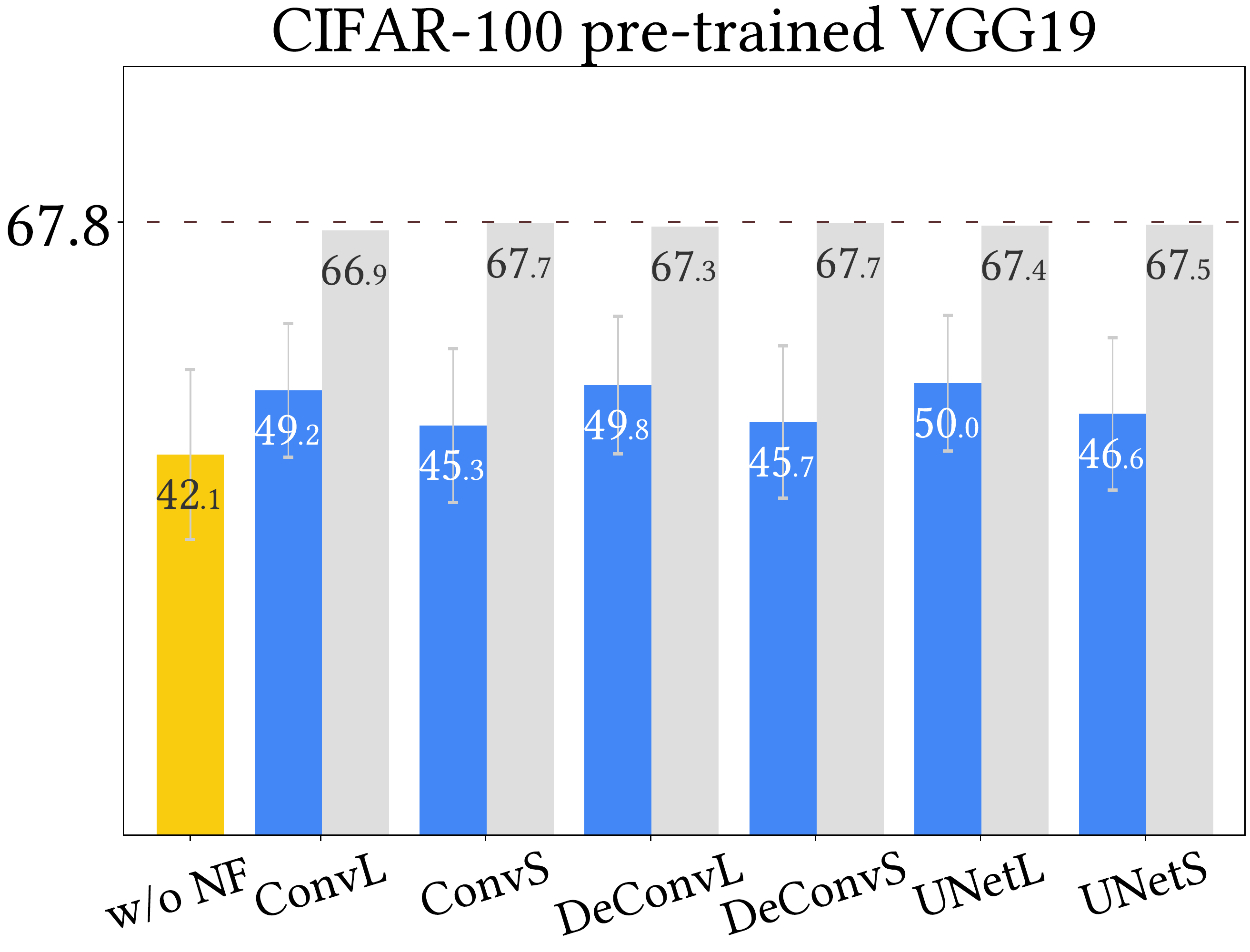}
     \end{subfigure}
     \end{adjustbox}\\
     {\small (c) CIFAR-100, $0.35\%$ Bit Error Rate\\}
        \caption{Experimental results on CIFAR-100.}
        \label{fig:cifar100_experimental_results_appendix}
\end{figure}

%% file: assets/_tables/_tables_Appendix_Appendixtransferabilityp1CIFAR10.tex
\begin{table*}[ht]
\renewcommand{\arraystretch}{1.2}
\centering
\caption{Transfer results on CIFAR-10: NeuralFuse trained on SM with 1\% BER} 
\label{table:Appendixtransferabilityp1CIFAR10}
\begin{adjustbox}{max width=\linewidth}
\begin{threeparttable}
\begin{tabular}{ c|c|c|cc|crr|crr }

\hline
\multirow{2}{*}{SM} & \multirow{2}{*}{TM} & \multirow{2}{*}{BER} & \multirow{2}{*}{CA} & \multirow{2}{*}{PA} & \multicolumn{3}{c|}{ConvL (1\%)} & \multicolumn{3}{c}{UNetL (1\%)} \\
& & & & & CA (NF) & PA (NF) & RP & CA (NF) & PA (NF) & RP \\
\hline

\multirow{9}{*}{ResNet18} & ResNet18 & 0.5\% & 92.6 & 70.1 $\pm$ 11.6 & 89.8 & 89.5 $\pm$ 0.2 & 19.4 & 86.6 & 86.2 $\pm$ 0.3 & 16.1 \\
\cline{2-11}

& \multirow{2}{*}{ResNet50} & 1\% & \multirow{2}{*}{92.6} & 26.1 $\pm$ ~~9.4 & \multirow{2}{*}{89.5} & 36.0 $\pm$ ~19 & 9.9 & \multirow{2}{*}{85.2} & 38.8 $\pm$ ~19 & 12.7 \\
&  & 0.5\% & & 61.0 $\pm$ 10.3 & & 75.1 $\pm$ ~10 & 14.1 & & 77.1 $\pm$ 5.0 & 16.1\\ 
\cline{2-11}

& \multirow{2}{*}{VGG11} & 1\% & \multirow{2}{*}{88.4} & 42.2 $\pm$ 11.6 & \multirow{2}{*}{88.4} & 62.5 $\pm$ 8.4 & 20.3 & \multirow{2}{*}{76.8} & 61.1 $\pm$ 8.5 & 18.9 \\
&  & 0.5\% & & 63.6 $\pm$ ~~9.3 & & 81.0 $\pm$ 4.6 & 17.4 & & 73.7 $\pm$ 3.0 & 10.1\\ 
\cline{2-11}

& \multirow{2}{*}{VGG16} & 1\% & \multirow{2}{*}{90.3} & 35.7 $\pm$ ~~7.9 & \multirow{2}{*}{89.6} & 63.3 $\pm$ ~18 & 27.6  & \multirow{2}{*}{85.2} & 59.9 $\pm$ ~16 & 24.2 \\
&  & 0.5\% & & 66.6 $\pm$ ~~8.1 & & 85.0 $\pm$ 3.4 & 18.4 & & 80.2 $\pm$ 4.5 & 13.6\\ 
\cline{2-11}

& \multirow{2}{*}{VGG19} & 1\% & \multirow{2}{*}{90.5} & 36.0 $\pm$ 12.0 & \multirow{2}{*}{89.6} & 50.7 $\pm$ ~22 & 14.7 & \multirow{2}{*}{85.3} & 51.1 $\pm$ ~16 & 15.1 \\
&  & 0.5\% & & 64.2 $\pm$ 12.4 & & 80.2 $\pm$ 8.7 & 16.0 & & 76.5 $\pm$ 7.8 & 12.3\\ 
\hline

\multirow{9}{*}{VGG19} & \multirow{2}{*}{ResNet18} & 1\% & \multirow{2}{*}{92.6} & 38.9 $\pm$ 12.4 & \multirow{2}{*}{89.8} & 61.0 $\pm$ ~17 & 22.1 & \multirow{2}{*}{87.0} & 69.7 $\pm$ ~11 & 30.8 \\
&  & 0.5\% & & 70.1 $\pm$ 11.6 & & 86.1 $\pm$ 6.9 & 16.0 & & 84.2 $\pm$ 3.0 & 14.1 \\
\cline{2-11}

& \multirow{2}{*}{ResNet50} & 1\% & \multirow{2}{*}{92.6} & 26.1 $\pm$ ~~9.4 & \multirow{2}{*}{89.9} & 34.0 $\pm$ ~19 & 7.9 & \multirow{2}{*}{87.0} & 44.2 $\pm$ ~17 & 18.1 \\
&  & 0.5\% & & 61.0 $\pm$ 10.3 & & 76.5 $\pm$ ~10 & 15.5 & & 80.7 $\pm$ 4.2 & 19.7 \\
\cline{2-11}

& \multirow{2}{*}{VGG11} & 1\% & \multirow{2}{*}{88.4} & 42.2 $\pm$ 11.6 & \multirow{2}{*}{89.7} & 76.5 $\pm$ 7.0 & 34.3 & \multirow{2}{*}{87.1} & 79.9 $\pm$ 5.6 & 37.7 \\
&  & 0.5\% & & 63.6 $\pm$ ~~9.3 & & 88.0 $\pm$ 2.1 & 24.4 & & 85.4 $\pm$ 0.8 & 21.8 \\
\cline{2-11}

& \multirow{2}{*}{VGG16} & 1\% & \multirow{2}{*}{90.3} & 35.7 $\pm$ ~~7.9 & \multirow{2}{*}{89.6} & 75.5 $\pm$ ~12 & 39.8 & \multirow{2}{*}{87.2} & 78.9 $\pm$ 7.8 & 43.2 \\
&  & 0.5\% & & 66.6 $\pm$ ~~8.1 & & 88.9 $\pm$ 0.6 & 22.3 & & 86.2 $\pm$ 0.3 & 19.6 \\
\cline{2-11}

& VGG19 & 0.5\% & 90.5 & 64.2 $\pm$ 12.4 & 89.8 & 89.6 $\pm$ 8.7 & 25.4 & 87.4 & 86.8 $\pm$ 0.4 & 22.6 \\
\hline

\end{tabular}
    \begin{tablenotes}
      \small
      \item {\textit{Note}. SM: source model, used for training generators; TM: target model, used for testing generators; BER: the bit-error rate of the target model; CA ($\%$): clean accuracy; PA ($\%$): perturbed accuracy; NF: NeuralFuse; and RP: total recovery percentage of PA (NF) vs. PA}
    \end{tablenotes}
\end{threeparttable}
\end{adjustbox}
\end{table*}

%% file: assets/_tables/_tables_Appendix_Appendixtransferabilityp15GTSRB.tex
\begin{table*}[ht]
\renewcommand{\arraystretch}{1.2}
\centering
\caption{Transfer results on GTSRB: NeuralFuse trained on SM with 1.5\% BER} 
\label{table:Appendixtransferabilityp15GTSRB}
\begin{adjustbox}{max width=\linewidth}
\begin{threeparttable}
\begin{tabular}{ c|c|c|cc|crr|crr }

\hline
\multirow{2}{*}{SM} & \multirow{2}{*}{TM} & \multirow{2}{*}{BER} & \multirow{2}{*}{CA} & \multirow{2}{*}{PA} & \multicolumn{3}{c|}{ConvL (1.5\%)} & \multicolumn{3}{c}{UNetL (1.5\%)} \\
& & & & & CA (NF) & PA (NF) & RP & CA (NF) & PA (NF) & RP \\
\hline

\multirow{10}{*}{ResNet18} & \multirow{2}{*}{ResNet18} & 1\% & \multirow{2}{*}{95.5} & 36.9 $\pm$ 16.0 & \multirow{2}{*}{95.7} & 93.9 $\pm$ 1.9 & 57.0 & \multirow{2}{*}{94.9} & 94.4 $\pm$ 0.4 & 57.5 \\
&  & 0.5\% & & 75.2 $\pm$ 12.7 & & 95.7 $\pm$ 0.2 & 20.5 & & 94.8 $\pm$ 0.2 & 19.6 \\
\cline{2-11}

& \multirow{2}{*}{ResNet50} & 1\% & \multirow{2}{*}{95.0} & 29.5 $\pm$ 16.9 & \multirow{2}{*}{94.4} & 37.0 $\pm$ ~22 & 7.5 &  \multirow{2}{*}{94.4} & 47.1 $\pm$ ~23 & 17.6 \\
&  & 0.5\% & & 74.0 $\pm$ ~13.0 & & 77.5 $\pm$ ~13 & 3.5 & & 84.8 $\pm$ 9.5 & 10.8\\ 
\cline{2-11}

& \multirow{2}{*}{VGG11} & 1\% & \multirow{2}{*}{91.9} & 34.9 $\pm$ 12.4 & \multirow{2}{*}{92.8} & 45.2 $\pm$ ~10 & 10.3 &  \multirow{2}{*}{91.4} & 50.5 $\pm$ ~13 & 15.6 \\
&  & 0.5\% & & 64.9 $\pm$ 10.8 & & 79.4 $\pm$ 5.8 & 14.5 & & 83.9 $\pm$ 4.2 & 19.0\\ 
\cline{2-11}

& \multirow{2}{*}{VGG16} & 1\% & \multirow{2}{*}{95.2} & 15.1 $\pm$ ~~6.8 & \multirow{2}{*}{95.4} & 31.1 $\pm$ ~13 & 15.9 &  \multirow{2}{*}{94.6} & 36.8 $\pm$ ~12 & 21.7 \\
&  & 0.5\% & & 58.8 $\pm$ ~~8.9 & & 84.5 $\pm$ 8.3 & 25.8 & & 86.0 $\pm$ 8.6 & 27.2\\ 
\cline{2-11}

& \multirow{2}{*}{VGG19} & 1\% & \multirow{2}{*}{95.5} & 36.6 $\pm$ ~~6.8 & \multirow{2}{*}{95.0} & 56.4 $\pm$ ~15 & 19.8 &  \multirow{2}{*}{94.3} & 60.8 $\pm$ ~15 & 24.2 \\
&  & 0.5\% & & 69.1 $\pm$ 11.1 & & 86.9 $\pm$ 3.4 & 17.8 & & 87.7 $\pm$ 3.8 & 18.6\\ 
\hline

\multirow{10}{*}{VGG19} & \multirow{2}{*}{ResNet18} & 1\% & \multirow{2}{*}{95.5} & 36.9 $\pm$ 16.0 & \multirow{2}{*}{88.4} & 50.3 $\pm$ ~12 & 13.4 & \multirow{2}{*}{92.8} & 63.7 $\pm$ ~16 & 26.8 \\
&  & 0.5\% & & 75.2 $\pm$ 12.7 & & 77.9 $\pm$ 7.4 & 2.7 & & 87.5 $\pm$ 3.9 & 12.3 \\
\cline{2-11}

& \multirow{2}{*}{ResNet50} & 1\% & \multirow{2}{*}{95.0} & 29.5 $\pm$ 16.9 & \multirow{2}{*}{87.5} & 29.7 $\pm$ ~17 & 0.2  & \multirow{2}{*}{92.5} & 40.4 $\pm$ ~21 & 10.9 \\
&  & 0.5\% & & 74.0 $\pm$ ~13.0 & & 67.9 $\pm$ ~17 & -6.1 & & 77.5 $\pm$ ~15 & 3.5 \\
\cline{2-11}

& \multirow{2}{*}{VGG11} & 1\% & \multirow{2}{*}{91.9} & 34.9 $\pm$ 12.4 & \multirow{2}{*}{89.7} & 47.1 $\pm$ ~11 & 12.2 &  \multirow{2}{*}{93.5} & 60.0 $\pm$ ~12 & 25.1 \\
&  & 0.5\% & & 64.9 $\pm$ 10.8 & & 76.3 $\pm$ 5.1 & 11.4 & & 86.0 $\pm$ 3.8 & 21.1 \\
\cline{2-11}

& \multirow{2}{*}{VGG16} & 1\% & \multirow{2}{*}{95.2} & 15.1 $\pm$ ~~6.8 & \multirow{2}{*}{93.0} & 29.2 $\pm$ ~15 & 14.1 &  \multirow{2}{*}{93.0} & 38.5 $\pm$ ~16 & 23.4 \\
&  & 0.5\% & & 58.8 $\pm$ ~~8.9 & & 75.7 $\pm$ ~12 & 16.9 & & 79.9 $\pm$ 8.3 & 21.1 \\
\cline{2-11}

& \multirow{2}{*}{VGG19} & 1\% & \multirow{2}{*}{95.5} & 36.6 $\pm$ ~~6.8 & \multirow{2}{*}{95.1} & 87.4 $\pm$ 6.0 & 50.8 & \multirow{2}{*}{94.6} & 88.7 $\pm$ 5.0 & 52.1 \\
&  & 0.5\% & & 69.1 $\pm$ 11.1 & & 92.4 $\pm$ 2.4 & 23.3 & & 92.4 $\pm$ 2.2 & 23.3 \\
\hline

\end{tabular}
    \begin{tablenotes}
      \small
      \item {\textit{Note}. SM: source model, used for training generators; TM: target model, used for testing generators; BER: the bit-error rate of the target model; CA ($\%$): clean accuracy; PA ($\%$): perturbed accuracy; NF: NeuralFuse; and RP: total recovery percentage of PA (NF) vs. PA}
    \end{tablenotes}
\end{threeparttable}
\end{adjustbox}
\end{table*}

%% file: assets/_tables/_tables_Appendix_Appendixtransferabilityp1GTSRB.tex
\begin{table*}[ht]
\vspace{-5mm}
\renewcommand{\arraystretch}{1.2}
\centering
\caption{Transfer results on GTSRB: NeuralFuse trained on SM with 1\% BER}
\label{table:Appendixtransferabilityp1GTSRB}
\begin{adjustbox}{max width=\linewidth}
\begin{threeparttable}
\begin{tabular}{c|c|c|cc|crr|crr }

\hline
\multirow{2}{*}{SM} & \multirow{2}{*}{TM} & \multirow{2}{*}{BER} & \multirow{2}{*}{CA} & \multirow{2}{*}{PA} & \multicolumn{3}{c|}{ConvL (1\%)} & \multicolumn{3}{c}{UNetL (1\%)} \\
& & & & & CA (NF) & PA (NF) & RP & CA (NF) & PA (NF) & RP \\
\hline

\multirow{9}{*}{ResNet18} & ResNet18 & 0.5\% & 95.5 & 75.2 $\pm$ 12.7 & 95.7 & 95.3 $\pm$ 0.5 & 20.1 & 96.2 & 95.7 $\pm$ 0.3 & 20.5 \\
\cline{2-11}

& \multirow{2}{*}{ResNet50} & 1\% & \multirow{2}{*}{95.0} & 29.5 $\pm$ 16.9 & \multirow{2}{*}{94.5} & 35.6 $\pm$ ~21 & 6.1 & \multirow{2}{*}{95.6} & 42.6 $\pm$ ~23 & 13.1 \\
&  & 0.5\% & & 74.0 $\pm$ 13.0 & & 78.8 $\pm$ ~13 & 4.8 & & 87.3 $\pm$ 9.0 & 13.3\\ 
\cline{2-11}

& \multirow{2}{*}{VGG11} & 1\% & \multirow{2}{*}{91.9} & 34.9 $\pm$ 12.4 & \multirow{2}{*}{93.1} & 45.8 $\pm$ ~11 & 10.9 & \multirow{2}{*}{94.0} & 47.1 $\pm$ ~14 & 12.2 \\
&  & 0.5\% & & 64.9 $\pm$ 10.8 & & 81.8 $\pm$ 5.0 & 16.9 & & 84.2 $\pm$ 4.8 & 19.3\\ 
\cline{2-11}

& \multirow{2}{*}{VGG16} & 1\% & \multirow{2}{*}{95.2} & 15.1 $\pm$ ~~6.8 & \multirow{2}{*}{95.5} & 26.5 $\pm$ ~12 & 11.4 & \multirow{2}{*}{95.5} & 32.4 $\pm$ ~11 & 17.3 \\
&  & 0.5\% & & 58.8 $\pm$ ~~8.9 & & 82.2 $\pm$ 9.0 & 23.4 & & 85.4 $\pm$ 6.7 & 26.6\\ 
\cline{2-11}

& \multirow{2}{*}{VGG19} & 1\% & \multirow{2}{*}{95.5} & 36.6 $\pm$ ~~6.8 & \multirow{2}{*}{94.9} & 53.2 $\pm$ ~14 & 16.6 & \multirow{2}{*}{95.6} & 60.9 $\pm$ ~15 & 24.3 \\
&  & 0.5\% & & 69.1 $\pm$ 11.1 & & 85.4 $\pm$ 4.5 & 16.3 & & 87.5 $\pm$ 3.7 & 18.4\\ 
\hline

\multirow{9}{*}{VGG19} & \multirow{2}{*}{ResNet18} & 1\% & \multirow{2}{*}{95.5} & 36.9 $\pm$ 16.0 & \multirow{2}{*}{93.7} & 53.1 $\pm$ ~16 & 16.2 & \multirow{2}{*}{95.0} & 63.4 $\pm$ ~18 & 26.5 \\
&  & 0.5\% & & 75.2 $\pm$ 12.7 & & 83.9 $\pm$ 7.6 & 8.7 & & 89.7 $\pm$ 4.8 & 14.5 \\
\cline{2-11}

& \multirow{2}{*}{ResNet50} & 1\% & \multirow{2}{*}{95.0} & 29.5 $\pm$ 16.9 & \multirow{2}{*}{92.8} & 30.6 $\pm$ ~18 & 1.1 &  \multirow{2}{*}{95.4} & 38.9 $\pm$ ~22 & 9.4 \\
&  & 0.5\% & & 74.0 $\pm$ 13.0 & & 74.7 $\pm$ ~18 & 0.7 & & 81.5 $\pm$ ~16 & 7.5 \\
\cline{2-11}

& \multirow{2}{*}{VGG11} & 1\% & \multirow{2}{*}{91.9} & 34.9 $\pm$ 12.4 & \multirow{2}{*}{93.7} & 50.6 $\pm$ ~11 & 15.7 & \multirow{2}{*}{95.1} & 58.9 $\pm$ ~15 & 24.0 \\
&  & 0.5\% & & 64.9 $\pm$ 10.8 & & 82.3 $\pm$ 5.1 & 17.4 & & 87.5 $\pm$ 3.7 & 22.6 \\
\cline{2-11}

& \multirow{2}{*}{VGG16} & 1\% & \multirow{2}{*}{95.2} & 15.1 $\pm$ ~~6.8 & \multirow{2}{*}{95.2} & 27.8 $\pm$ ~15 & 12.7 & \multirow{2}{*}{95.2} & 33.5 $\pm$ ~14 & 18.4 \\
&  & 0.5\% & & 58.8 $\pm$ ~~8.9 & & 79.0 $\pm$ ~12 & 20.2 & & 81.8 $\pm$ 7.8 & 23.0 \\
\cline{2-11}

& VGG19 & 0.5\% & 95.5 & 69.1 $\pm$ 11.1 & 96.0 & 94.0 $\pm$ 2.2 & 24.9 & 95.4 & 93.9 $\pm$ 2.1 & 24.8 \\
\hline

\end{tabular}
    \begin{tablenotes}
      \small
      \item {\textit{Note}. SM: source model, used for training generators; TM: target model, used for testing generators; BER: the bit-error rate of the target model; CA ($\%$): clean accuracy; PA ($\%$): perturbed accuracy; NF: NeuralFuse; and RP: total recovery percentage of PA (NF) vs. PA}
    \end{tablenotes}
\end{threeparttable}
\end{adjustbox}
\vspace{-3.5cm}
\end{table*}

%% file: assets/_tables/_tables_Appendix_Appendixtransferabilityp1CIFAR100.tex
\begin{table*}[ht]
\setlength{\tabcolsep}{3mm}
\renewcommand{\arraystretch}{1.15}
\centering
\caption{Transfer results on CIFAR-100: NeuralFuse trained on SM with 1\% BER} 
\label{table:Appendixtransferabilityp1CIFAR100}
\begin{adjustbox}{max width=\linewidth}
\begin{threeparttable}
\begin{tabular}{ c|c|c|cc|crr|crr }
\hline
\multirow{2}{*}{SM} & \multirow{2}{*}{TM} & \multirow{2}{*}{BER} & \multirow{2}{*}{CA} & \multirow{2}{*}{PA} & \multicolumn{3}{c|}{ConvL (1\%)} & \multicolumn{3}{c}{UNetL (1\%)} \\
& & & & & CA (NF) & PA (NF) & RP & CA (NF) & PA (NF) & RP \\
\hline

\multirow{14}{*}{ResNet18} & \multirow{2}{*}{ResNet18} & 0.5\% & \multirow{2}{*}{73.7} & 20.9 $\pm$ ~~7.4 & \multirow{2}{*}{54.8} & 35.8 $\pm$ 5.2 & 14.9 & \multirow{2}{*}{50.6} & 39.3 $\pm$ 2.8 & 18.4 \\
&  & 0.35\% & & 31.4 $\pm$ ~~7.6 & & 41.7 $\pm$ 3.7 & 10.3 & & 43.3 $\pm$ 1.4 & 11.9\\ 
\cline{2-11}

& \multirow{3}{*}{ResNet50} & 1\% & \multirow{3}{*}{73.5} & ~~3.0 $\pm$ ~~1.8 & \multirow{3}{*}{44.9} & 2.2 $\pm$ 2.0 & -0.8 & \multirow{3}{*}{41.5} & 2.4 $\pm$ 1.9 & -0.6 \\
&  & 0.5\% & & 21.3 $\pm$ ~~7.0 & & 15.9 $\pm$ 8.2 & -5.4 & & 17.1 $\pm$ 7.1 & -4.2\\ 
&  & 0.35\% & & 35.7 $\pm$ ~~8.6 & & 23.7 $\pm$ 7.1 & -12.0 & & 26.2 $\pm$ 5.6 & -9.5\\ 
\cline{2-11}

& \multirow{3}{*}{VGG11} & 1\% & \multirow{3}{*}{64.8} & ~~8.2 $\pm$ ~~5.7 & \multirow{3}{*}{41.2} & 9.8 $\pm$ 5.6 & 1.6 & \multirow{3}{*}{37.5} & 10.2 $\pm$ 5.1 & 2.0 \\
&  & 0.5\% & & 23.9 $\pm$ ~~9.4 & & 24.2 $\pm$ 5.9 & 0.3 & & 24.5 $\pm$ 4.7 & 0.6\\ 
&  & 0.35\% & & 31.3 $\pm$ 10.0 & & 29.0 $\pm$ 5.4 & -2.3 & & 28.2 $\pm$ 4.5 & -3.1\\
\cline{2-11}

& \multirow{3}{*}{VGG16} & 1\% & \multirow{3}{*}{67.8} & ~~7.0 $\pm$ ~~3.5 & \multirow{3}{*}{44.0} & 7.9 $\pm$ 3.7 & 0.9 & \multirow{3}{*}{39.5} & 10.1 $\pm$ 4.5 & 3.1 \\
&  & 0.5\% & & 22.4 $\pm$ ~~7.0 & & 22.4 $\pm$ 7.6 & 0.0 & & 26.3 $\pm$ 5.3 & 3.9\\ 
&  & 0.35\% & & 31.1 $\pm$ ~~7.2 & & 28.1 $\pm$ 5.9 & -3.0 & & 30.6 $\pm$ 3.6 & -0.5\\ 
\cline{2-11}

& \multirow{3}{*}{VGG19} & 1\% & \multirow{3}{*}{67.8} & 10.6 $\pm$ ~~4.3 & \multirow{3}{*}{44.2} & 13.5 $\pm$ 6.1 & 2.9 & \multirow{3}{*}{40.7} & 15.6 $\pm$ 6.2 & 5.0 \\
&  & 0.5\% & & 34.0 $\pm$ ~~9.6 & & 27.9 $\pm$ 4.8 & -6.1 & & 29.3 $\pm$ 4.6 & -4.7\\ 
&  & 0.35\% & & 42.1 $\pm$ ~~9.4 & & 33.2 $\pm$ ~48 & -8.9 & & 32.8 $\pm$ 3.9 & -9.3\\ 
\hline

\multirow{14}{*}{VGG19} & \multirow{3}{*}{ResNet18} & 1\% & \multirow{3}{*}{73.7} & ~~4.6 $\pm$ ~~2.9 & \multirow{3}{*}{55.5} & 5.8 $\pm$ 3.7 & 1.2 & \multirow{3}{*}{57.3} & 6.8 $\pm$ 4.4 & 2.2 \\
&  & 0.5\% & & 20.9 $\pm$ ~~7.4 & & 24.6 $\pm$ 6.3 & 3.7 & & 28.1 $\pm$ 5.9 & 7.2 \\
&  & 0.35\% & & 31.4 $\pm$ ~~7.6 & & 31.1 $\pm$ 5.0 & -0.3 & & 36.4 $\pm$ 4.5 & 5.0 \\
\cline{2-11}

& \multirow{3}{*}{ResNet50} & 1\% & \multirow{3}{*}{73.5} & ~~3.0 $\pm$ ~~1.8 & \multirow{3}{*}{56.1} & 2.8 $\pm$ 2.1 & -0.2 & \multirow{3}{*}{56.1} & 3.7 $\pm$ 2.4 & 0.7 \\
&  & 0.5\% & & 21.3 $\pm$ ~~7.0 & & 18.9 $\pm$ 8.6 & -2.4 & & 22.8 $\pm$ 8.5 & 1.5 \\
&  & 0.35\% & & 35.7 $\pm$ ~~8.6 & & 28.7 $\pm$ 8.2 & -7.0 & & 33.7 $\pm$ 7.0 & -2.0 \\
\cline{2-11}

& \multirow{3}{*}{VGG11} & 1\% & \multirow{3}{*}{64.8} & ~~8.2 $\pm$ ~~5.7 & \multirow{3}{*}{52.8} & 12.3 $\pm$ 8.4 & 4.1 & \multirow{3}{*}{53.9} & 15.4 $\pm$ 9.4 & 7.2 \\
&  & 0.5\% & & 23.9 $\pm$ ~~9.4 & & 30.0 $\pm$ 9.3 & 6.1 & & 33.3 $\pm$ 7.2 & 9.4 \\
&  & 0.35\% & & 31.3 $\pm$ 10.0 & & 36.5 $\pm$ 7.7 & 5.2 & & 38.8 $\pm$ 6.5 & 7.5 \\
\cline{2-11}

& \multirow{3}{*}{VGG16} & 1\% & \multirow{3}{*}{67.8} & ~~7.0 $\pm$ ~~3.5 & \multirow{3}{*}{53.6} & 11.2 $\pm$ 4.4 & 4.2 & \multirow{3}{*}{55.2} & 13.6 $\pm$ 5.2 & 6.6 \\
&  & 0.5\% & & 22.4 $\pm$ ~~7.0 & & 32.4 $\pm$ 7.3 & 10.0 &  & 35.9 $\pm$ 6.2 & 13.5 \\
&  & 0.35\% & & 31.1 $\pm$ ~~7.2 & & 39.4 $\pm$ 6.3 & 8.3 & & 42.4 $\pm$ 4.9 & 11.3 \\
\cline{2-11}

& \multirow{2}{*}{VGG19} & 0.5\% & \multirow{2}{*}{67.8} & 34.0 $\pm$ ~~9.6 & \multirow{2}{*}{59.4} & 50.2 $\pm$ 3.1 & 16.2 & \multirow{2}{*}{58.7} & 49.1 $\pm$ 3.5 & 15.1 \\
&  & 0.35\% & & 42.1 $\pm$ ~~9.4 & & 53.1 $\pm$ 2.3 & 11.0 & & 52.0 $\pm$ 3.1 & 9.9 \\
\hline

\end{tabular}
    \begin{tablenotes}
      \small
      \item {\textit{Note}. SM: source model, used for training generators; TM: target model, used for testing generators; BER: the bit-error rate of the target model; CA ($\%$): clean accuracy; PA ($\%$): perturbed accuracy; NF: NeuralFuse; and RP: total recovery percentage of PA (NF) vs. PA}
    \end{tablenotes}
\end{threeparttable}
\end{adjustbox}
\end{table*}

%% file: assets/_tables/_tables_Appendix_Appendixtransferabilityp05CIFAR100.tex
\begin{table*}[ht]
\vspace{-2mm}
\setlength{\tabcolsep}{3mm}
\renewcommand{\arraystretch}{1.15}
\centering
\caption{Transfer results on CIFAR-100: NeuralFuse trained on SM with 0.5\% BER}
\label{table:Appendixtransferabilityp05CIFAR100}
\begin{adjustbox}{max width=\linewidth}
\begin{threeparttable}
\begin{tabular}{ c|c|c|cc|crr|crr }

\hline
\multirow{2}{*}{SM} & \multirow{2}{*}{TM} & \multirow{2}{*}{BER} & \multirow{2}{*}{CA} & \multirow{2}{*}{PA} & \multicolumn{3}{c|}{ConvL (0.5\%)} & \multicolumn{3}{c}{UNetL (0.5\%)} \\
& & & & & CA (NF) & PA (NF) & RP & CA (NF) & PA (NF) & RP \\
\hline

\multirow{9}{*}{ResNet18} & ResNet18 & 0.35\% & 73.7 & 31.4 $\pm$ ~~7.6 & 65.2 & 47.7 $\pm$ 4.9 & 16.3 & 66.2 & 49.2 $\pm$ 4.1 & 17.8 \\
\cline{2-11}

& \multirow{2}{*}{ResNet50} & 0.5\% & \multirow{2}{*}{73.5} & 21.3 $\pm$ ~~7.0 & \multirow{2}{*}{62.5} & 24.0 $\pm$ 9.9 & 2.8 & \multirow{2}{*}{63.5} & 26.4 $\pm$ 9.1 & 5.1 \\
&  & 0.35\% & & 35.7 $\pm$ ~~8.6 & & 36.3 $\pm$ 8.9 & 0.6 & & 39.4 $\pm$ 8.1 & 3.7\\ 
\cline{2-11}

& \multirow{2}{*}{VGG11} & 0.5\% & \multirow{2}{*}{64.8} & 23.9 $\pm$ ~~9.4 & \multirow{2}{*}{59.2} & 33.0 $\pm$ 9.8 & 9.2 & \multirow{2}{*}{61.1} & 34.2 $\pm$ 9.8 & 10.3 \\
&  & 0.35\% & & 31.3 $\pm$ 10.0 & & 40.4 $\pm$ 8.7 & 9.1 & & 41.4 $\pm$ 9.0 & 10.1\\ 
\cline{2-11}

& \multirow{2}{*}{VGG16} & 0.5\% & \multirow{2}{*}{67.8} & 22.4 $\pm$ ~~7.0 & \multirow{2}{*}{59.5} & 34.7 $\pm$ 8.0 & 12.3 & \multirow{2}{*}{61.4} & 37.5 $\pm$ 6.8 & 15.2 \\
&  & 0.35\% & & 31.1 $\pm$ ~~7.2 & & 42.9 $\pm$ 6.0 & 11.8 & & 45.3 $\pm$ 4.9 & 14.2\\ 
\cline{2-11}

& \multirow{2}{*}{VGG19} & 0.5\% & \multirow{2}{*}{67.8} & 34.0 $\pm$ ~~9.6 & \multirow{2}{*}{61.6} & 43.7 $\pm$ 6.2 & 9.6 & \multirow{2}{*}{62.0} & 45.0 $\pm$ 6.3 & 11.0 \\
&  & 0.35\% & & 42.1 $\pm$ ~~9.4 & & 49.0 $\pm$ 5.5 & 6.8 & & 50.5 $\pm$ 5.3 & 8.3\\ 
\hline

\multirow{9}{*}{VGG19} & \multirow{2}{*}{ResNet18} & 0.5\% & \multirow{2}{*}{73.7} & 20.9 $\pm$ ~~7.4 & \multirow{2}{*}{66.1} & 24.9 $\pm$ 6.7 & 4.0 & \multirow{2}{*}{67.8} & 27.7 $\pm$ 6.8 & 6.8 \\
&  & 0.35\% & & 31.4 $\pm$ ~~7.6 & & 34.4 $\pm$ 5.4 & 3.0 & & 38.1 $\pm$ 5.6 & 6.7 \\
\cline{2-11}

& \multirow{2}{*}{ResNet50} & 0.5\% & \multirow{2}{*}{73.5} & 21.3 $\pm$ ~~7.0 & \multirow{2}{*}{66.2} & 22.7 $\pm$ 7.8 & 1.4 & \multirow{2}{*}{66.7} & 25.4 $\pm$ 8.0 & 4.2 \\
&  & 0.35\% & & 35.7 $\pm$ ~~8.6 & & 35.5 $\pm$ 7.7 & -0.2 & & 38.8 $\pm$ 7.5 & 3.2 \\
\cline{2-11}

& \multirow{2}{*}{VGG11} & 0.5\% & \multirow{2}{*}{64.8} & 23.9 $\pm$ ~~9.4 & \multirow{2}{*}{59.9} & 29.3 $\pm$ ~10 & 5.4 & \multirow{2}{*}{61.0} & 31.2 $\pm$ 9.8 & 7.4 \\
&  & 0.35\% & & 31.3 $\pm$ 10.0 & & 36.6 $\pm$ 9.5 & 5.3 & & 38.1 $\pm$ 9.0 & 6.8 \\
\cline{2-11}

& \multirow{2}{*}{VGG16} & 0.5\% & \multirow{2}{*}{67.8} & 22.4 $\pm$ ~~7.0 & \multirow{2}{*}{62.5} & 30.8 $\pm$ 7.3 & 8.4 & \multirow{2}{*}{62.6} & 33.0 $\pm$ 7.3 & 10.7 \\
&  & 0.35\% & & 31.1 $\pm$ ~~7.2 & & 40.0 $\pm$ 6.5 & 8.9 &  & 42.5 $\pm$ 5.9 & 11.3 \\
\cline{2-11}

& VGG19 & 0.35\% & 67.8 & 42.1 $\pm$ ~~9.4 & 65.6 & 52.0 $\pm$ 6.2 & 9.8 & 65.5 & 52.6 $\pm$ 6.1 & 10.4 \\
\hline

\end{tabular}
    \begin{tablenotes}
      \small
      \item {\textit{Note}. SM: source model, used for training generators; TM: target model, used for testing generators; BER: the bit-error rate of the target model; CA ($\%$): clean accuracy; PA ($\%$): perturbed accuracy; NF: NeuralFuse; and RP: total recovery percentage of PA (NF) vs. PA}
    \end{tablenotes}
\end{threeparttable}
\end{adjustbox}
\vspace{-4cm}
\end{table*}

%% file: assets/_tables/_tables_Appendix_AppendixtransferabilityDoublep15CIFAR10.tex
\begin{table*}[ht]
\renewcommand{\arraystretch}{1.15}
\centering
\caption{Transfer results on CIFAR-10: NeuralFuse trained on two SM with 1.5\% BER}
\label{table:AppendixtransferabilityDoublep15CIFAR10}
\begin{adjustbox}{max width=\linewidth}
\begin{threeparttable}
\begin{tabular}{ c|c|c|cc|crr|crr }

\hline
\multirow{2}{*}{SM} & \multirow{2}{*}{TM} & \multirow{2}{*}{BER} & \multirow{2}{*}{CA} & \multirow{2}{*}{PA} & \multicolumn{3}{c|}{ConvL (1.5\%)} & \multicolumn{3}{c}{UNetL (1.5\%)} \\
& & & & & CA (NF) & PA (NF) & RP & CA (NF) & PA (NF) & RP \\
\hline

\multirow{9}{*}{\begin{tabular}{@{}c@{}}ResNet18\\+\\VGG19\end{tabular}} & \multirow{2}{*}{ResNet18} & 1\% & \multirow{2}{*}{92.6} & 38.9 $\pm$ 12.4 & \multirow{2}{*}{89.4} & 88.1 $\pm$ 1.0 & 49.2 & \multirow{2}{*}{86.3} & 85.4 $\pm$ 0.5 & 46.5 \\
&  & 0.5\% & & 70.1 $\pm$ 11.6 & & 89.2 $\pm$ 0.2 & 19.1 & & 86.1 $\pm$ 0.2 & 16.0\\ 
\cline{2-11}

& \multirow{2}{*}{ResNet50} & 1\% & \multirow{2}{*}{92.6} & 26.1 $\pm$ ~~9.4 & \multirow{2}{*}{89.3} & 44.0 $\pm$ ~22 & 17.9 & \multirow{2}{*}{86.1} & 50.9 $\pm$ ~20 & 24.8 \\
&  & 0.5\% & & 61.0 $\pm$ 10.3 & & 80.3 $\pm$ 6.7 & 19.3 & & 78.6 $\pm$ 3.9 & 17.6\\ 
\cline{2-11}

& \multirow{2}{*}{VGG11} & 1\% & \multirow{2}{*}{88.4} & 42.2 $\pm$ 11.6 & \multirow{2}{*}{89.1} & 77.0 $\pm$ 5.6 & 34.8 & \multirow{2}{*}{85.9} & 82.3 $\pm$ 4.1 & 40.1 \\
&  & 0.5\% & & 63.6 $\pm$ ~~9.3 & & 87.5 $\pm$ 1.6 & 23.9 & & 85.0 $\pm$ 0.6 & 21.4\\ 
\cline{2-11}

& \multirow{2}{*}{VGG16} & 1\% & \multirow{2}{*}{90.3} & 35.7 $\pm$ ~~7.9 & \multirow{2}{*}{89.1} & 80.5 $\pm$ 8.6 & 44.8  & \multirow{2}{*}{85.7} & 81.4 $\pm$ 5.5 & 45.7 \\
&  & 0.5\% & & 66.6 $\pm$ ~~8.1 & & 88.2 $\pm$ 0.7 & 21.6 & & 85.0 $\pm$ 0.7 & 18.4\\ 
\cline{2-11}

& \multirow{2}{*}{VGG19} & 1\% & \multirow{2}{*}{90.5} & 36.0 $\pm$ 12.0 & \multirow{2}{*}{89.2} & 75.1 $\pm$ ~17 & 39.1 & \multirow{2}{*}{86.1} & 83.0 $\pm$ 3.4 & 47.0 \\
&  & 0.5\% & & 64.2 $\pm$ 12.4 & & 89.0 $\pm$ 0.2 & 24.8 & & 85.9 $\pm$ 0.4 & 21.7\\ 
\hline

\end{tabular}
    \begin{tablenotes}
      \small
      \item {\textit{Note}. SM: source model, used for training generators; TM: target model, used for testing generators; BER: the bit-error rate of the target model; CA ($\%$): clean accuracy; PA ($\%$): perturbed accuracy; NF: NeuralFuse; and RP: total recovery percentage of PA (NF) vs. PA}
    \end{tablenotes}
\end{threeparttable}
\end{adjustbox}\vspace{-1mm}
\end{table*}

%% file: assets/_tables/low_precision_bit_errors/cifar10.tex
\begin{table*}[ht]
    \centering
    \caption{Reduced-precision Quantization and with 0.5\% BER on CIFAR-10 pre-trained models.}
    \label{tab:appendix_low_precision_bit_errors_cifar10}
\begin{adjustbox}{max width=\linewidth}
\begin{threeparttable}
\begin{tabular}{c|c|c|c|crr|crr}
\toprule[1.5pt]
Base & \multirow{2}{*}{\#Bits} & \multirow{2}{*}{CA} & \multirow{2}{*}{PA} & \multicolumn{3}{c|}{ConvL (1\%)} & \multicolumn{3}{c}{UNetL (1\%)} \\
Model &  &  &  & CA (NF)  & PA (NF) & RP    & CA (NF)  & PA (NF) & RP  \\
\midrule
\multirow{7}[0]{*}{ResNet18} & 8     & 92.6  & 70.1 $\pm$ 11.6 & 89.8  & 89.5 $\pm$ ~~0.2 & 19.4  & 86.6  & 86.2 $\pm$ ~~0.3 & 16.1 \\
      & 7     & 92.5  & 68.8 $\pm$ 10.4 & 89.8  & 89.5 $\pm$ ~~1.7 & 20.7  & 86.5  & 86.0 $\pm$ ~~0.5 & 17.2 \\
      & 6     & 92.6  & 68.4 $\pm$ 11.2 & 89.7  & 89.5 $\pm$ ~~0.2 & 21.1  & 86.6  & 85.9 $\pm$ ~~0.3 & 17.5 \\
      & 5     & 92.4  & 52.7 $\pm$ 14.1 & 89.7  & 90.0 $\pm$ ~~0.7 & 37.3  & 86.5  & 85.5 $\pm$ ~~0.8 & 32.8 \\
      & 4     & 91.8  & 26.3 $\pm$ 12.7 & 89.8  & 58.7 $\pm$ 24.5 & 32.4  & 86.6  & 64.9 $\pm$ 22.5 & 38.6 \\
      & 3     & 84.8  & 11.3 $\pm$ ~~1.8 & 89.8  & 12.8 $\pm$ ~~5.8 & 1.5   & 86.0  & 14.8 $\pm$ 10.0 & 3.5 \\
      & 2     & 10.0  & 10.0 $\pm$ ~~0.0 & 10.0  & 10.0 $\pm$ ~~0.0 & 0.0   & 10.0  & 10.0 $\pm$ ~~0.0 & 0.0 \\
\midrule
\multirow{7}[0]{*}{VGG19} & 8     & 90.5  & 64.2 $\pm$ 12.4 & 89.8  & 89.6 $\pm$ ~~8.7 & 25.4  & 87.4  & 86.8 $\pm$ ~~0.4 & 22.6 \\
      & 7     & 90.3  & 66.5 $\pm$ ~~8.5 & 89.8  & 89.6 $\pm$ ~~0.2 & 23.1  & 87.4  & 86.7 $\pm$ ~~0.3 & 20.2 \\
      & 6     & 90.1  & 59.8 $\pm$ 13.2 & 89.9  & 89.4 $\pm$ ~~3.8 & 29.6  & 87.4  & 86.4 $\pm$ ~~0.7 & 26.6 \\
      & 5     & 90.2  & 37.7 $\pm$ 14.1 & 89.8  & 78.0 $\pm$ 15.8 & 40.3  & 87.2  & 79.8 $\pm$ ~~0.8 & 42.1 \\
      & 4     & 87.5  & 14.7 $\pm$ ~~6.0 & 89.8  & 27.8 $\pm$ 18.9 & 13.1  & 87.2  & 34.4 $\pm$ 20.5 & 19.7 \\
      & 3     & 78.3  & 10.5 $\pm$ ~~1.5 & 89.7  & 10.9 $\pm$ ~~2.6 & 0.4   & 86.8  & 11.0 $\pm$ ~~2.9 & 0.5 \\
      & 2     & 10.0  & 10.0 $\pm$ ~~0.0 & 10.0  & 10.0 $\pm$ ~~0.0 & 0.0   & 10.0  & 10.0 $\pm$ ~~0.0 & 0.0 \\
\bottomrule[1.5pt]
\end{tabular}%
    \begin{tablenotes}
      \small
      \item {\textit{Note}. CA ($\%$): clean accuracy; PA ($\%$): perturbed accuracy; NF: NeuralFuse; and RP: total recovery percentage of PA (NF) vs. PA}
    \end{tablenotes}
\end{threeparttable}
\end{adjustbox}
\vspace{-8mm}
\end{table*}

%% file: assets/_figures/low_precision_quantization/cifar10.tex
\begin{figure}[ht]
     \centering
     \hfill
     \begin{subfigure}[b]{\textwidth}
         \centering
         \includegraphics[width=.8\textwidth, trim=-0.2cm 0cm 0.2cm 0.1cm, clip]{assets/figures_paths/precision_loss_and_biterror/legend.pdf}
     \end{subfigure}
     \hfill
     \begin{subfigure}[b]{0.45\textwidth}
         \centering
         \includegraphics[width=\textwidth]{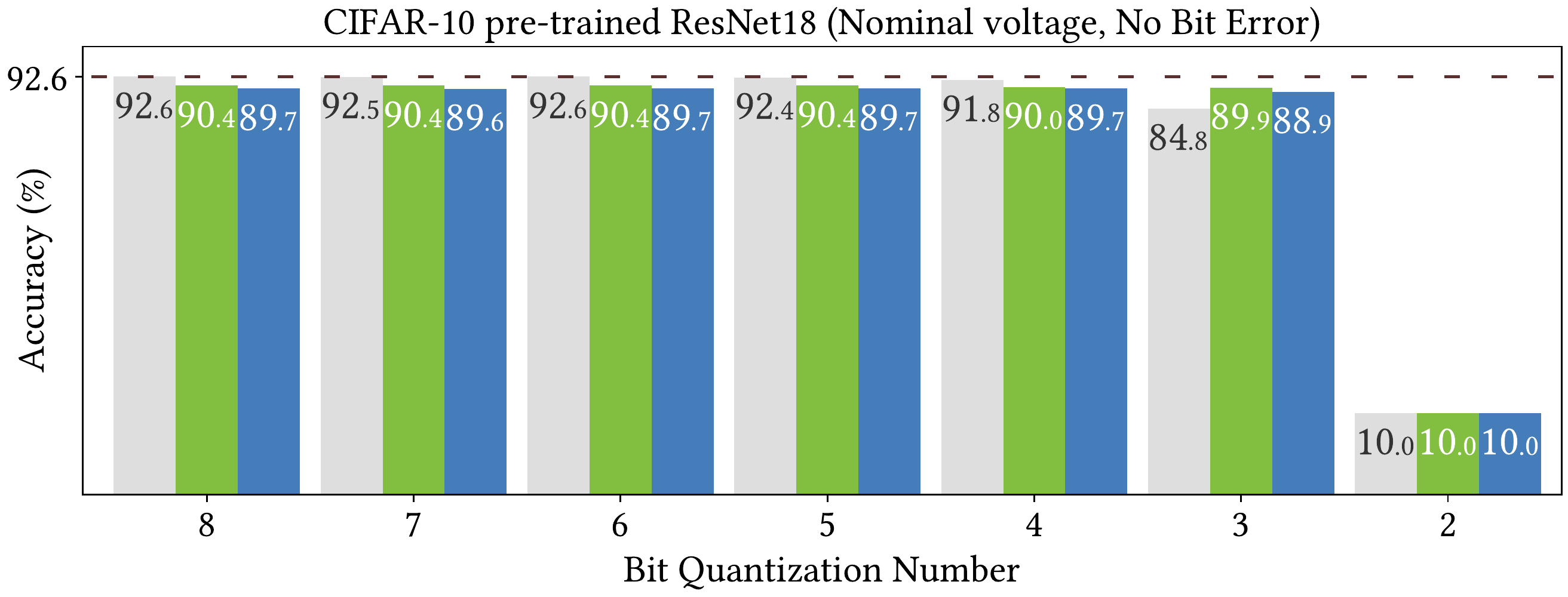}
         \caption{Base Model: ResNet18, no bit error.}
     \end{subfigure}
     \hspace{5mm}
     \begin{subfigure}[b]{0.45\textwidth}
         \centering
         \includegraphics[width=\textwidth]{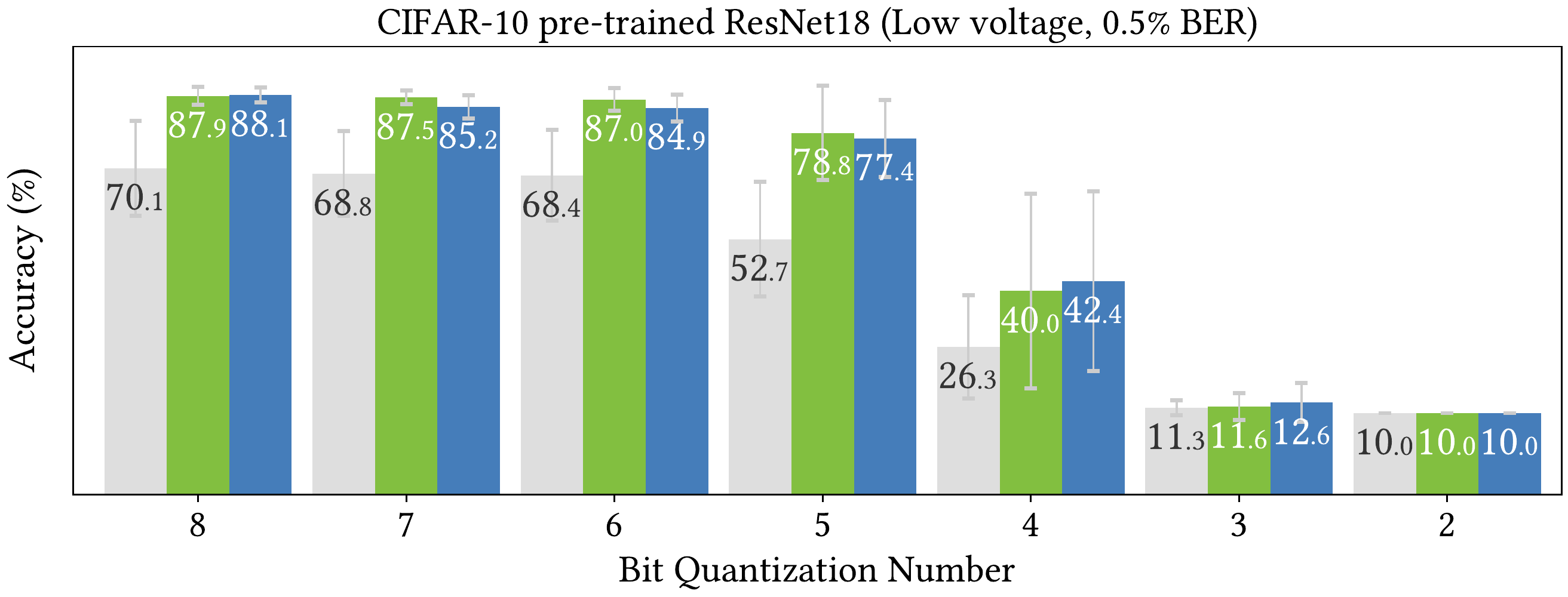}
         \caption{Base Model: ResNet18, 0.5\% B.E.R.}
     \end{subfigure}
     \hfill
     \newline
     \hfill
     \begin{subfigure}[b]{0.45\textwidth}
         \centering
         \includegraphics[width=\textwidth]{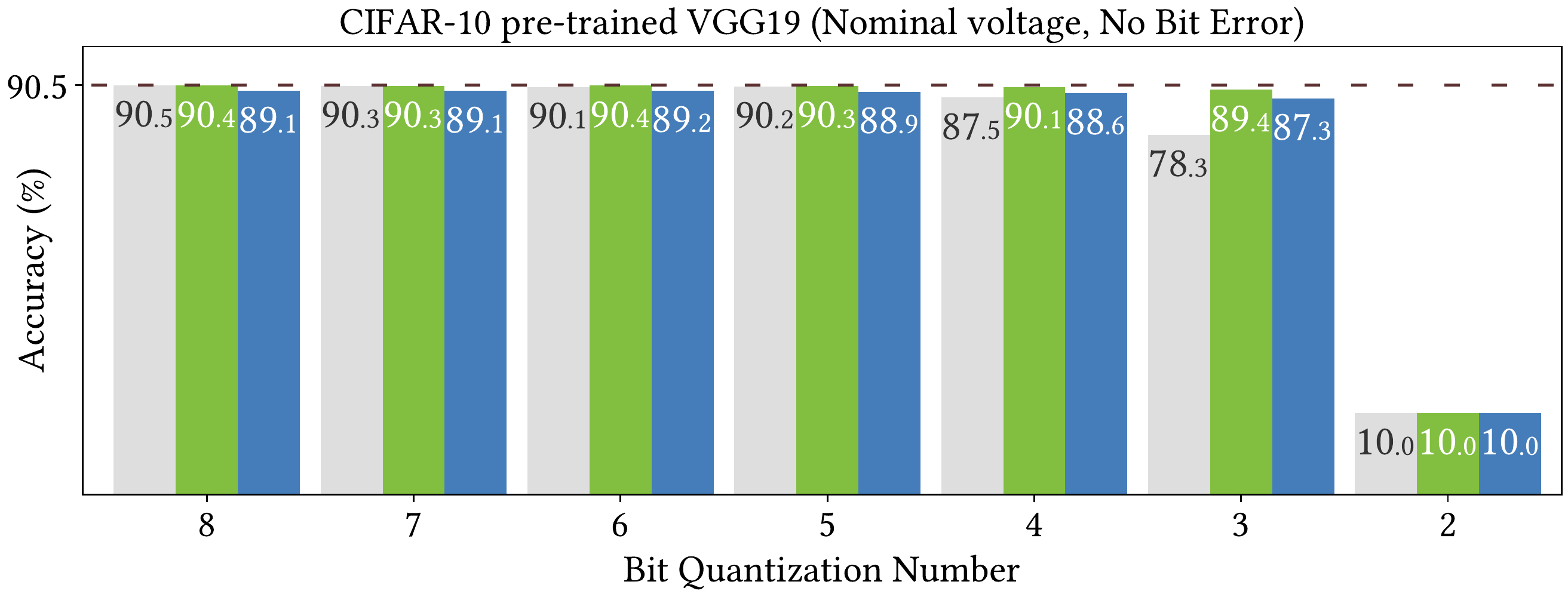}
         \caption{Base Model: VGG19, no bit error.}\vspace{-2mm}
     \end{subfigure}
     \hspace{5mm}
     \begin{subfigure}[b]{0.45\textwidth}
         \centering
         \includegraphics[width=\textwidth]{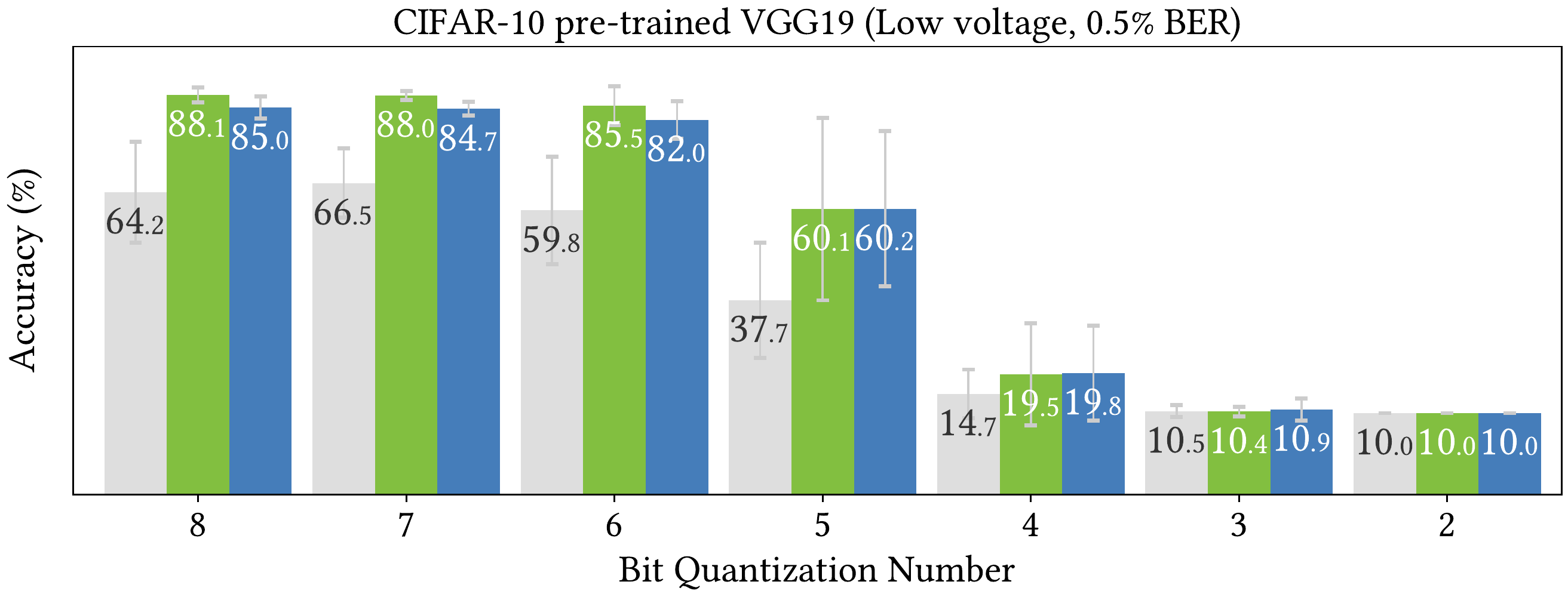}
         \caption{Base Model: VGG19, 0.5\% B.E.R.}\vspace{-2mm}
     \end{subfigure}
     \hfill
     \newline\vspace{-2mm}
        \caption{Results of Reduced-precision and bit errors (0.5\%) on CIFAR-10 pre-trained base models.}
        \label{fig:appendix_low_precision_cifar10}
\end{figure}

%% file: assets/_tables/low_precision_bit_errors/gtsrb.tex
\begin{table*}[ht]
\setlength{\tabcolsep}{3mm}
    \centering
    \caption{Reduced-precision Quantization and with 0.5\% BER on GTSRB pre-trained models.}
    \label{tab:appendix_low_precision_bit_errors_gtsrb}
\begin{adjustbox}{max width=\linewidth}
\begin{threeparttable}
\begin{tabular}{c|c|c|c|crr|crr}
\toprule[1.5pt]
Base & \multirow{2}{*}{\#Bits} & \multirow{2}{*}{CA} & \multirow{2}{*}{PA} & \multicolumn{3}{c|}{ConvL (1\%)} & \multicolumn{3}{c}{UNetL (1\%)} \\
Model &  &  &  & CA (NF)  & PA (NF) & RP    & CA (NF)  & PA (NF) & RP  \\
\midrule
\multirow{7}[0]{*}{ResNet18} & 8     & 95.5  & 75.2 $\pm$ 12.7 & 95.7  & 95.3 $\pm$ ~~0.5 & 20.1  & 96.2  & 95.7 $\pm$ ~~0.3 & 20.5 \\
      & 7     & 95.5  & 69.5 $\pm$ 10.6 & 95.7  & 95.3 $\pm$ ~~0.3 & 25.8  & 96.2  & 95.9 $\pm$ ~~0.3 & 26.4 \\
      & 6     & 95.4  & 67.2 $\pm$ 14.4 & 95.7  & 95.2 $\pm$ ~~0.5 & 28.0  & 96.2  & 95.7 $\pm$ ~~0.5 & 28.5 \\
      & 5     & 95.4  & 48.6 $\pm$ 18.2 & 95.8  & 92.6 $\pm$ ~~5.1 & 44.0  & 96.2  & 94.8 $\pm$ ~~2.5 & 46.2 \\
      & 4     & 92.6  & 24.6 $\pm$ ~~9.8 & 95.9  & 75.6 $\pm$ 16.2 & 51.0  & 96.2  & 86.6 $\pm$ ~~9.5 & 62.0 \\
      & 3     & 67.7  & ~~5.3 $\pm$ ~~3.5 & 95.4  & 18.4 $\pm$ 15.3 & 13.1  & 96.2  & 25.3 $\pm$ 22.5 & 20.0 \\
      & 2     & ~~3.8   & ~~3.8 $\pm$ ~~0.0 & ~~4.1   & 3.8 $\pm$ ~~0.0 & 0.0  & ~~3.8   & 3.8 $\pm$ ~~0.0 & 0.0 \\
\midrule
\multirow{7}[0]{*}{VGG19} & 8     & 95.5  & 69.1 $\pm$ 11.1 & 96.0  & 94.0 $\pm$ ~~2.2 & 24.9  & 95.4  & 93.9 $\pm$ ~~2.1 & 24.8 \\
      & 7     & 95.6  & 66.1 $\pm$ 14.8 & 96.0  & 92.2 $\pm$ ~~5.7 & 26.1  & 95.4  & 92.6 $\pm$ ~~3.7 & 26.5 \\
      & 6     & 95.3  & 64.2 $\pm$ ~~8.4 & 96.0  & 92.2 $\pm$ ~~5.7 & 28.0  & 95.4  & 92.3 $\pm$ ~~2.3 & 28.1 \\
      & 5     & 95.2  & 48.2 $\pm$ 14.0 & 96.0  & 92.2 $\pm$ ~~5.7 & 44.0  & 95.4  & 86.2 $\pm$ ~~8.4 & 38.0 \\
      & 4     & 92.0  & 18.2 $\pm$ 14.3 & 93.0  & 92.2 $\pm$ ~~5.7 & 74.0  & 95.0  & 49.6 $\pm$ 22.8 & 31.4 \\
      & 3     & 60.0  & ~~2.0 $\pm$ ~~0.9 & 87.3  & 92.2 $\pm$ ~~5.7 & 90.2  & 87.2  & 1.7 $\pm$ ~~0.9 & -0.3 \\
      & 2     & ~~5.9   & ~~3.8 $\pm$ ~~0.0 & ~~5.9   & 3.8 $\pm$ ~~0.0 & 0.0   & ~~5.9   & 3.8 $\pm$ ~~0.0 & 0.0 \\
\bottomrule[1.5pt]
\end{tabular}%
    \begin{tablenotes}
      \small
      \item {\textit{Note}. CA ($\%$): clean accuracy; PA ($\%$): perturbed accuracy; NF: NeuralFuse; and RP: total recovery percentage of PA (NF) vs. PA}
    \end{tablenotes}
\end{threeparttable}
\end{adjustbox}\vspace{-2mm}
\end{table*}

%% file: assets/_figures/low_precision_quantization/gtsrb.tex
\begin{figure}[ht]
     \centering
     \hfill
     \begin{subfigure}[b]{\textwidth}
         \centering
         \includegraphics[width=.8\textwidth, trim=-0.2cm 0cm 0.2cm 0.1cm, clip]{assets/figures_paths/precision_loss_and_biterror/legend.pdf}
     \end{subfigure}
     \hfill
     \begin{subfigure}[b]{0.45\textwidth}
         \centering
         \includegraphics[width=\textwidth]{assets/figures_paths/precision_loss_and_biterror/GTSRB_ResNet18.pdf}
         \caption{Base Model: ResNet18, no bit error.}
     \end{subfigure}
     \hspace{5mm}
     \begin{subfigure}[b]{0.45\textwidth}
         \centering
         \includegraphics[width=\textwidth]{assets/figures_paths/precision_loss_and_biterror/GTSRB_ResNet18_5ber.pdf}
         \caption{Base Model: ResNet18, 0.5\% B.E.R.}
     \end{subfigure}
     \newline
     \begin{subfigure}[b]{0.45\textwidth}
         \centering
         \includegraphics[width=\textwidth]{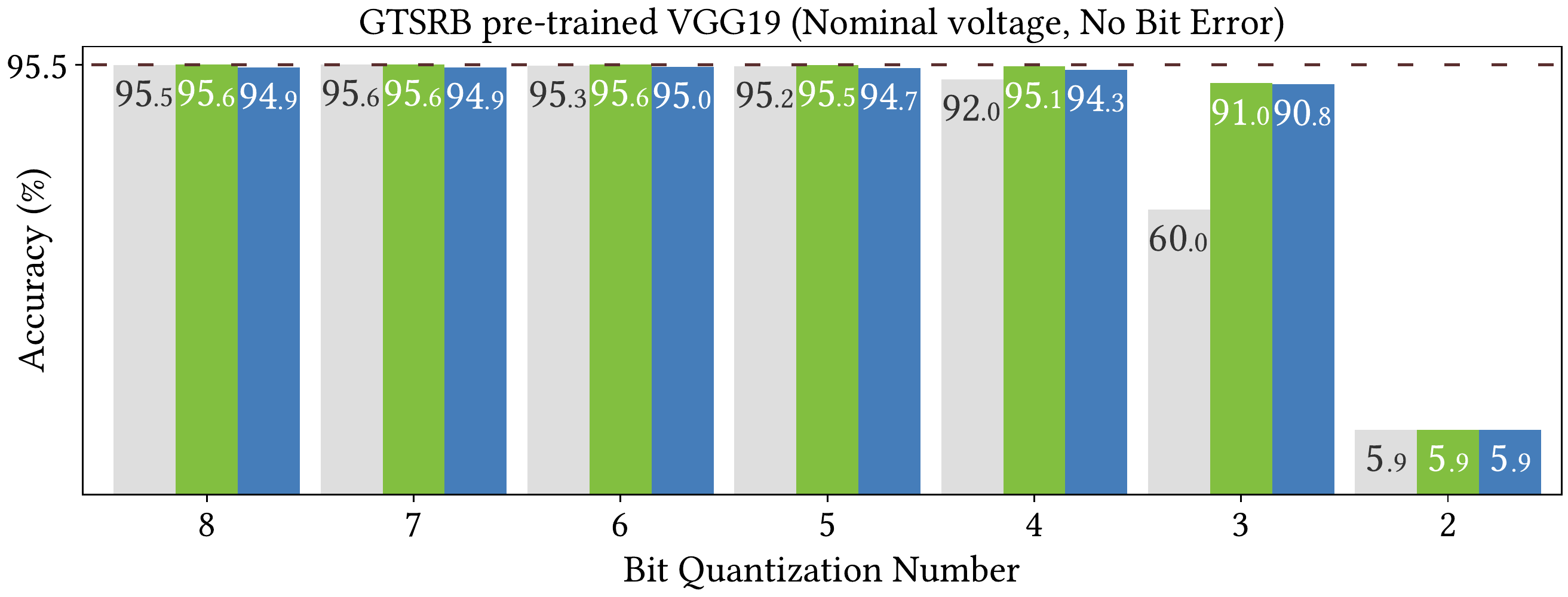}
         \caption{Base Model: VGG19, no bit error.}
     \end{subfigure}
     \hspace{5mm}
     \begin{subfigure}[b]{0.45\textwidth}
         \centering
         \includegraphics[width=\textwidth]{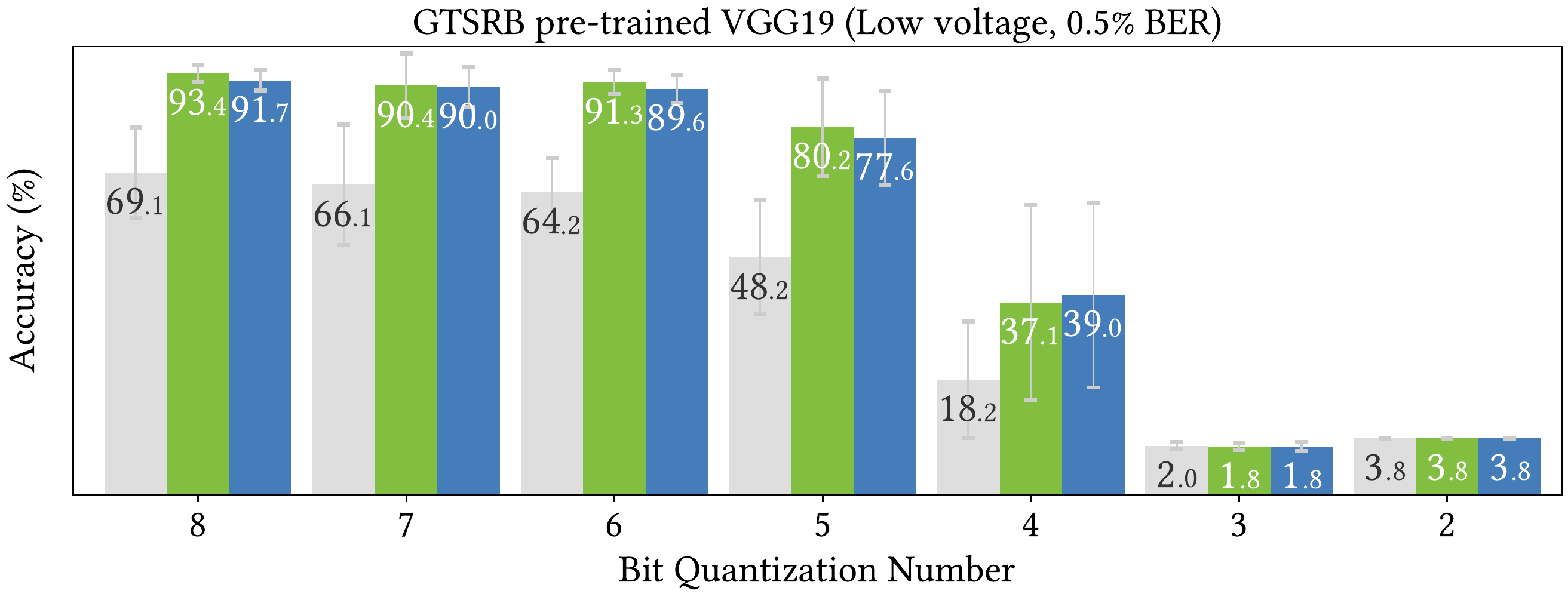}
         \caption{Base Model: VGG19, 0.5\% B.E.R.}
     \end{subfigure}\newline
\vspace{-5mm}
        \caption{Results of Reduced-precision and bit errors (0.5\%) on GTSRB pre-trained base models.}
        \label{fig:appendix_low_precision_gtsrb}
\vspace{-25mm}
\end{figure}

%% file: assets/_tables/low_precision_bit_errors/imagenet10.tex
\begin{table*}[ht]
    \centering
    \caption{Reduced-precision Quantization and with 0.5\% BER on ImageNet-10 pre-trained models.}
    \label{tab:appendix_low_precision_bit_errors_imagenet10}

\begin{adjustbox}{max width=\linewidth}
\begin{threeparttable}
\begin{tabular}{c|c|c|c|crr|crr}
\toprule[1.5pt]
Base & \multirow{2}{*}{\#Bits} & \multirow{2}{*}{CA} & \multirow{2}{*}{PA} & \multicolumn{3}{c|}{ConvL (0.5\%)} & \multicolumn{3}{c}{UNetL (0.5\%)} \\
Model &  &  &  & CA (NF)  & PA (NF) & RP    & CA (NF)  & PA (NF) & RP  \\
\midrule
\multirow{7}[0]{*}{ResNet18} & 8     & 92.2  & 72.3 $\pm$ ~~7.0 & 94.0  & 88.0 $\pm$ ~~2.0 & 15.7  & 94.0  & 88.1 $\pm$ ~~1.4 & 15.8 \\
      & 7     & 92.4  & 70.6 $\pm$ 13.0 & 94.2  & 86.7 $\pm$ ~~4.1 & 16.1  & 93.6  & 87.8 $\pm$ ~~3.5 & 17.2 \\
      & 6     & 92.4  & 68.9 $\pm$ ~~9.9 & 94.2  & 85.1 $\pm$ ~~4.8 & 16.2  & 93.6  & 86.4 $\pm$ ~~3.7 & 17.5 \\
      & 5     & 91.0  & 60.9 $\pm$ 13.0 & 94.2  & 82.5 $\pm$ ~~6.8 & 21.6  & 94.0  & 83.2 $\pm$ ~~5.9 & 22.3 \\
      & 4     & 91.4  & 47.4 $\pm$ ~~9.8 & 93.8  & 68.6 $\pm$ ~~9.8 & 21.2  & 92.6  & 68.7 $\pm$ ~~9.2 & 21.3 \\
      & 3     & 85.2  & 28.8 $\pm$ 11.8 & 89.2  & 44.1 $\pm$ 14.0 & 15.3  & 89.4  & 42.7 $\pm$ 14.2 & 13.9 \\
      & 2     & 10.0  & 10.0 $\pm$ ~~0.0 & 10.0  & 10.0 $\pm$ ~~0.0 & 0.0   & 10.0  & 10.0 $\pm$ ~~0.0 & 0.0 \\
\midrule
\multirow{7}[0]{*}{VGG19} & 8     & 92.4  & 37.2 $\pm$ 11.0 & 91.4  & 75.5 $\pm$ ~~8.8 & 38.3  & 89.4  & 77.9 $\pm$ ~~6.1 & 40.7 \\
      & 7     & 92.0  & 27.3 $\pm$ ~~6.6 & 91.2  & 59.3 $\pm$ 13.0 & 32.0  & 89.4  & 65.4 $\pm$ 10.0 & 38.1 \\
      & 6     & 92.4  & 27.9 $\pm$ ~~6.4 & 91.0  & 59.7 $\pm$ 11.8 & 31.8  & 89.4  & 64.9 $\pm$ ~~9.9 & 37.0 \\
      & 5     & 92.0  & 15.1 $\pm$ ~~4.4 & 91.6  & 23.1 $\pm$ ~~0.7 & 8.0   & 89.0  & 27.9 $\pm$ ~~8.8 & 12.8 \\
      & 4     & 89.4  & 12.2 $\pm$ ~~2.7 & 90.8  & 14.0 $\pm$ ~~4.3 & 1.8   & 89.6  & 14.6 $\pm$ ~~4.9 & 2.4 \\
      & 3     & 46.8  & ~~9.9 $\pm$ ~~0.5 & 83.2  & 10.4 $\pm$ ~~0.6 & 0.5   & 84.2  & 9.9 $\pm$ ~~0.7 & 0.0 \\
      & 2     & 10.0  & 10.0 $\pm$ ~~0.0 & 10.0  & 10.0 $\pm$ ~~0.0 & 0.0   & 10.0  & 10.0 $\pm$ ~~0.0 & 0.0 \\
\bottomrule[1.5pt]
\end{tabular}%
    \begin{tablenotes}
      \small
      \item {\textit{Note}. CA ($\%$): clean accuracy; PA ($\%$): perturbed accuracy; NF: NeuralFuse; and RP: total recovery percentage of PA (NF) vs. PA}
    \end{tablenotes}
\end{threeparttable}
\end{adjustbox}\vspace{-6mm}
\end{table*}

%% file: assets/_figures/low_precision_quantization/imagenet10.tex
\begin{figure}[ht]
     \centering
     \hfill
     \begin{subfigure}[b]{\textwidth}
         \centering
         \includegraphics[width=.8\textwidth, trim=-0.2cm -0.1cm 0.2cm 0.1cm, clip]{assets/figures_paths/precision_loss_and_biterror/legend.pdf}
     \end{subfigure}
    \hfill
     \begin{subfigure}[b]{0.45\textwidth}
         \centering
         \includegraphics[width=\textwidth]{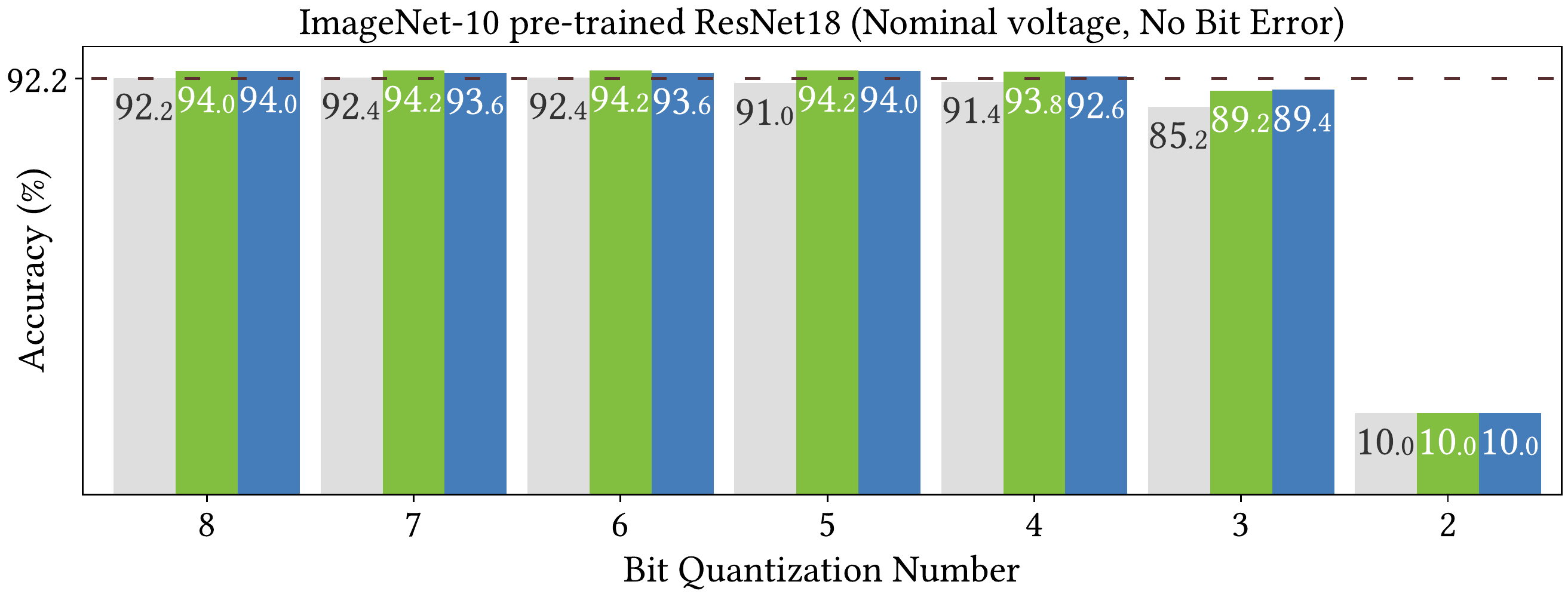}
         \caption{Base Model: ResNet18, no bit error.}
     \end{subfigure}
     \hspace{5mm}
     \begin{subfigure}[b]{0.45\textwidth}
         \centering
         \includegraphics[width=\textwidth]{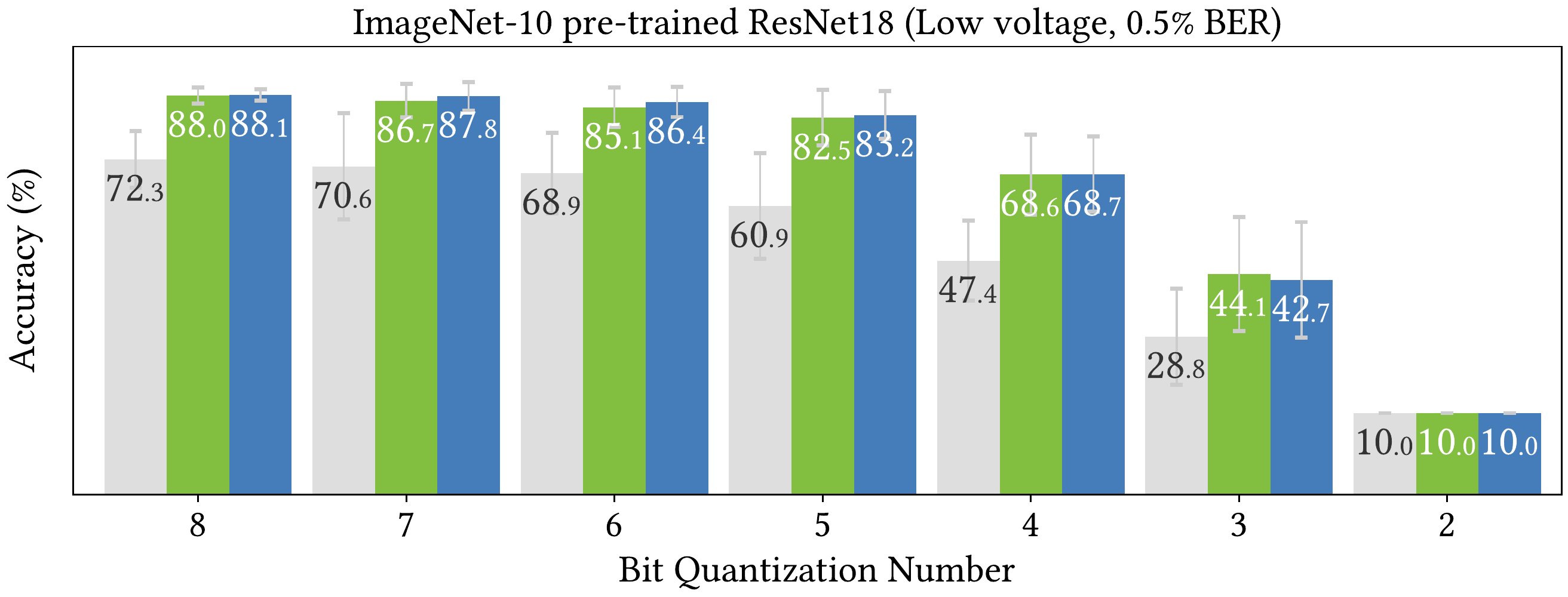}
         \caption{Base Model: ResNet18, 0.5\% B.E.R.}
     \end{subfigure}
     \newline\vspace{2mm}
     \begin{subfigure}[b]{0.45\textwidth}
         \centering
         \includegraphics[width=\textwidth]{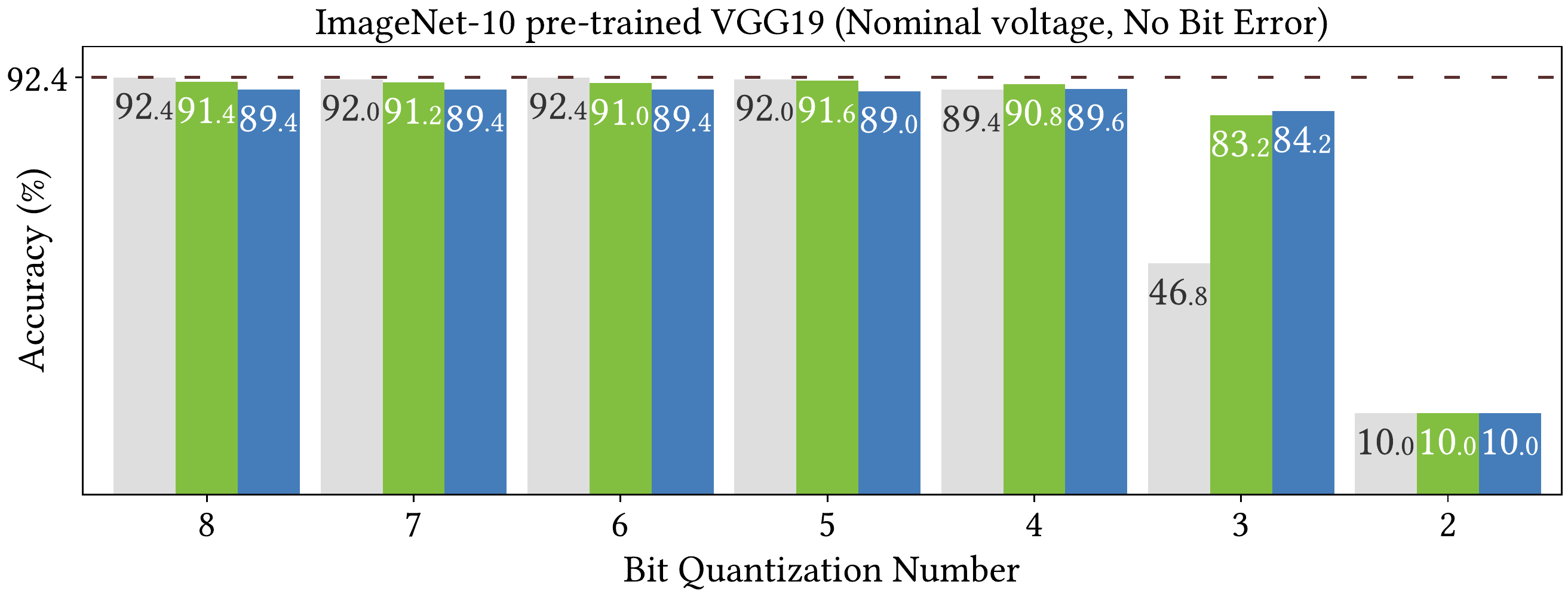}
         \caption{Base Model: VGG19, no bit error.}
     \end{subfigure}
     \hspace{5mm}
     \begin{subfigure}[b]{0.45\textwidth}
         \centering
         \includegraphics[width=\textwidth]{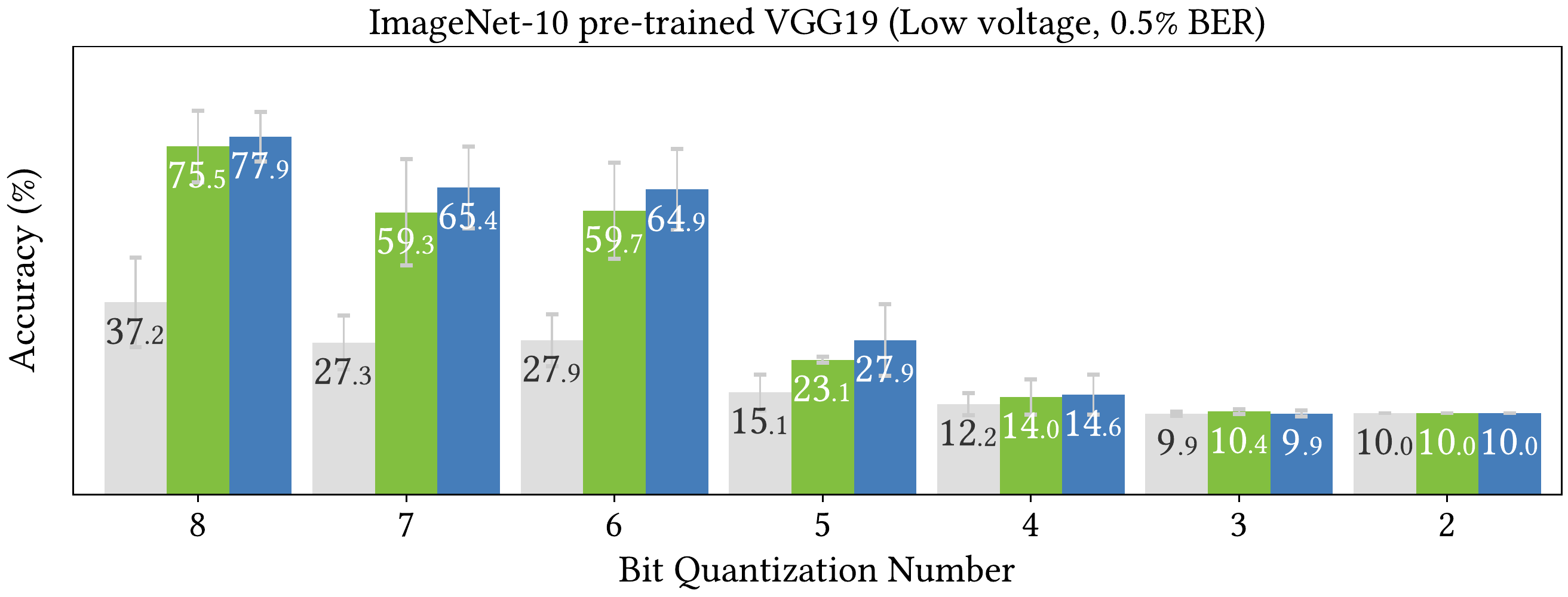}
         \caption{Base Model: VGG19, 0.5\% B.E.R.}
     \end{subfigure}\newline\vspace{-5mm}
        \caption{Results of Reduced-precision and bit errors (0.5\%) on ImageNet-10 pre-trained base models.}
     \label{fig:appendix_low_precision_imagenet10}
     \vspace{-5mm}
\end{figure}

%% file: assets/_tables/_tables_Appendix_AdversarialTraining.tex
\begin{table*}[ht]
\centering
\caption{Performance of the generator trained by adversarial training under K flip bits on ResNet18 with CIFAR-10. The results show that the generator trained by adversarial training cannot achieve high accuracy against bit errors under a $1\%$ bit error rate.}
\label{table:AppendixAdvTrainingp1CIFAR10}
    \begin{adjustbox}{max width=\linewidth}
    \begin{threeparttable}
        \begin{tabular}{ r|cc|ccr }
            \toprule[1.5pt]
            K-bits & CA & PA & CA (NF) & PA (NF) & RP \\ 
            \midrule
            100 & \multirow{5}{*}{92.6} & \multirow{5}{*}{38.9 $\pm$ 12.4} & 92.4 & 38.3 $\pm$ 12.1 & -0.6 \\ 
            500 & & &  92.1 & 38.7 $\pm$ 12.5 & -0.2 \\
            5,000 & & & 92.6 & 38.9 $\pm$ 12.5 & ~~0 \\ 
            20,000 & & & 60.1 & 23.0 $\pm$ ~8.1 & -16 \\
            100,000 & & & 71.1 & 23.6 $\pm$ ~6.6 & -16 \\
            \bottomrule[1.5pt]
        \end{tabular}
    \begin{tablenotes}
      \small
      \item {\textit{Note}. CA ($\%$): clean accuracy; PA ($\%$): perturbed accuracy; NF: NeuralFuse; and RP: total recovery percentage of PA (NF) vs. PA}
    \end{tablenotes}
    \end{threeparttable}
    \end{adjustbox}
\end{table*}

%% file: assets/_tables/_tables_Appendix_AppendixRobustModelPreResnet18CIFAR10.tex
\begin{table*}[ht]
\centering
\begin{threeparttable}
\caption{Performance of NeuralFuse trained with rubust CIFAR-10 pre-trained PreAct ResNet18. The results show that NeuralFuse can be used together with a robust model and further improve perturbed accuracy under both 1\% and 0.5\% BER}
\label{table:AppendixRobustCifar10Table}

\begin{tabular}{ c|c|c|cc|crr }
\toprule[1.5pt] 

Base Model & BER & NF &  CA & PA & CA (NF) & PA (NF) & RP  \\
\midrule 

\multirow{12}{*}{\begin{tabular}{@{}c@{}}PreAct \\ ResNet18\end{tabular} } & \multirow{6}{*}{1\%} & ConvL & \multirow{6}{*}{89.7}  & \multirow{6}{*}{23.0 $\pm$ 9.3} & 87.6  & 53.7 $\pm$ ~26 & 30.7 \\
&& ConvS & 
& & 83.1 & 34.6 $\pm$ ~15 & 11.6 \\
&& DeConvL & 
& & 87.7 & 55.4 $\pm$ ~27 & 32.4 \\
&& DeConvS & 
& & 82.9 & 32.4 $\pm$ ~14 & 9.4 \\
&& UNetL & 
& & 86.1 & 60.4 $\pm$ ~28 & 37.4 \\
&& UNetS & 
& & 80.4 & 51.9 $\pm$ ~24 & 28.9 \\
\addlinespace
\cline{2-8}
\addlinespace

& \multirow{6}{*}{0.5\%} & ConvL & \multirow{6}{*}{89.7} & \multirow{6}{*}{63.2 $\pm$ 8.7} & 89.2  & 87.8 $\pm$ 1.1 & 24.6 \\
&& ConvS & 
& & 89.2 & 74.0 $\pm$ 6.5 & 10.8 \\
&& DeConvL & 
& & 89.0 & 87.4 $\pm$ 1.1 & 24.2 \\
&& DeConvS & 
& & 89.9 & 74.4 $\pm$ 7.0 & 11.2 \\
&& UNetL & 
& & 87.5 & 85.9 $\pm$ 0.8 & 22.7 \\
&& UNetS & 
& & 88.2 & 80.4 $\pm$ 3.9 & 17.2 \\
\bottomrule[1.5pt]
\end{tabular}
    \begin{tablenotes}
      \small
      \item {\textit{Note}. BER: the bit-error rate of the base model; CA ($\%$): clean accuracy; PA ($\%$): perturbed accuracy; NF: NeuralFuse; and RP: total recovery percentage of PA (NF) vs. PA}
    \end{tablenotes}
    \end{threeparttable}
\end{table*}

%% file: assets/_figures/visualization.tex
\begin{figure}[h!t]
     \centering
     \begin{subfigure}[b]{0.31\textwidth}
         \centering
         \includegraphics[width=\textwidth]{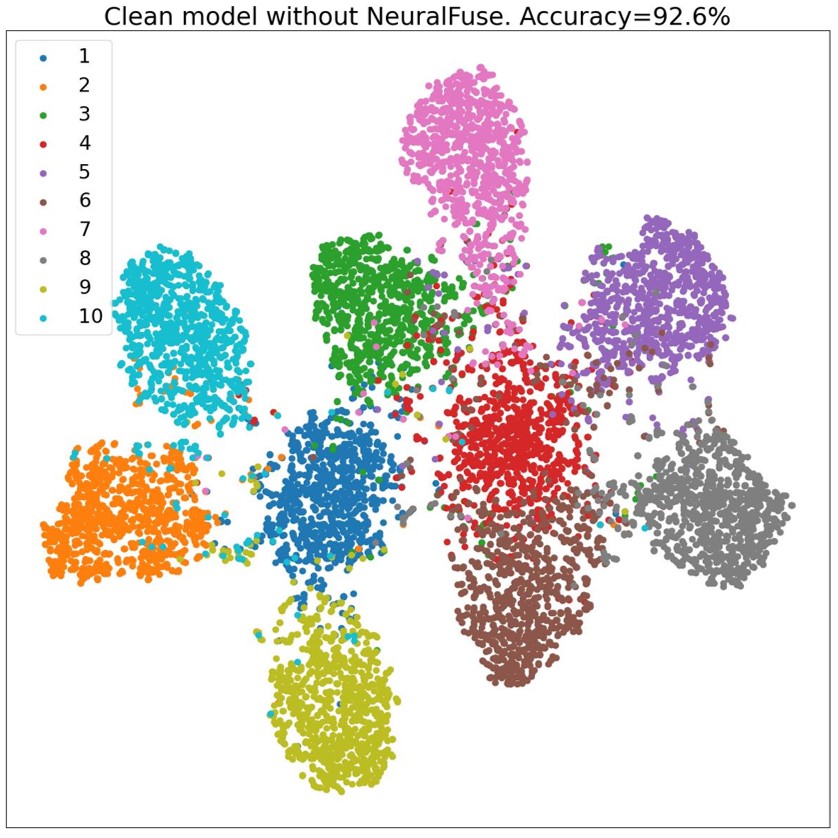}
         \caption{}
     \end{subfigure}
     \begin{subfigure}[b]{0.31\textwidth}
         \centering
         \includegraphics[width=\textwidth]{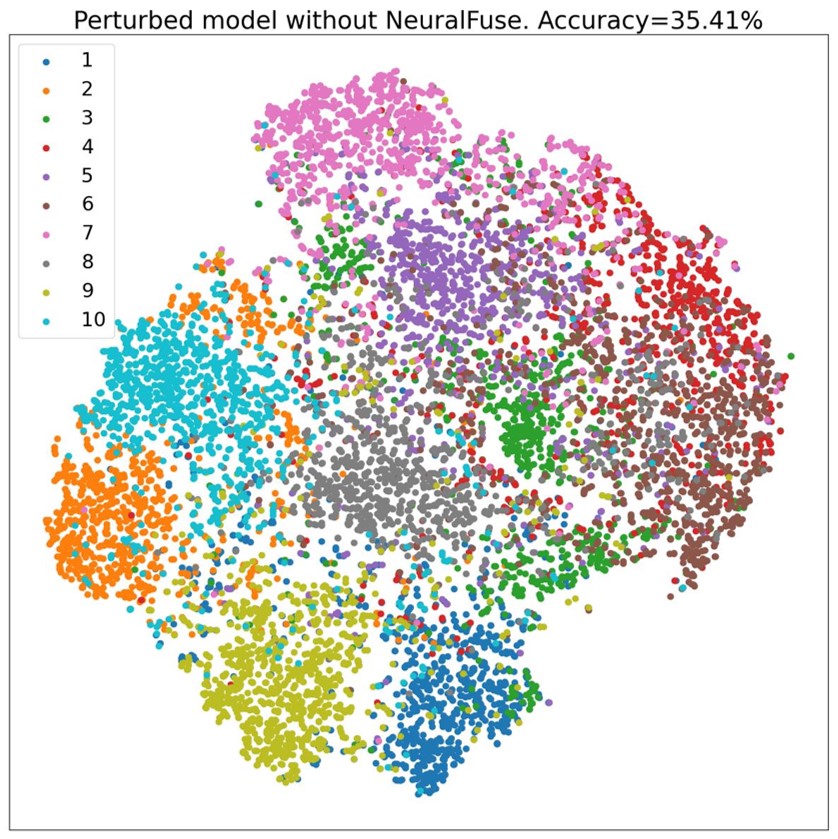}
         \caption{}
     \end{subfigure}
     \begin{subfigure}[b]{0.31\textwidth}
         \centering
         \includegraphics[width=\textwidth]{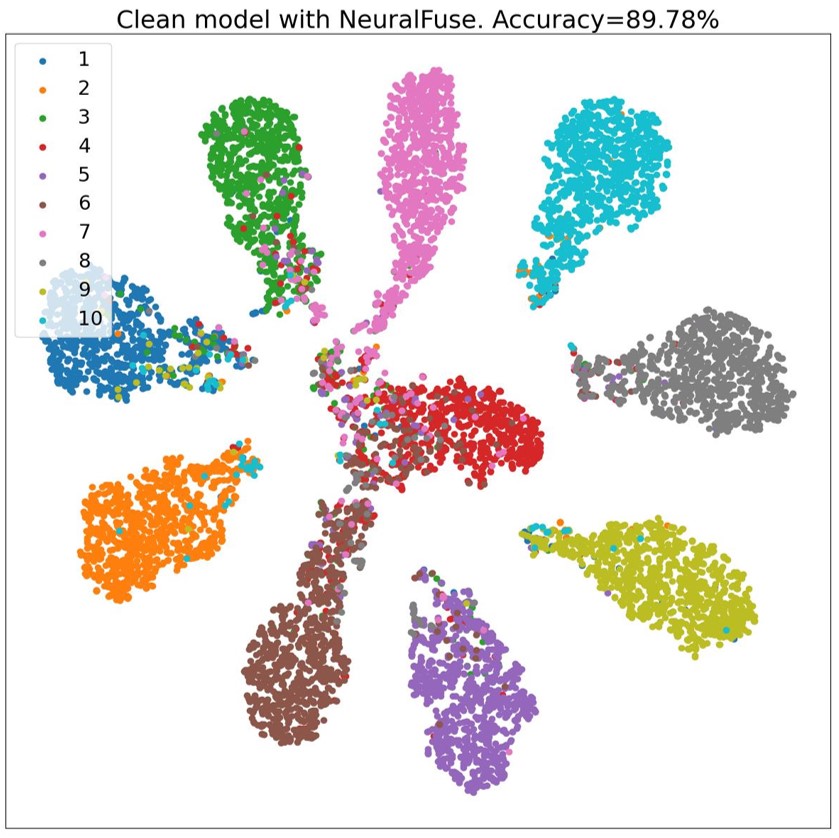}
         \caption{}
     \end{subfigure}
     \newpage
     \begin{subfigure}[b]{0.31\textwidth}
         \centering
         \includegraphics[width=\textwidth]{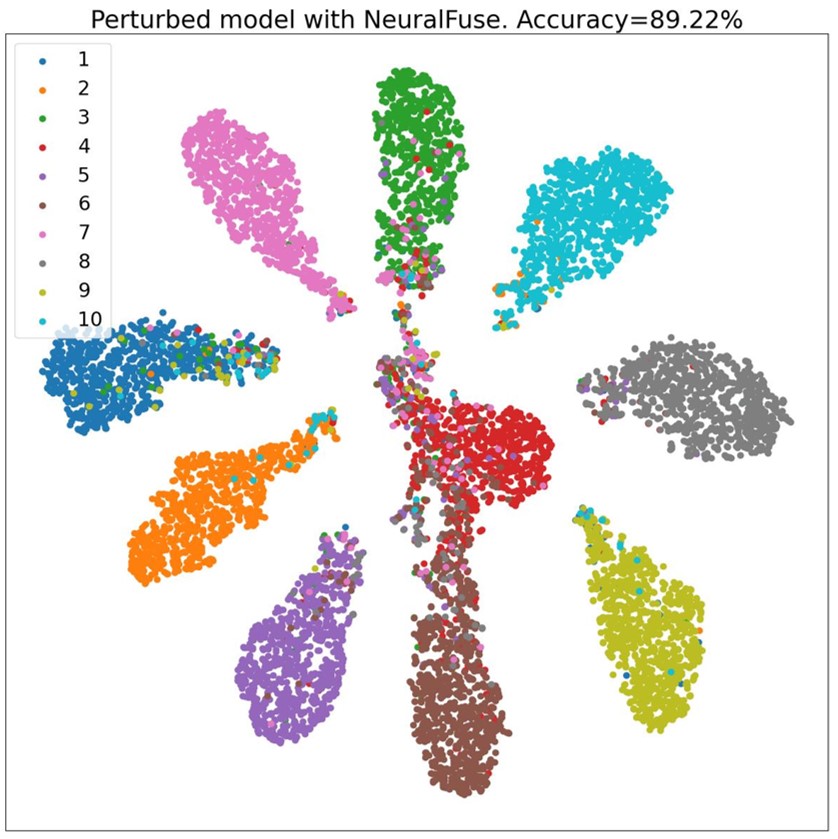}
         \caption{}
     \end{subfigure}
     \begin{subfigure}[b]{0.31\textwidth}
         \centering
         \includegraphics[width=\textwidth]{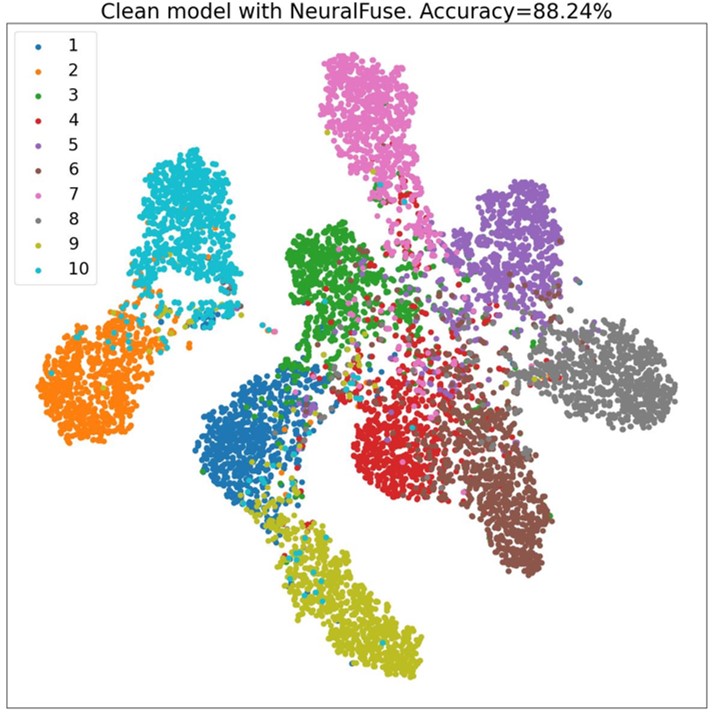}
         \caption{}
     \end{subfigure}
     \begin{subfigure}[b]{0.31\textwidth}
         \centering
         \includegraphics[width=\textwidth]{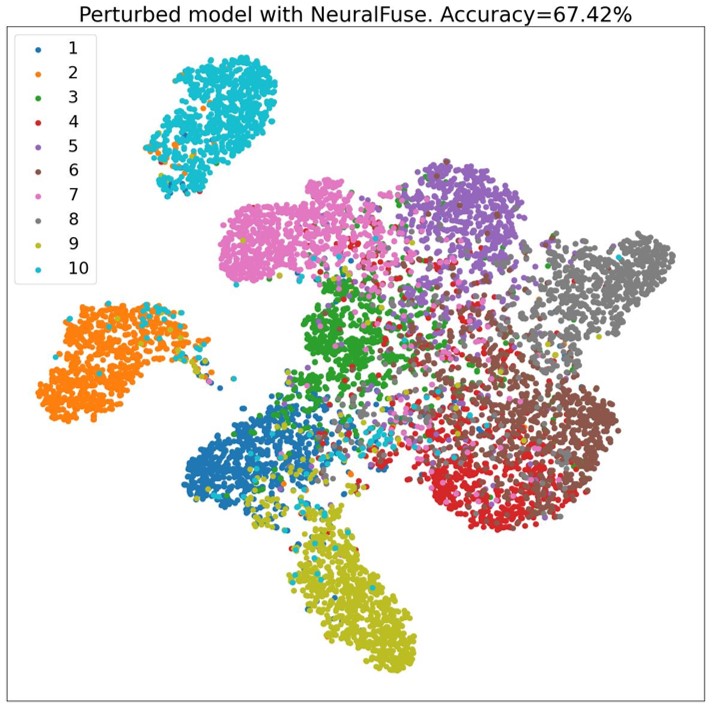}
         \caption{}
     \end{subfigure}
        \caption{t-SNE results for ResNet18 trained by CIFAR-10 under $1\%$ of bit error rate. (a) Clean model. (b) Perturbed model. (c) Clean model with ConvL. (d) Perturbed model with ConvL. (e) Clean model with ConvS. (f) Perturbed model with ConvS.}
        \label{fig:TSNE}
\end{figure}

%% file: assets/_figures/visualization_transformed.tex
\begin{figure*}[ht]
\centering
    \begin{adjustbox}{max width=\linewidth}
        \begin{tabular}{rlrl}
            \parbox[m]{1.5cm}{{\vspace{-7mm}\begin{flushright}Clean\end{flushright}}} &
            \includegraphics[width=.28\linewidth]{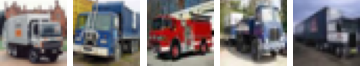} &
            \includegraphics[width=.28\linewidth]{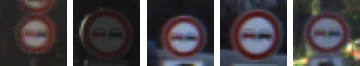} &
            \includegraphics[width=.28\linewidth]{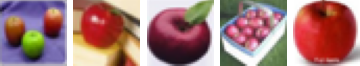} \\
            
            \parbox[m]{1.5cm}{{\vspace{-7mm}\begin{flushright}ConvL\end{flushright}}} &
            \includegraphics[width=.28\linewidth]{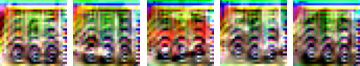} &
            \includegraphics[width=.28\linewidth]{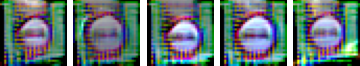} &
            \includegraphics[width=.28\linewidth]{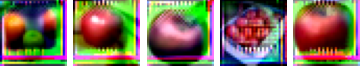} \\
            
            \parbox[m]{1.5cm}{{\vspace{-7mm}\begin{flushright}ConvS\end{flushright}}} &
            \includegraphics[width=.28\linewidth]{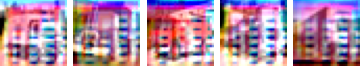}&
            \includegraphics[width=.28\linewidth]{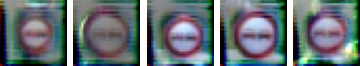} &
            \includegraphics[width=.28\linewidth]{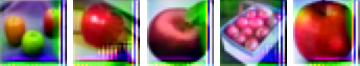} \\
            
            \parbox[m]{1.5cm}{{\vspace{-7mm}\begin{flushright}DeConvL\end{flushright}}} &
            \includegraphics[width=.28\linewidth]{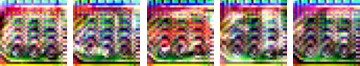} &
            \includegraphics[width=.28\linewidth]{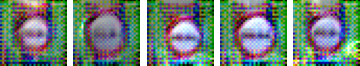} &
            \includegraphics[width=.28\linewidth]{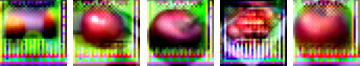} \\
            
            \parbox[m]{1.5cm}{{\vspace{-7mm}\begin{flushright}DeConvS\end{flushright}}} &
            \includegraphics[width=.28\linewidth]{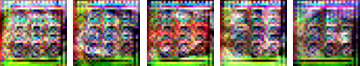} &
            \includegraphics[width=.28\linewidth]{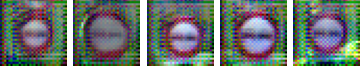} &
            \includegraphics[width=.28\linewidth]{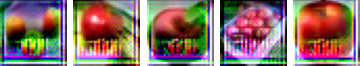} \\
            
            \parbox[m]{1.5cm}{{\vspace{-7mm}\begin{flushright}UNetL\end{flushright}}} &
            \includegraphics[width=.28\linewidth]{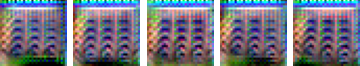} &
            \includegraphics[width=.28\linewidth]{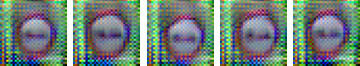} &
            \includegraphics[width=.28\linewidth]{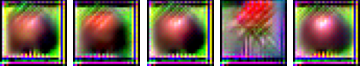}\\
            
            \parbox[m]{1.5cm}{{\vspace{-7mm}\begin{flushright}UNetS\end{flushright}}} &
            \subcaptionbox{\textit{Truck} class in CIFAR-10.}{\includegraphics[width=.28\linewidth]{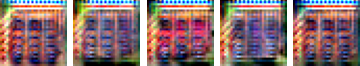}}&
            \subcaptionbox{\textit{No Passing} sign in GTSRB.}{\includegraphics[width=.28\linewidth]{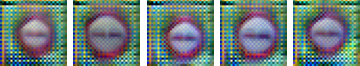}} &
            \subcaptionbox{\textit{Apple} class in CIFAR-100.}{\includegraphics[width=.28\linewidth]{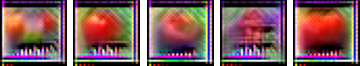}} \\
        \end{tabular}
    \end{adjustbox}
\caption{Visualization of transformed images from different NeuralFuse generators trained with ResNet18 at $1\%$ bit error rate.}
\label{fig:VisualizationImages}
\end{figure*}

%% file: assets/_figures/visualization_transformed_all.tex
\begin{figure*}[ht]
        \begin{tabular}{rlr}
            \parbox[m]{1.3cm}{{\vspace{-5mm}\begin{flushright}Clean\end{flushright}}} &
            \includegraphics[width=.42\linewidth]{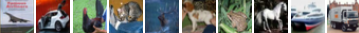} &
            \includegraphics[width=.42\linewidth]{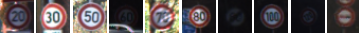} \\
            
            \parbox[m]{1.3cm}{{\vspace{-5mm}\begin{flushright}ConvL\end{flushright}}} &
            \includegraphics[width=.42\linewidth]{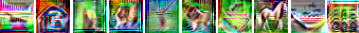} &
            \includegraphics[width=.42\linewidth]{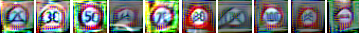} \\
            
            \parbox[m]{1.3cm}{{\vspace{-5mm}\begin{flushright}ConvS\end{flushright}}}&
            \includegraphics[width=.42\linewidth]{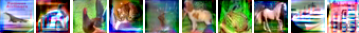}&
            \includegraphics[width=.42\linewidth]{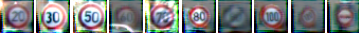} \\
            
            \parbox[m]{1.3cm}{{\vspace{-5mm}\begin{flushright}DeConvL\end{flushright}}} &
            \includegraphics[width=.42\linewidth]{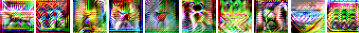} &
            \includegraphics[width=.42\linewidth]{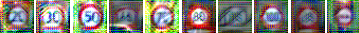} \\
            
            \parbox[m]{1.3cm}{{\vspace{-5mm}\begin{flushright}DeConvS\end{flushright}}} &
            \includegraphics[width=.42\linewidth]{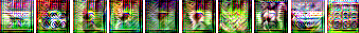} &
            \includegraphics[width=.42\linewidth]{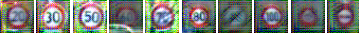} \\
            
            \parbox[m]{1.3cm}{{\vspace{-5mm}\begin{flushright}UNetL\end{flushright}}} &
            \includegraphics[width=.42\linewidth]{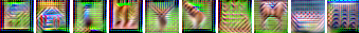} &
            \includegraphics[width=.42\linewidth]{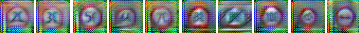} \\
            
            \parbox[m]{1.3cm}{{\vspace{-5mm}\begin{flushright}UNetS\end{flushright}}} &
            \subcaptionbox{10 classes sampled from CIFAR-10}{\includegraphics[width=.42\linewidth]{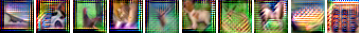}}&
            \subcaptionbox{10 traffic signs sampled from GTSRB}{\includegraphics[width=.42\linewidth]{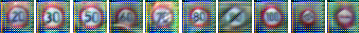}} \\
        \vspace{-2mm}
        \end{tabular}
        \begin{tabular}{cc}
            \parbox[m]{1.3cm}{{\vspace{-5mm}\begin{flushright}Clean\end{flushright}}}  &
            \includegraphics[width=0.865\linewidth]{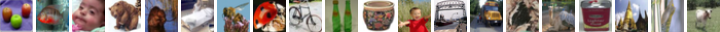} \\
            
            \parbox[m]{1.3cm}{{\vspace{-5mm}\begin{flushright}ConvL\end{flushright}}}  &
            \includegraphics[width=0.865\linewidth]{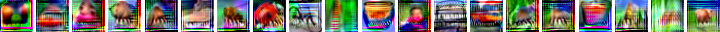} \\
            
            \parbox[m]{1.3cm}{{\vspace{-5mm}\begin{flushright}ConvS\end{flushright}}} &
            \includegraphics[width=0.865\linewidth]{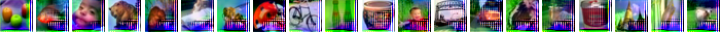} \\
            
            \parbox[m]{1.3cm}{{\vspace{-5mm}\begin{flushright}DeConvL\end{flushright}}}  &
            \includegraphics[width=0.865\linewidth]{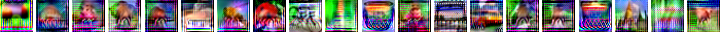} \\
            
            \parbox[m]{1.3cm}{{\vspace{-5mm}\begin{flushright}DeConvS\end{flushright}}}  &
            \includegraphics[width=0.865\linewidth]{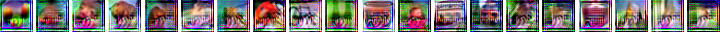} \\
            
            \parbox[m]{1.3cm}{{\vspace{-5mm}\begin{flushright}UNetL\end{flushright}}}  &
            \includegraphics[width=0.865\linewidth]{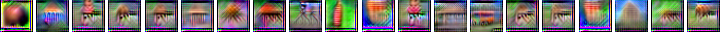} \\
            
            \parbox[m]{1.3cm}{{\vspace{-5mm}\begin{flushright}UNetS\end{flushright}}}  &
            {\subcaptionbox{20 classes sampled from CIFAR-100}{\includegraphics[width=0.865\linewidth]{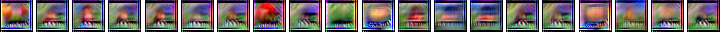}}} \\
        \end{tabular}
\vspace{-1mm}
\caption{Visualization of transformed images from different NeuralFuse generators trained by ResNet18 with $1\%$ bit error rate.}
\label{fig:VisualizationImagesALL}
\vspace{-1mm}
\end{figure*}